\documentclass[10pt,twocolumn,letterpaper]{article}

\usepackage[pagenumbers]{cvpr}
\usepackage{times}
\usepackage{epsfig}
\usepackage{graphicx}
\usepackage{amsmath}
\usepackage{amssymb}
\usepackage{booktabs}
\usepackage{float}
\newfloat{algorithm}{tbp}{loa}
\floatname{algorithm}{Algorithm}
\usepackage{multirow}
\usepackage{xcolor}
\usepackage{colortbl}
\usepackage{hyperref}
\definecolor{citeLinkBlue}{HTML}{1F4E8C}
\definecolor{refLinkPurple}{HTML}{8E74C9}
\hypersetup{
  colorlinks=true,
    linkcolor={blue!70!black},
    citecolor={blue!70!black},
    urlcolor={blue!70!black},
}
\usepackage{tabularx}
\usepackage{makecell}
\usepackage{pifont}
\usepackage{cuted}
\usepackage[most]{tcolorbox}
\usepackage{tikz}
\usetikzlibrary{shapes.geometric, arrows.meta, positioning, calc, fit, backgrounds}

\newcommand{\llbracket}{[\![}
\newcommand{\rrbracket}{]\!]}

\makeatletter
\renewcommand\paragraph{\@startsection{paragraph}{4}{\z@}%
  {1ex plus 0.4ex minus 0.2ex}%
  {-1em}%
  {\normalfont\normalsize\bfseries}}
\makeatother

\newcommand{\ours}{WorldRoamBench}

\newcommand{\vlmjudge}{Qwen3-VL-PLUS}
\newcommand{\cmark}{\textcolor{green!50!black}{$\checkmark$}}
\newcommand{\xmark}{\textcolor{red!60!black}{$\times$}}  
\newcommand{\modelhead}[2]{\makecell[c]{\raisebox{-0.15\height}{\includegraphics[height=1.05em]{#1}}\\[-0.15em]{\tiny\bfseries #2}}}
\newcommand{\Act}{\raisebox{-0.3\height}{\includegraphics[height=1.2em]{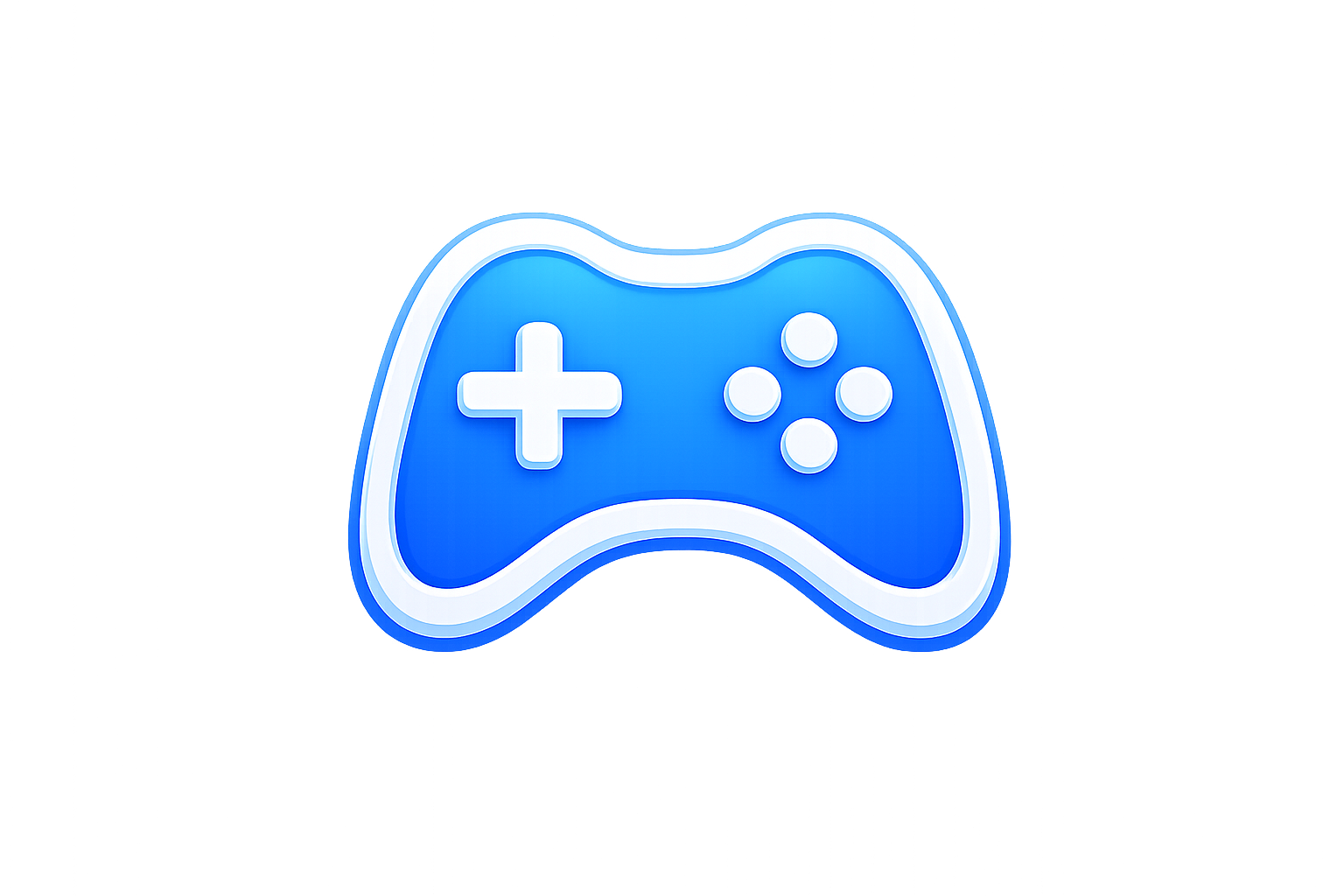}}}
\newcommand{\Pose}{\raisebox{-0.3\height}{\includegraphics[height=1.2em]{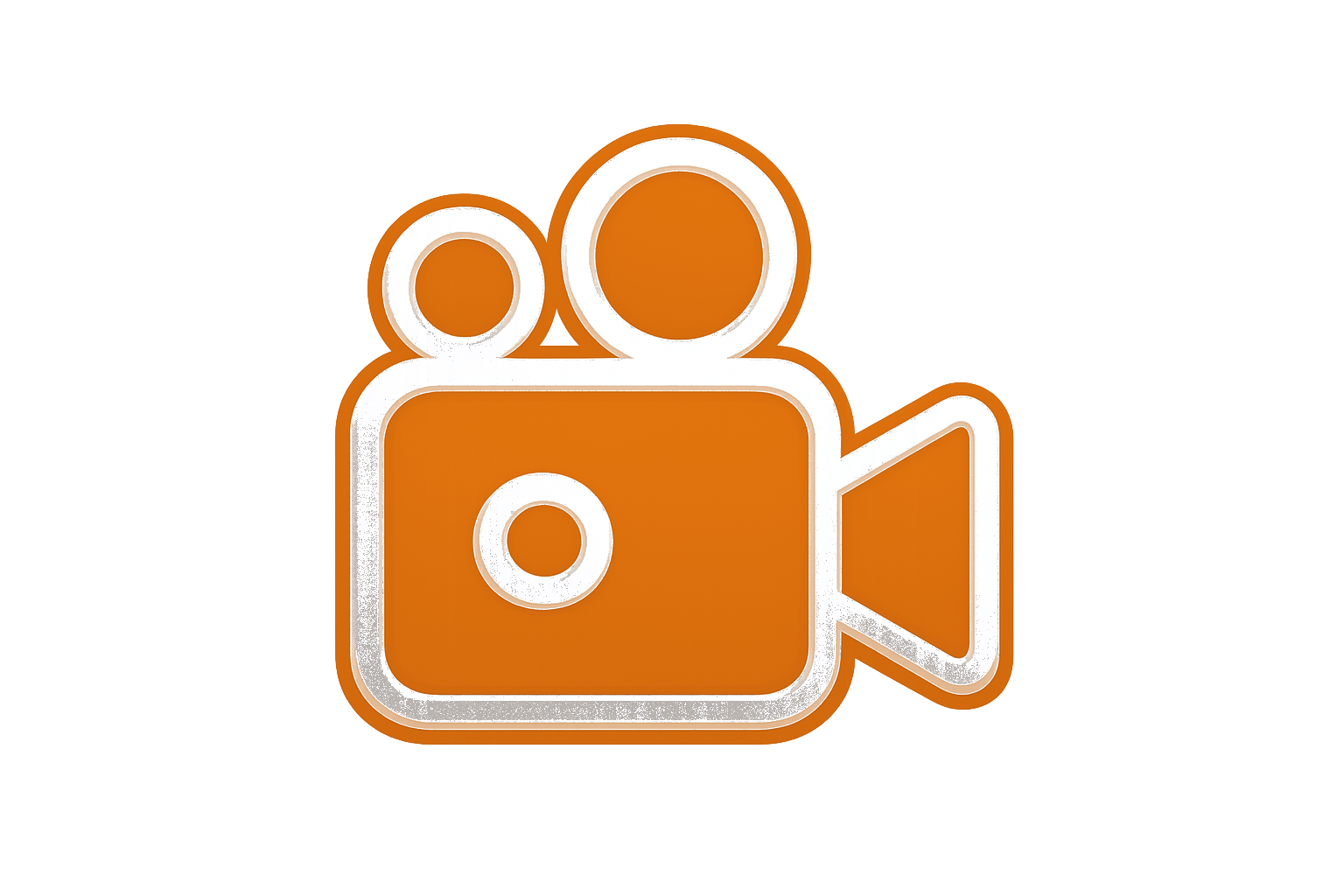}}}
\newcommand{\Txt}{\raisebox{-0.3\height}{\includegraphics[height=1.2em]{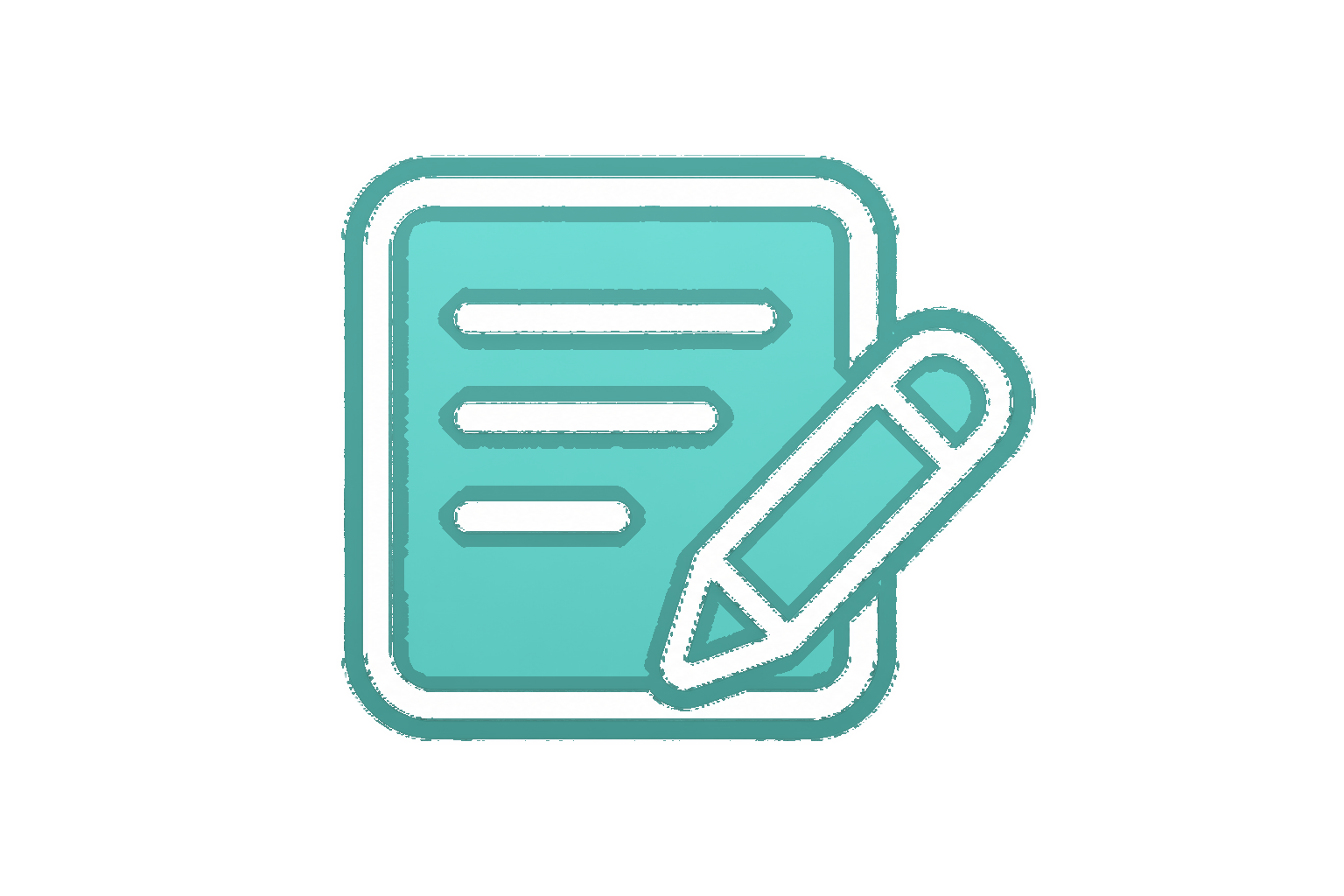}}}
\newcommand{\titlebadge}[3]{%
\href{#1}{%
\tikz[baseline=(badge.base)]{%
\node[
  draw=black!18,
	  fill=black!2,
	  rounded corners=9pt,
	  line width=0.35pt,
	  inner xsep=5pt,
	  inner ysep=1.5pt
	] (badge) {\raisebox{-0.16\height}{\includegraphics[height=0.78em]{#2}}\hspace{0.25em}\textbf{\small #3}};%
}%
}%
}
\definecolor{metricTotal}{HTML}{FFF3C9}
\definecolor{metricScore}{HTML}{DAEAFC}
\definecolor{metricSub}{HTML}{F5FAFA}
\definecolor{closedCol}{HTML}{F2F3F5}
\definecolor{metricAccent}{HTML}{ECE2F8}
\definecolor{metricAction}{HTML}{ECE2F8}
\definecolor{metricVisual}{HTML}{E6F2F3}
\definecolor{metricPhysics}{HTML}{F7E8D7}
\definecolor{metricMemory}{HTML}{DAEAFC}
\definecolor{metricTotalB}{HTML}{EAF1F0}
\definecolor{metricScoreB}{HTML}{DDECDD}
\definecolor{metricSubB}{HTML}{F1F3FA}
\definecolor{closedColB}{HTML}{F0F2F4}
\definecolor{barblue}{HTML}{8DBEF2}
\definecolor{bargold}{HTML}{F0CFA7}
\definecolor{bargreen}{HTML}{A7D7CF}
\definecolor{barmauve}{HTML}{C9B2EA}
\definecolor{barteal}{HTML}{9BD8E0}
\definecolor{barpurple}{HTML}{D5C3F2}
\definecolor{barslate}{HTML}{B9C7D4}
\definecolor{barrose}{HTML}{EDB8C6}
\definecolor{barolive}{HTML}{D9D7A3}
\definecolor{barviolet}{HTML}{BDAAE8}
\definecolor{bargray}{HTML}{D7DDE3}
\definecolor{barsage}{HTML}{C3D8A8}
\definecolor{medalgold}{HTML}{C9932D}
\definecolor{medalsilver}{HTML}{8E9AA8}
\definecolor{medalbronze}{HTML}{B97854}
\definecolor{headerBlue}{HTML}{2F80ED}
\tcbset{
  d3box/.style={
    enhanced,
    colback=blue!2,
    colframe=blue!45!black,
    colbacktitle=blue!12,
    coltitle=black,
    fonttitle=\bfseries,
    boxrule=0.4pt,
    arc=2pt,
    segmentation hidden,
    left=5pt,
    right=5pt,
    top=5pt,
    bottom=5pt
  }
}

\title{%
\texorpdfstring{%
\vspace*{-0.90in}
\begin{minipage}{\textwidth}
\raggedright
\raisebox{-0.18\height}{\includegraphics[height=0.22in]{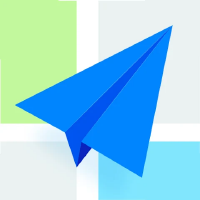}}%
\hspace{0.25em}\raisebox{0.035in}{%
{\sffamily\bfseries\small AMAP CV Lab, Alibaba Group}}\\[-0.68em]
{\color{headerBlue}\rule{\linewidth}{0.55pt}}
\end{minipage}\\[1.15em]
}{}%
\ours{}: An Open-World Benchmark for Long-Horizon Stability of Interactive World Models}
\makeatletter
\patchcmd{\@maketitle}{\vskip .5em}{\vskip 0pt}{}{}
\patchcmd{\@maketitle}{\vspace*{12pt}}{\vspace*{2pt}}{}{}
\makeatother
\author{
\normalsize
Ting-Bing Xu\textsuperscript{1,*}\quad
Jiacheng Sui\textsuperscript{1,*}\quad
Zhe Gao\textsuperscript{2,*}\quad
Kewei Shi\textsuperscript{2,*}\quad
Wenjin Yang\textsuperscript{1}\quad
Zhicheng Liu\textsuperscript{1,$\dagger$}\quad
Zhaoxu Sun\textsuperscript{1}\\
\normalsize
Mingchao Sun\textsuperscript{1}\quad
Hongyu Pan\textsuperscript{1}\quad
Fan Jiang\textsuperscript{1}\quad
Mu Xu\textsuperscript{1,$\ddagger$}\quad
Qi Fan\textsuperscript{2,\S}\quad
Yang Gao\textsuperscript{2}\quad
Yong Li\textsuperscript{3}\quad
Baoquan Chen\textsuperscript{4}\\
{\footnotesize
\textsuperscript{1}AMAP CV Lab, Alibaba Group\quad
\textsuperscript{2}Nanjing University\quad
\textsuperscript{3}Tsinghua University\quad
\textsuperscript{4}Peking University}\\[6pt]
\normalsize
\titlebadge{https://worldroam.amap.com}{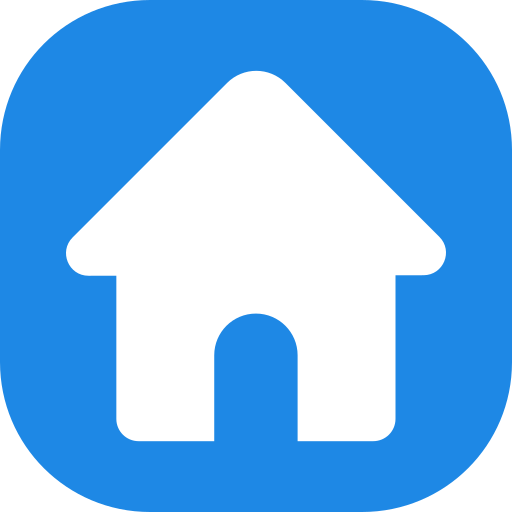}{Homepage}\quad
\titlebadge{https://worldroam.amap.com/\#dataset}{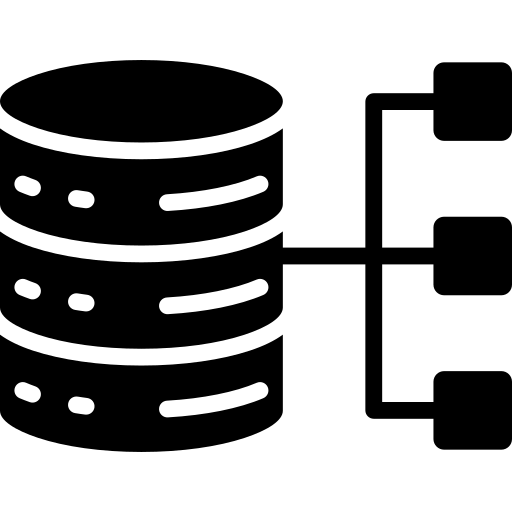}{Dataset}\quad
\titlebadge{https://worldroam.amap.com}{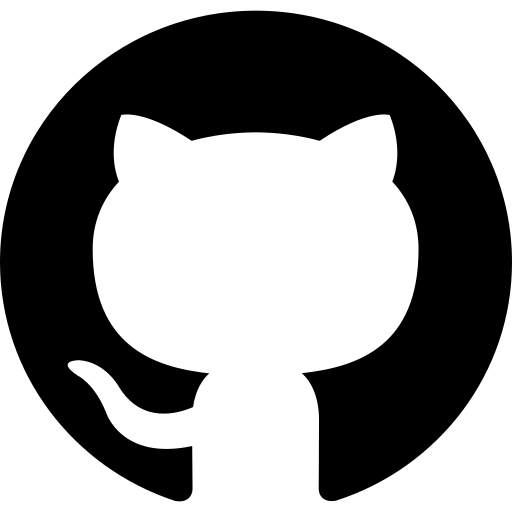}{GitHub}\quad
\titlebadge{https://worldroam.amap.com/\#leaderboard}{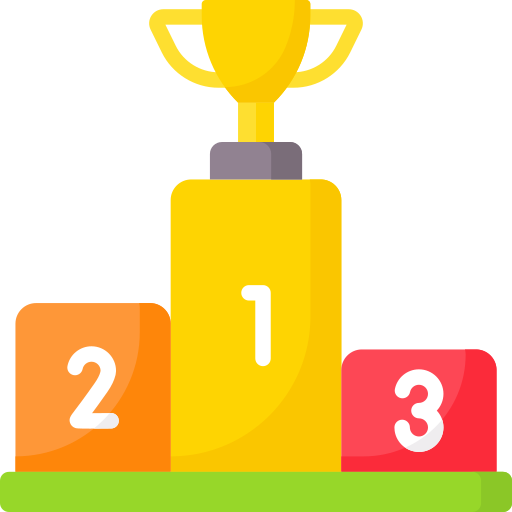}{Leaderboard}
}

\begin{document}
\maketitle
\begingroup
\renewcommand{\thefootnote}{\fnsymbol{footnote}}
\makeatletter
\renewcommand{\@makefntext}[1]{\noindent#1}
\makeatother
\footnotetext[1]{\scriptsize\raggedright\textsuperscript{*}Equal contribution; \textsuperscript{$\dagger$}Project Leader; \textsuperscript{$\ddagger$}Project Sponsor.\\
\textsuperscript{\S}Corresponding author: Qi Fan (\href{mailto:fanqi@nju.edu.cn}{fanqi@nju.edu.cn}).}
\endgroup

\begingroup
\setlength{\stripsep}{0.25em}
\begin{strip}
\centering
\includegraphics[width=.92\textwidth]{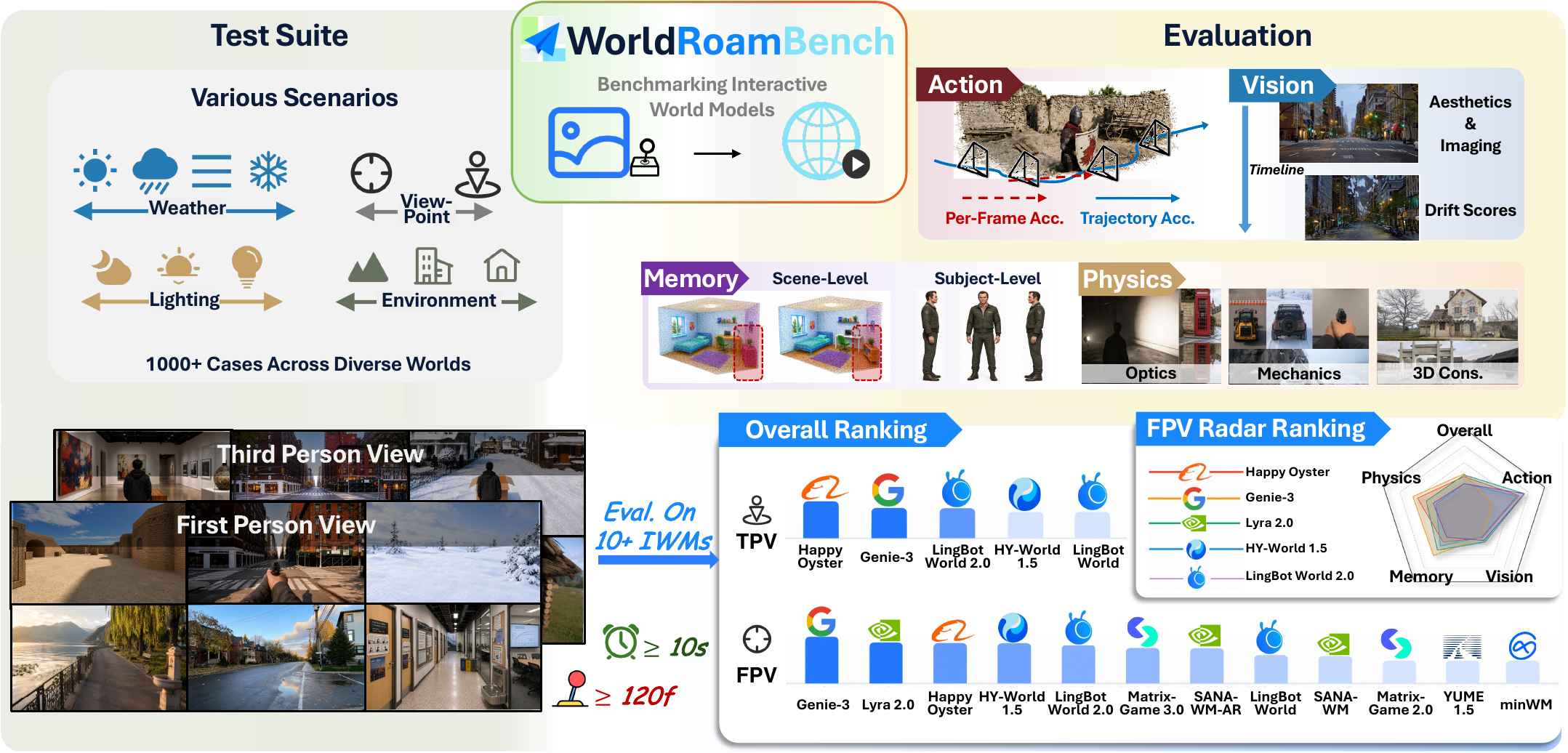}
\captionof{figure}{\textbf{Illustration of \ours{}.} \ours{} evaluates interactive world models across action following, visual quality, memory, and interaction physics, covering open-source and closed-source models in diverse scenarios and viewing conditions.}
\label{fig:teaser}
\vspace{0.5em}
\end{strip}
\endgroup

\begin{abstract}
Despite rapid progress in interactive world models (IWMs), short-horizon performance does not establish sustained action following, visual stability, physical plausibility, or memory. We introduce \ours{}, an open-world benchmark for \emph{long-horizon stability} across four dimensions, each with innovations: (i)~\textbf{Action}: per-frame action metric bypassing cross-model semantic scale disparity and exposing failures hidden by trajectory; (ii)~\textbf{\mbox{Vision}}: sliding-window drift metric capturing non-monotonic mid-sequence collapse missed by start-vs-end comparisons; (iii)~\textbf{Physics}: evaluation of physical plausibility across mechanics, optics, and 3D consistency, gated by camera-motion and subject-tracking checks; (iv)~\textbf{\mbox{Memory}}: a trajectory-aware protocol reducing confounding from action-following errors, evaluating scene memory via transition-localized 3D point-cloud reconstruction and subject memory via tracking-plus-VLM reasoning. The benchmark comprises 1000+ test cases across Nature, Urban, and Indoor scenes in first/third-person views with WASD 10--60\,s continuous interaction. Evaluating 10+ open/closed-source models reveals none reliably satisfies all dimensions; even the best achieves only moderate scores. Advances on \ours{} are steps toward IWMs that are stable, physically grounded, memory-faithful, and deployable in real-world applications.
\end{abstract}

\section{Introduction}
\label{sec:intro}

Imagine exploring an infinite, AI-generated world by pressing WASD keys on your keyboard: moving forward through forests, turning to gaze at distant mountains, and returning to find the same landscape waiting as you left it.
This vision of interactive world models (IWMs)~\cite{genie2024,matrixgame2,hyworld15} is rapidly approaching reality: recent models generate photorealistic environments at interactive frame rates, accepting discrete keyboard inputs to control camera movement.

Yet as these models proliferate, a critical question remains unanswered: \textit{How well do they respond to user inputs over extended interactions?}
Existing benchmarks (\Cref{tab:comparison}), such as MIND~\cite{mind2026}, WorldMark~\cite{worldmark2026}, iWorld-Bench~\cite{iworldbench2026}, WildWorld~\cite{wildworld2026}, WBench~\cite{ying2026wbench}, and WorldOlympiad~\cite{worldolympiad2026}, have advanced evaluation methodology. However, most evaluate only short clips of ${\sim}$5--10s~\cite{mind2026,wildworld2026,iworldbench2026,ying2026wbench}, leaving \emph{long-horizon exploration} largely unexamined. Moreover, most lack systematic evaluation of \emph{interaction physics}, such as mechanics, optics, and 3D consistency~\cite{mind2026,worldmark2026,iworldbench2026,wildworld2026,ying2026wbench}, and only a few assess \emph{memory consistency} during revisit, often under the assumption of perfect action execution~\cite{mind2026,iworldbench2026}.

\begin{table*}[t]
\centering
\small
\caption{\textbf{Comparison with existing interactive world model benchmarks.} \ours{} jointly evaluates discrete-action accuracy, visual drift, interaction physics, and trajectory-aware scene/subject memory. $^\dagger$WBench and iWorld-Bench report 10+ models because they include general video generation models beyond strictly interactive world models. $^\ddagger$WBench, iWorld-Bench, and WorldOlympiad include a small number of third-person cases but do not provide a dedicated third-person view (TPV) leaderboard.}
\label{tab:comparison}
\setlength{\tabcolsep}{3pt}
\renewcommand{\arraystretch}{1.4}
\resizebox{\textwidth}{!}{%
\begin{tabular}{l|ccccccc|cc|cc|cc}
\toprule
& \multicolumn{7}{c|}{\textbf{Eval. Metrics}} & \multicolumn{2}{c|}{\textbf{Eval. Models}} & \multicolumn{2}{c|}{\textbf{Leaderboards}} & \multicolumn{2}{c}{\textbf{Categories}} \\
\cmidrule(lr){2-8} \cmidrule(lr){9-10} \cmidrule(lr){11-12} \cmidrule(lr){13-14}
\textbf{Benchmark} & \textbf{Action Eval.} & \textbf{Discrete-action} & \textbf{Visual} & \textbf{Interaction} & \textbf{Revisit} & \textbf{Traj.-Aware} & \textbf{Time} & \textbf{Open-} & \textbf{Closed-} & \textbf{First-} & \textbf{Third-} & \textbf{Game} & \textbf{Real} \\
 & \textbf{Granularity} & \textbf{Accuracy} & \textbf{Drift} & \textbf{Physics} & \textbf{Memory} & \textbf{Memory} & \textbf{Scale/Video} & \textbf{Source} & \textbf{Source} & \textbf{Person} & \textbf{Person} & \textbf{World} & \textbf{World} \\
\midrule
WildWorld~\cite{wildworld2026} & Segment & \xmark & \xmark & \xmark & \xmark & \xmark & ${\sim}$5--10s & 5 & 0 & \cmark & \xmark & \cmark & \xmark \\
iWorld-Bench~\cite{iworldbench2026}$^{\dagger\ddagger}$ & Traj. & \xmark & \xmark & \xmark & \cmark & \xmark & ${\sim}$5--10s & 10+ & 0 & \cmark & \xmark & \cmark & \cmark \\
WorldOlympiad~\cite{worldolympiad2026}$^\ddagger$ & Multi-chunk & \xmark & \xmark & \cmark & \xmark & \xmark & ${\sim}$10s & 8 & 0 & \cmark & \xmark & \cmark & \cmark \\
MIND~\cite{mind2026} & Traj. & \xmark & \xmark & \xmark & \cmark & \xmark & ${\sim}$5--10s & 2 & 0 & \cmark & \cmark & \cmark & \xmark \\
WBench~\cite{ying2026wbench}$^{\dagger\ddagger}$ & Multi-turn & \xmark & \xmark & \cmark & \cmark & \xmark & ${\sim}$5--10s & 10+ & 2 & \cmark & \xmark & \cmark & \cmark \\
WorldMark~\cite{worldmark2026} & Frame/segment & \xmark & \xmark & \xmark & \cmark & \cmark & 20/40/60s & 10 & 0 & \cmark & \cmark & \xmark & \cmark \\
\midrule
\textbf{\ours{} (Ours)} & \textbf{Traj. + per-frame} & \textbf{\cmark} & \textbf{\cmark} & \textbf{\cmark} & \textbf{\cmark} & \textbf{\cmark} & \textbf{10--60s} & \textbf{10} & \textbf{2} & \textbf{\cmark} & \textbf{\cmark} & \textbf{\cmark} & \textbf{\cmark} \\
\bottomrule
\end{tabular}%
}
\renewcommand{\arraystretch}{1.0}
\end{table*}

\begin{figure}[t]
\centering
\includegraphics[width=\columnwidth]{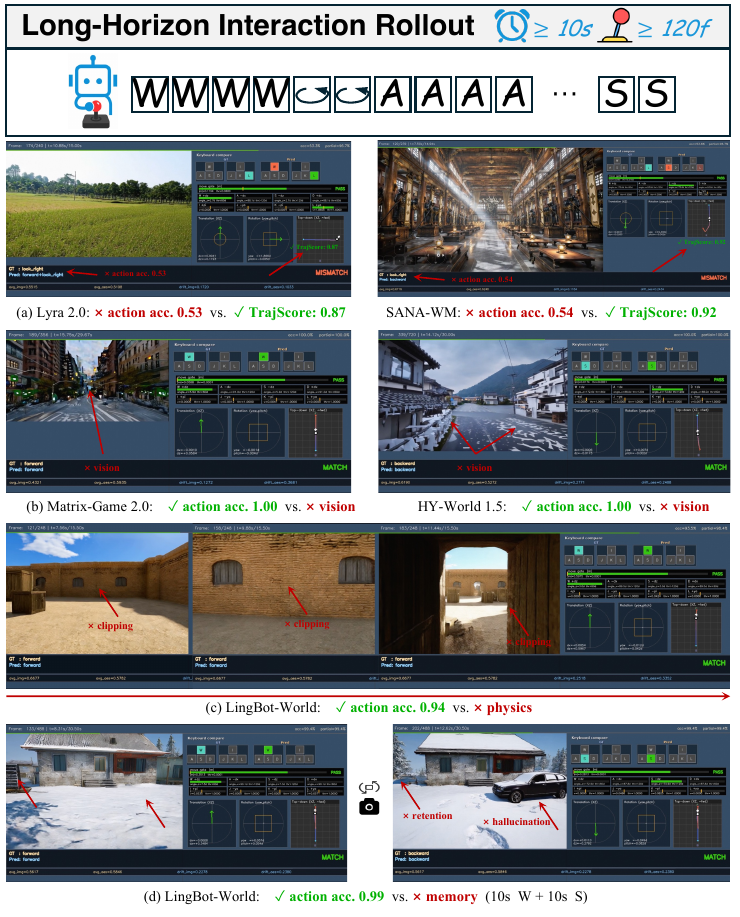}
\caption{\textbf{Failures revealed by long-horizon interaction.} Extended rollouts expose failures often missed by short clips: high trajectory scores hide per-step action mismatches, visual quality degrades, physical constraints are violated, and revisited scenes are regenerated inconsistently.}
\label{fig:intro_failures}
\end{figure}

We argue the true challenge for interactive world models lies in \emph{long-horizon stability}: maintaining faithful, physically plausible, visually coherent, and memory-consistent generation as users continuously roam the generated world. Four critical aspects emerge where current models are most fragile under extended interaction:
1)~\textbf{Long-horizon action following.} Trajectory-based benchmarks~\cite{mind2026,iworldbench2026,ying2026wbench} recover 3D camera trajectories from generated videos and evaluate geometric errors such as RPE and ATE\footnote{RPE: Relative Pose Error; ATE: Absolute Trajectory Error.}. These metrics have two fundamental limitations: first, cross-model semantic scale disparity, where models produce vastly different displacement and rotation magnitudes for identical actions, makes trajectory comparison inherently unfair; second, trajectory alignment can mask per-frame failures, as a model may ignore a keystroke then over-compensate, yielding acceptable trajectory error but poor frame-level fidelity. As shown in \Cref{fig:intro_failures}(a), models frequently mispredict discrete actions at switch boundaries, a failure mode invisible to trajectory-level evaluation.
2)~\textbf{Long-horizon visual stability.} Existing benchmarks measure average visual quality (Imaging and Aesthetic scores) but rarely assess \emph{temporal degradation}. In long-horizon rollout, autoregressive error accumulation causes progressive quality collapse: color drift, blurring, or structural breakdown. \Cref{fig:intro_failures}(b) shows a model may follow actions correctly yet produce visually collapsed frames after hundreds of generation steps.
3)~\textbf{Long-horizon interaction physics.} Recent benchmarks such as WBench~\cite{ying2026wbench} and WorldOlympiad~\cite{worldolympiad2026} assess interaction physics. We examine physical consistency under sustained user control, including collision response, clipping, gravity, optical consistency, and rigid-body 3D coherence. \Cref{fig:intro_failures}(c) shows a model following the intended direction yet violating physical constraints (e.g., passing through obstacles).
4)~\textbf{Long-horizon memory.} MIND~\cite{mind2026} and iWorld-Bench~\cite{iworldbench2026} evaluate memory through symmetric revisit tests, comparing frames at index $t$ and $T{-}t$. However, this assumes perfectly executed actions, which breaks in practice: cumulative drift makes the ``return'' frame spatially offset. Moreover, \Cref{fig:intro_failures}(d) shows that even when actions prescribe an ideal symmetric return path, models regenerate a different scene upon revisit, revealing memory limitations.

To fill these gaps, we introduce \ours{}---\textbf{R}esponsive \textbf{O}pen-world \textbf{A}ction \textbf{M}odel \textbf{Bench}mark---which focuses on \emph{long-horizon interactive stability} (\cref{fig:teaser}). Unlike benchmarks evaluated on brief ${\sim}$5--10s clips, \ours{} evaluates models as they \emph{roam} through generated open-domain worlds for ${\sim}$10--60s of continuous WASD/IJKL interaction, covering all four dimensions: \hyperref[sec:action]{action following}, \hyperref[sec:visual]{visual quality}, \hyperref[sec:physics]{interaction physics}, and \hyperref[sec:memory]{memory}. Contributions are as follows:

\begin{enumerate}
    \item \textbf{Action.} We propose a keystroke-level per-frame action metric that bypasses cross-model semantic scale disparity and reveals per-step failures hidden by trajectory.
    
    \item \textbf{Vision.} We introduce a sliding-window temporal drift metric over imaging and aesthetic scores that captures non-monotonic mid-sequence collapse missed by per-frame or start-vs-end measures.
    
    \item \textbf{Physics.} We design a validity-gated physics evaluation over \emph{mechanics} (collision, clipping, deformation, gravity), \emph{optics} (reflection, shadow occlusion), and \emph{3D consistency}, scoring only rollouts that provide usable motion evidence: pose-verified response in FPV and continuous, trackable subject evidence in TPV.
    
    \item \textbf{Memory.} We present a trajectory-aware dual-track protocol, designed to \emph{reduce} the sensitivity of memory scoring to action-following error, that evaluates \emph{scene memory} via transition-localized 3D point-cloud reconstruction and \emph{subject memory} via tracking-plus-\vlmjudge{} reasoning. 

    \item \textbf{Leaderboard.} We build the most comprehensive leaderboard to date, evaluating 10+ state-of-the-art open- and closed-source IWMs on 1000+ test cases, with dedicated leaderboards for first- and third-person views and short-, mid-, and long-horizon rankings.
\end{enumerate}

Throughout our evaluation, we identify three recurring weaknesses in current IWMs: \textbf{(i) limited long-horizon stability.} From short to long rollouts, visual quality declines, action accuracy drops for most models, and memory deteriorates most sharply, exposing failures obscured by short clips. Sensitivity to action switching also varies substantially across models; \textbf{(ii) weak adherence to physical constraints.} Models can follow the commanded direction yet generate physically implausible motion, such as passing through solid obstacles. Even leading models achieve only moderate mechanics scores, while open-source models lag behind; and \textbf{(iii) unreliable long-term memory.} Most models struggle to retain previously observed scene content, often regenerating inconsistent geometry upon revisit. Reliable long-term memory is therefore still rare among IWMs.

\begin{figure*}[t]
\centering
\includegraphics[width=0.98\textwidth]{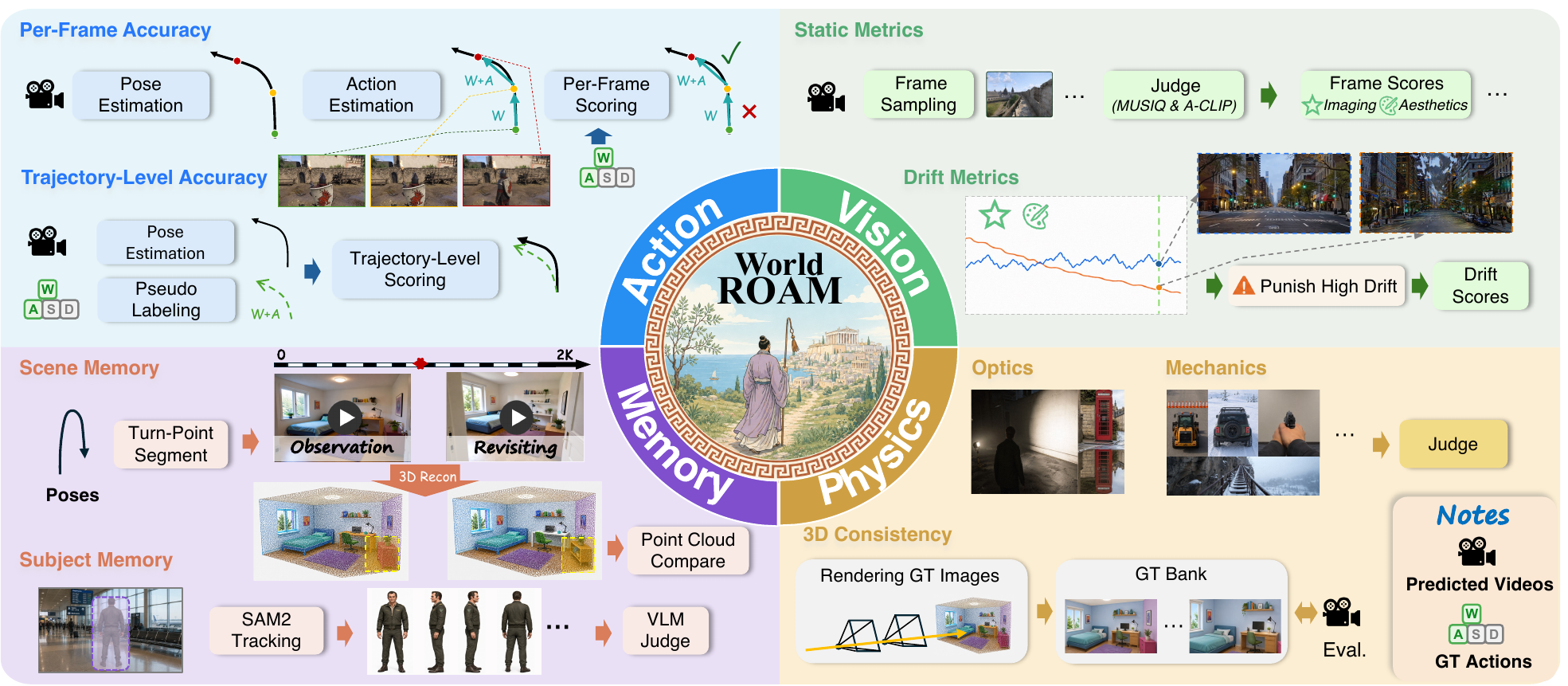}
\caption{\textbf{Overview of the \ours{} evaluation pipeline.} Given videos generated from shared initial frames and WASD/IJKL action programs, \ours{} evaluates four complementary dimensions: action following through pose-aligned frame scoring, visual quality through frame-level and drift-aware metrics, interaction physics through mechanics, optics, and 3D-consistency tests, and memory through turning-point-aware scene reconstruction and third-person subject memory.}
\label{fig:roam-architecture}
\end{figure*}

\section{Related Work}
\label{sec:related}

\subsection{Interactive Video World Models}
Unlike passive video generation models such as Wan~2.7~\cite{wan2025}, Kling~3.0~\cite{kling3_2025}, Sora~2~\cite{sora2_2025}, Veo~3.1~\cite{veo3_2025}, and Seedance~2.0~\cite{seedance2_2025}, which synthesize videos from text or image prompts without real-time user control, interactive world models generate visual environments conditioned on user actions in real time.
Genie~\cite{genie2024} pioneered this paradigm by learning latent action spaces from unlabeled video.
Recent models explore three overlapping control modalities.
\textbf{Discrete-action control}, used by Matrix-Game~2.0/3.0~\cite{matrixgame2,matrixgame3} and HY-World~1.5~\cite{hyworld15}, conditions generation on keyboard commands and can be combined with camera-pose inputs.
\textbf{Camera-pose control}, adopted by LingBot-World/2.0~\cite{lingbotworld,lingbot-world-v2}, SANA-WM/AR~\cite{sanawm}, Lyra~2.0~\cite{lyra2_2026}, and minWM~\cite{minwm2026}, specifies motion through continuous camera trajectories.
\textbf{Language-based control}, exemplified by Yume~1.5~\cite{yume15}, uses text instructions to guide motion and world events.
Closed-source models such as Genie~3~\cite{genie3_2025} (Google DeepMind) and Happy Oyster~\cite{happyoyster2026} (Alibaba) further push visual quality and temporal coherence.
To compare models across these control interfaces, we use a common benchmark protocol: an initial image and a sequence of discrete actions $\{a_1, a_2, \ldots, a_T\}$, where $a_t \in \{\text{W, A, S, D, I, J, K, L, idle}\}$, passed directly or translated by deterministic adapters before the model produces a video rollout $\{f_1, f_2, \ldots, f_T\}$.

\subsection{Benchmarks for World Models}
Evaluation benchmarks for world models span several paradigms.
\textbf{Video generation} benchmarks (VBench~\cite{vbench2024,vbenchpp2024,vbench2_2025}, WorldScore~\cite{worldscore2025}, WorldModelBench~\cite{worldmodelbench2025}) assess visual quality, text alignment, or physical commonsense without closed-loop interactive control.
\textbf{Domain-specific} benchmarks target narrow scenarios: ACT-Bench~\cite{actbench2024} and WorldLens~\cite{worldlens2025} for autonomous driving; EWMBench~\cite{ewmbench2025} and WorldArena~\cite{worldarena2026} for embodied robotic manipulation.
\textbf{Physics and interaction fidelity} benchmarks (PhyGenBench~\cite{phygenbench2024}, WorldBench~\cite{worldbench2026ucla}, Omni-WorldBench~\cite{omniworldbench2026}) evaluate physical plausibility or causal interaction correctness on prompt-conditioned generation without discrete-keystroke control.
\textbf{Interactive world model} benchmarks are most relevant to our setting.
MIND~\cite{mind2026} introduced memory via closed-loop revisit tests; WorldMark~\cite{worldmark2026} unified WASD actions with three difficulty tiers; iWorld-Bench~\cite{iworldbench2026} proposed a control quadruple with six levels; WBench~\cite{ying2026wbench} unified multi-modal inputs for multi-turn evaluation; WorldOlympiad~\cite{worldolympiad2026} jointly assessed physics, geometry, and interaction across chunk-by-chunk rollouts; and WildWorld~\cite{wildworld2026} evaluated gamepad-based action following in a single game.

However, as shown in \Cref{tab:comparison}, interactive benchmarks leave several aspects underexplored: most are limited to short horizons of ${\sim}$5--10s, action evaluation remains largely trajectory-level despite cross-model semantic scale disparity, long-horizon interaction physics is rarely measured, and memory protocols often assume perfect ground-truth (GT) frame alignment.
\ours{} jointly evaluates action following, visual drift, interaction physics, and scene/subject memory in open-world, long-horizon (10--60s) interactions. Its action suite crosses three rollout durations with three switch counts, enabling separate comparisons across durations and switch counts.

\subsection{Action Evaluation Methodologies}
Action evaluation in world models generally follows four paradigms.
\textbf{IDM-based} approaches (VPT~\cite{vpt2022}, MineWorld~\cite{mineworld2025}) train an Inverse Dynamics Model to classify per-frame actions from video, achieving high accuracy in Minecraft but limited to that domain.
\textbf{Trajectory-based} approaches (MIND~\cite{mind2026}, iWorld-Bench~\cite{iworldbench2026}) recover 3D camera poses via SLAM or ViPE and compute geometric errors.
\textbf{Flow-based} approaches, exemplified by WorldMark~\cite{worldmark2026}, characterize action responses through direction accuracy, direction purity, response latency, and motion stability.
\textbf{VLM-based} approaches (WildWorld~\cite{wildworld2026}) use vision-language models to judge action consistency at the segment level.
\ours{} combines pose-decoded strict and partial action accuracy with TrajScore to assess command agreement and geometric fidelity.

\subsection{Physics Evaluation Methodologies}

Physics evaluation has been explored in both general video generation and world-model benchmarks.
General video benchmarks such as VBench~\cite{vbench2024}, VBench-2.0~\cite{vbench2_2025}, and OmniBench~\cite{omnibench2025} include physical plausibility among broader visual and temporal quality dimensions.
Physics-focused benchmarks further define explicit physical reasoning tasks: PhyGenBench~\cite{phygenbench2024} covers diverse physical laws, WorldModelBench~\cite{worldmodelbench2025} evaluates violations such as gravity, penetration, fluid dynamics, and conservation, and WM-ABench~\cite{wmabench2025} studies mechanical interactions through controlled simulation.
Recent world-model benchmarks also touch physics from different angles: WorldScore~\cite{worldscore2025} includes 3D and photometric consistency, WorldBench~\cite{worldbench2026ucla} evaluates physical plausibility in generated worlds, WorldLens~\cite{worldlens2025} studies closed-loop degradation, and Omni-WorldBench~\cite{omniworldbench2026} introduces multi-level causal interaction scenarios.

However, existing physics tests are still mostly designed for passive observations or coarse semantic judgments.
They rarely ask whether a user's action should trigger a specific physical consequence, and they do not provide case-specific metrics for contact, occlusion, reflection, shadows, or traversability.
\ours{} complements prior benchmarks by evaluating fine-grained interaction physics in 3D-grounded scenes under explicit WASD/IJKL control.

\subsection{Memory Evaluation Methodologies}
Existing memory evaluation protocols commonly rely on frame-pair similarity.
MIND~\cite{mind2026}, for example, constructs symmetric action sequences and compares temporally symmetric frames using pixel-level metrics such as LPIPS.
This formulation implicitly assumes perfect action execution and uniform action magnitude, such that symmetric temporal indices correspond to the same spatial location.
However, interactive models may drift, under- or over-execute commands, and accumulate timing offsets, causing frame-pair metrics to conflate control error with memory degradation.
WBench~\cite{ying2026wbench} partially addresses temporal misalignment by estimating poses and selecting the revisit frame closest to the initial frame, but the resulting metric remains frame-level and first-frame-centric, rather than assessing preservation of the broader visited scene.

A related line of work evaluates short-term geometric consistency rather than memory.
WBench's spatial consistency, WorldMark's geometry consistency~\cite{worldmark2026}, and WorldOlympiad's geometric evaluation~\cite{worldolympiad2026} measure whether nearby generated views remain 3D-coherent.
While such metrics characterize local spatial consistency, they do not evaluate whether a model retains previously generated or observed content after a long interactive loop.
\ours{} therefore formulates memory as revisit evaluation under imperfect control, combining action-aware transition localization with scene-level 3D reconstruction to assess preservation beyond a single frame pair.

\section{\ours{} Benchmark Design}
\label{sec:method}

\subsection{Overview}
\ours{} evaluates interactive world models through four questions (\cref{fig:roam-architecture}):
(1)~\textbf{Action Following}: does the model respond correctly to each keystroke?
(2)~\textbf{Visual Quality}: does the generated world remain perceptually faithful and temporally stable?
(3)~\textbf{Interaction Physics}: does the world obey basic physical laws during interaction?
(4)~\textbf{Memory}: does the model preserve what it previously generated?
Together, these dimensions provide a comprehensive assessment of interactive world models, with each capturing capabilities not fully reflected by the others.

\subsection{Evaluation Dimension 1: Action Following}
\label{sec:action}

\begin{figure}[t]
\centering
\includegraphics[width=\columnwidth]{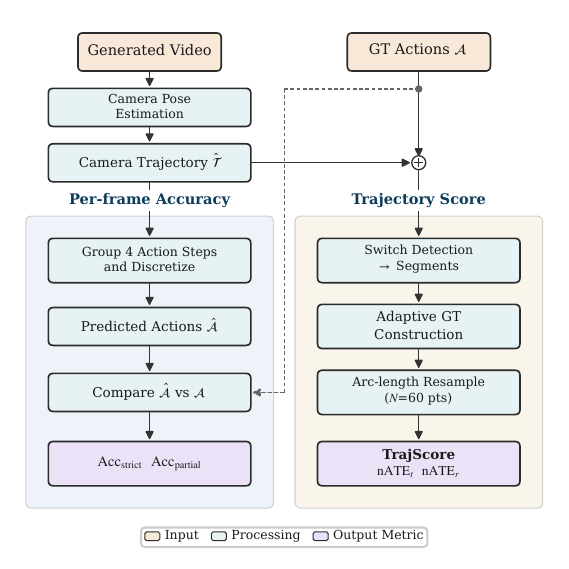}
\caption{\textbf{Action following evaluation.} Estimated camera poses support two complementary metrics: discrete action agreement over short motion windows and geometric alignment with an action-conditioned reference trajectory.}
\label{fig:action_pipeline}
\end{figure}

We evaluate action following through two complementary measures: \emph{discrete action accuracy}, which assesses frame-level command agreement, and \emph{TrajScore}, which assesses the geometric fidelity of the camera path. Both use the estimated camera trajectory (\Cref{fig:action_pipeline}).

\subsubsection{Per-frame Discrete Action Accuracy}
\label{sec:per_frame_action}

Let $\mathcal{A}=(a_1,\ldots,a_T)$ be the command sequence, with $a_t$ controlling the transition from $I_{t-1}$ to $I_t$. The atomic vocabulary comprises eight primitives: \texttt{forward}\,(W), \texttt{backward}\,(S), \texttt{left}\,(A), \texttt{right}\,(D), \texttt{look\_left}\,(J, yaw$-$), \texttt{look\_right}\,(L, yaw$+$), \texttt{look\_up}\,(I, pitch$+$), and \texttt{look\_down}\,(K, pitch$-$). Commands may also be valid combinations of these primitives or \texttt{no\_op}.

\paragraph{Pose-based action decoding.}
From the aligned video, we estimate camera-to-world poses $\hat{\mathcal{T}}=(\hat{P}_0,\ldots,\hat{P}_T)$. We use MegaSaM~\cite{megasam} with a DA-v3~\cite{depthanythingv3} depth prior, selected for its highest mean action-decoding ceiling among six pose pipelines (\Cref{sec:appendix_pose_pipelines}).

Following CompassReward~\cite{worldcompass2026}, we infer actions from relative camera motion. Translation direction determines movement primitives, while signed yaw and pitch determine look primitives; magnitude thresholds suppress negligible motion. We infer one action for each non-overlapping group of four frames; the window-size ablation is reported in \Cref{sec:appendix_stride_ablation}.

The same action can induce different translation magnitudes across models and scenes. We therefore calibrate a fixed set of four translation-threshold candidates for each pose source using cross-model statistics. One candidate is selected per video, yielding a decoded sequence $\hat{\mathcal{A}}$ shared by all discrete metrics. Calibration and candidate selection are detailed in \Cref{sec:appendix_compass}.

\paragraph{Strict accuracy.}
Strict accuracy requires the predicted action to match the ground-truth command exactly, including every component of a compound action:
\begin{equation}
    \mathrm{Acc}_\mathrm{strict}
    = \frac{1}{T}\sum_{t=1}^{T}\llbracket \hat{a}_t=a_t\rrbracket,
    \label{eq:strict_accuracy}
\end{equation}
where $\llbracket\cdot\rrbracket$ is the indicator function.

\paragraph{Partial accuracy.}
A model may execute the commanded movement while introducing unintended rotation. Partial accuracy distinguishes such partial compliance from a complete mismatch. Let $\mathcal{P}(a)$ denote the set of atomic components of action $a$. Partial accuracy measures the fraction of frames in which the predicted and ground-truth actions share at least one component:
\begin{equation}
    \mathrm{Acc}_\mathrm{partial}
    = \frac{1}{T}\sum_{t=1}^{T}
    \llbracket \mathcal{P}(\hat{a}_t)\cap
    \mathcal{P}(a_t)\neq\emptyset\rrbracket.
    \label{eq:partial_accuracy}
\end{equation}
For example, if the command is \texttt{forward} but the prediction is \texttt{forward+look\_right}, the two component sets intersect at $\{\texttt{forward}\}$. The correct translation yields a partial match, while the unintended rotation prevents a strict match. The gap between partial and strict accuracy measures how often predictions are only partially correct; move and look accuracies separately assess translation and rotation agreement.

\subsubsection{Trajectory-level Metrics: TrajScore}
\label{sec:trajscore}

Discrete labels do not capture geometric deviations within an action: a path may be classified as \texttt{forward} despite substantial lateral drift. Inspired by WBench~\cite{ying2026wbench}, TrajScore complements discrete action accuracy by evaluating camera-path geometry through adaptive reference construction and arc-length alignment.

\paragraph{Adaptive reference construction.}
A key design choice in \ours{} is an \emph{adaptive reference trajectory}: subject to minimum-motion safeguards, it retains the model's own motion magnitude while imposing the ideal direction and shape prescribed by the action. This separates \emph{how far the model moves} from \emph{whether it follows the commanded geometry}, without requiring a fixed movement distance or turning angle across models.

We partition the command sequence into constant-action segments and construct an ideal path for each. For pure translation commands, the reference is a straight line in the commanded direction whose length matches the predicted segment's displacement. Look commands prescribe rotation in place for first-person views and an orbit around the subject for third-person views, with the orbit radius estimated from the predicted trajectory. Compound commands couple movement and turning: e.g., \texttt{forward+look\_right} in first-person view defines a curved path that advances while turning right.

\paragraph{Arc-length alignment.}
Direct frame-wise comparison between the predicted and reference trajectories can conflate geometric deviations with differences in traversal speed. Within each constant-action segment, we therefore resample both trajectories at equal fractions of their respective arc lengths to establish point correspondences. We compute $\mathrm{ATE}_t$ as the mean Euclidean position error and $\mathrm{ATE}_r$ as the mean angular orientation error between these paired samples. Reference construction, safeguards, and sampling details, including the treatment of zero-length paths, are provided in \Cref{sec:appendix_trajscore}.

\paragraph{Normalized geometric error.}
To express geometric errors relative to the amount of motion, we normalize translation and rotation errors by the predicted path length $L_\mathrm{path}$ and accumulated rotation $\Phi_\mathrm{total}$, respectively:
\begin{align}
    \mathrm{nATE}_t &= \mathrm{clamp}\!\left(\frac{\mathrm{ATE}_t}{\max(L_\mathrm{path},\,\ell_\mathrm{min})},0,1\right),\\
    \mathrm{nATE}_r &= \mathrm{clamp}\!\left(\frac{\mathrm{ATE}_r}{\max(\Phi_\mathrm{total},\,\phi_\mathrm{min})},0,1\right).
\end{align}
The denominator floors $\ell_\mathrm{min}$ and $\phi_\mathrm{min}$ stabilize low-motion cases. The trajectory score is
\begin{equation}
    \mathrm{TrajScore}=1-\frac{\mathrm{nATE}_t+\mathrm{nATE}_r}{2},
\end{equation}
with higher values indicating closer geometric alignment. TrajScore thus measures relative geometric fidelity, complementing frame-level command agreement.

\subsubsection{Action Score}
We aggregate the complementary scores by view:
\begin{equation}
    S_\mathrm{action}=
    \begin{cases}
    \frac{\mathrm{Acc}_\mathrm{strict}+\mathrm{Acc}_\mathrm{partial}+\mathrm{TrajScore}}{3}, & \mathrm{FPV},\\[6pt]
    \frac{\mathrm{Acc}_\mathrm{partial}+\mathrm{TrajScore}}{2}, & \mathrm{TPV}.
    \end{cases}
\end{equation}
In TPV, subject-centered camera orbits couple rotation and translation, potentially yielding compound predictions that fail strict matching; partial matching still credits the correctly decoded look action.
We also report normalized translation and rotation errors as diagnostics.

\subsection{Evaluation Dimension 2: Visual Quality}
\label{sec:visual}

Autoregressive world models must maintain visual quality not only on average but also \emph{over time}, as quality often degrades when generation errors accumulate. We therefore evaluate visual quality along two axes---\emph{absolute quality} and \emph{temporal drift}---using four complementary metrics.

\paragraph{Aesthetic Quality.}
The aesthetic appeal of generated frames is measured using a CLIP-based linear aesthetic predictor~\cite{schuhmann2022laion5b}. Each frame $I_t$ is encoded by a frozen CLIP ViT-L/14 backbone into a normalized feature vector $\mathbf{z}_t = \mathrm{norm}(\mathrm{CLIP}(I_t))$. A linear head $g:\mathbb{R}^{768}\!\to\!\mathbb{R}$, trained on human aesthetic ratings from the LAION-Aesthetics dataset, produces a per-frame score $s^\mathrm{aes}_t = g(\mathbf{z}_t)/10$. The video-level aesthetic quality is the mean over all frames:
\begin{equation}
    S_\mathrm{aes} = \frac{1}{T}\sum_{t=1}^{T} s^\mathrm{aes}_t.
\end{equation}

\paragraph{Imaging Quality.}
Low-level perceptual quality (sharpness, noise, compression artifacts) is assessed using MUSIQ~\cite{ke2021musiq} with the SPAQ-pretrained checkpoint, a multi-scale image quality transformer that operates without a reference image. MUSIQ processes each frame at multiple resolutions via a multi-scale representation and outputs a technical quality score $s^\mathrm{img}_t = \mathrm{MUSIQ}(I_t)/100$. The video-level imaging quality is:
\begin{equation}
    S_\mathrm{img} = \frac{1}{T}\sum_{t=1}^{T} s^\mathrm{img}_t.
\end{equation}

\paragraph{Temporal Drift Metrics.}
A defining failure mode of autoregressive world models is progressive quality degradation: early frames may be crisp and aesthetically pleasing, whereas later frames suffer from accumulated artifacts, color shifts, or detail loss. Degradation is not always monotonic: we frequently observe models that \emph{collapse mid-rollout and partially recover}, so contrasting a clip's start and end can dramatically underestimate the true visual instability. We therefore quantify drift over the entire rollout via a sliding-window formulation inspired by Helios~\cite{helios2025} but adapted to capture the largest excursion in quality.

For a per-frame score sequence $\{s_t\}_{t=1}^T$ (either aesthetic or imaging) generated at frame rate $f$, we slide a window of \emph{fixed duration} $W$ along the timeline with stride $R$, i.e.\ a window of $w = \mathrm{round}(fW)$ frames with stride $r = \mathrm{round}(fR)$ frames. The mean score of each window is $\bar{s}_i = \frac{1}{|\mathcal{T}_i|}\sum_{t \in \mathcal{T}_i} s_t$, where $\mathcal{T}_i$ denotes the frame indices in window $i$. The last window is anchored to the end of the clip, $\mathcal{T}_{\text{last}} = \{\,t : T - w < t \le T\,\}$, so tail frames are never dropped, and a clip shorter than one window degenerates to a single window covering the whole rollout. We then take the highest and lowest window means, $\bar{s}_\mathrm{best} = \max_i \bar{s}_i$ and $\bar{s}_\mathrm{worst} = \min_i \bar{s}_i$, as the best- and worst-quality moments along the rollout. The drift score is the relative drop from the best to the worst window:
\begin{equation}
    D = \frac{\bar{s}_\mathrm{best} - \bar{s}_\mathrm{worst}}{|\bar{s}_\mathrm{best}|}.
    \label{eq:drift}
\end{equation}
Normalizing by the best-window score expresses drift relative to each rollout's quality level, facilitating comparison across models. We use $W = 3\,$s and $R = 1\,$s throughout, so $D = 0.05$ means the worst 3-second window scores 5\% below the best. \Cref{sec:appendix_drift_sweep} reports the parameter sweep.


Two design choices matter. First, windows use a fixed duration in \emph{seconds} rather than a fixed fraction of the clip. This maintains a consistent temporal scale across duration buckets and avoids averaging brief collapses over increasingly long windows as rollout duration grows. Second, \emph{overlapping} windows ($R < W$) reduce sensitivity to window boundaries, mitigating the dilution of collapses split between disjoint windows. Because $D$ is a range statistic, longer rollouts contain more windows and have a larger expected best-to-worst spread; drift should be compared within, not across, duration buckets. Applying this formulation to aesthetic and imaging scores yields lower-is-better metrics, \emph{Drift\textsubscript{Aesthetic}} ($D_\mathrm{aes}\!\downarrow$) and \emph{Drift\textsubscript{Imaging}} ($D_\mathrm{img}\!\downarrow$).

\subsubsection{Visual Score}
Absolute quality scores measure average perceptual quality, while drift scores capture within-rollout quality variation. The lower-is-better drift metrics are converted into stability scores and averaged with the absolute quality metrics:
\begin{equation}
    S_{\text{visual}} =
    \frac{
    S_\mathrm{aes}
    + (1 - D_\mathrm{aes})
    + S_\mathrm{img}
    + (1 - D_\mathrm{img})
    }{4}.
\end{equation}
This score rewards both high perceptual quality and temporal stability during long-horizon autoregressive rollout. Because the aesthetic and imaging scores are non-negative, $\bar{s}_\mathrm{best} \ge \bar{s}_\mathrm{worst} \ge 0$, so in practice $D \in [0, 1]$ and each stability term $1 - D$ stays within $[0, 1]$.

\subsection{Evaluation Dimension 3: Interaction Physics}
\label{sec:physics}

\ours{} evaluates interaction physics---i.e., the physical consequences of user-driven interaction---as \emph{physical plausibility under controlled generation}. A model should not only follow the prescribed action, but also produce physically consistent consequences as the user interacts with the environment. We decompose physics into three complementary domains: \emph{mechanics}, which measures whether objects, terrain, and the subject obey basic interaction constraints; \emph{optics}, which measures whether light-dependent phenomena such as reflections and shadows remain plausible; and \emph{3D consistency}, which measures whether the generated scene preserves rigid spatial structure during controlled camera motion. All VLM-based judgments in the physics suite use \vlmjudge{}.

\subsubsection{Validity Gate}
\label{sec:validity}

Before scoring physics, a validity gate removes rollouts without usable motion evidence: FPV requires pose-verified camera translation, while TPV requires continuous subject tracking (with a \vlmjudge{} direction check when masks are unavailable). Invalid cases are excluded (Appendix~\ref{sec:appendix_validity_details}).

\subsubsection{Physics Protocols}
\label{sec:physics_protocols}

Mechanics evaluates whether visible interactions produce physically plausible outcomes across four criteria: \emph{collision}, \emph{clipping}, \emph{deformation}, and \emph{gravity}, the last covering falls, sinking, and ground-contact adherence on slopes and steps. The two views draw on different evidence: FPV rollouts are judged from the reconstructed camera trajectory, testing whether the motion exhibits the environmental influence expected under the commanded action, while TPV rollouts are judged by an evidence-gathering agent that localizes the critical event window, inspects it with dense cropped samples, and commits to a verdict only after re-checking its own collected evidence (Appendix~\ref{sec:appendix_mechanics_details}). Unless otherwise specified, each mechanics protocol returns a binary pass/fail score.

Optics evaluates whether light-dependent effects remain consistent throughout interaction. We consider two protocols. \emph{Reflection} is scored with a coarse-to-fine procedure: the coarse stage checks global reflection plausibility, while the fine stage assigns per-frame reflection scores over valid frames and normalizes them to $[0,1]$. \emph{Occlusion}, implemented as shadow correctness, checks whether expected shadows are present and whether their direction and shape are consistent with the scene lighting.

3D consistency measures whether the scene preserves rigid spatial structure under controlled camera motion. We score it by comparing sampled generated frames against reference views captured from the source 3D engine within landmark-defined regions of interest (Appendix~\ref{sec:appendix_physics_protocols}).

\subsubsection{Physics Score}
\label{sec:physics_score}

Each valid case receives a protocol-level score $s_c \in [0,1]$. Domain scores are therefore conditional on the model-specific subsets accepted by the validity gate; TPV Control Rate provides a complementary model-level coverage indicator but does not make these subsets identical. Let $\mathcal{C}_{\mathrm{mech}}$, $\mathcal{C}_{\mathrm{opt}}$, and $\mathcal{C}_{\mathrm{3d}}$ denote the valid cases for mechanics, optics, and 3D consistency, respectively. The domain scores are
\begin{equation}
S_{\mathrm{mech}}
=
\frac{1}{|\mathcal{C}_{\mathrm{mech}}|}
\sum_{c\in\mathcal{C}_{\mathrm{mech}}} s_c,
\qquad
S_{\mathrm{opt}}
=
\frac{1}{|\mathcal{C}_{\mathrm{opt}}|}
\sum_{c\in\mathcal{C}_{\mathrm{opt}}} s_c,
\end{equation}
with $S_{\mathrm{3d}}$ defined analogously when 3D-consistency cases are available. The final physics score averages all available physics domains:
\begin{equation}
S_{\mathrm{physics}}
=
\frac{1}{|\mathcal{D}|}
\sum_{d\in\mathcal{D}} S_d,
\qquad
\mathcal{D}\subseteq\{\mathrm{mech},\mathrm{opt},\mathrm{3d}\}.
\end{equation}
Equal weighting is used because the three domains capture complementary aspects of physical plausibility: mechanics tests interaction-level consequences, optics tests light-level consistency, and 3D consistency tests spatial coherence.

\subsection{Evaluation Dimension 4: Memory}
\label{sec:memory}

\begin{figure}[t]
\centering
\includegraphics[width=\columnwidth]{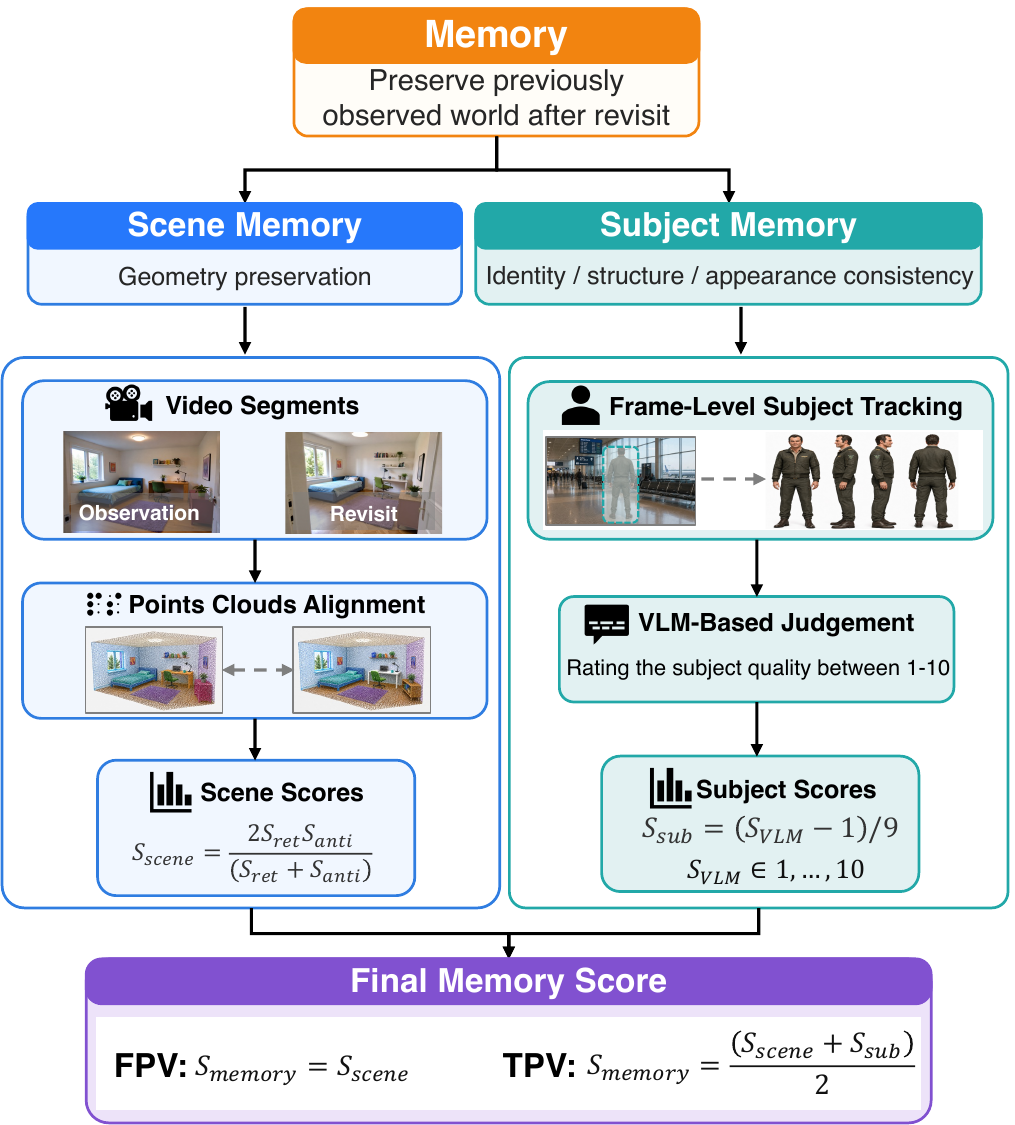}
\caption{\textbf{Memory evaluation pipeline.} \ours{} evaluates memory along two tracks: scene memory compares reconstructed geometry across the localized observation--revisit transition, while subject memory judges identity, structure, and appearance with a holistic \vlmjudge{}. Full details in Appendix~\ref{sec:appendix_memory_details}.}
\label{fig:memory_pipeline}
\end{figure}

\ours{} evaluates memory as \emph{revisit stability under imperfect control}. A model should preserve the previously observed world when the user returns to a region, even if the generated trajectory does not exactly match the ground-truth action path. As summarized in \Cref{fig:memory_pipeline}, we therefore separate memory into two complementary tracks. \emph{Scene memory} measures whether observed scene geometry is retained upon revisit, while \emph{subject memory} measures whether the protagonist in a third-person video preserves identity, structure, and appearance. This separation is important because scene memory is primarily geometric, whereas subject memory depends on fine-grained visual cues on a small moving region.

\subsubsection{Scene Memory}
\label{sec:scene_memory}

Scene memory asks whether the geometry observed before a revisit remains available after the model returns. Directly matching symmetric frame pairs is unreliable because current models under- or over-execute actions, accumulate timing offsets, and may not return to the same viewpoint. We therefore use a scene-level protocol rather than a frame-pair protocol. The rollout is split at the executed observation--revisit transition, both segments are reconstructed into point clouds, and the two clouds are aligned before scoring.

A reconstructed point cloud, however, is a \emph{discrete} sample of the scene rather than a continuous surface. Even two passes of the same trajectory over the same region do not yield point-wise coincident reconstructions: the achievable agreement between them is bounded below by the sampling density of the reconstruction itself. An absolute cross-segment distance is therefore not interpretable on its own, because the same numerical discrepancy means different things in a densely reconstructed indoor room and in a sparse distant landscape. We instead measure the cross-segment discrepancy \emph{relative to that bound}, so that the reported quantity is a signal-to-noise ratio rather than a distance.

Let $\mathcal{P}_{\mathrm{obs}}$ and $\mathcal{P}_{\mathrm{rev}}$ denote the aligned observation and revisit point clouds, and let $d(\mathbf{p}, \mathcal{Y})$ be the distance from a point $\mathbf{p}$ to its nearest neighbor in $\mathcal{Y}$. We estimate the sampling floor of a cloud $\mathcal{X}$ by splitting it into two disjoint random subsets $\mathcal{X}_1, \mathcal{X}_2$ and measuring how far apart one half sits from the other,
\begin{equation}
\sigma(\mathcal{X}) = \frac{1}{|\mathcal{X}_1|}\sum_{\mathbf{p}\in\mathcal{X}_1} d(\mathbf{p},\, \mathcal{X}_2),
\label{eq:sampling_floor}
\end{equation}
and we write the directed cross-segment discrepancy as
\begin{equation}
\delta(\mathcal{X}\!\rightarrow\!\mathcal{Y}) = \frac{1}{|\mathcal{X}|}\sum_{\mathbf{p}\in\mathcal{X}} d(\mathbf{p},\, \mathcal{Y}).
\label{eq:cross_discrepancy}
\end{equation}
Retention and anti-hallucination are then the two directed ratios
\begin{equation}
\begin{aligned}
S_{\mathrm{ret}} &= \min\!\left(1,\; \frac{\sigma(\mathcal{P}_{\mathrm{obs}})}{\delta(\mathcal{P}_{\mathrm{obs}}\!\rightarrow\!\mathcal{P}_{\mathrm{rev}})}\right), \\[5pt]
S_{\mathrm{anti}} &= \min\!\left(1,\; \frac{\sigma(\mathcal{P}_{\mathrm{rev}})}{\delta(\mathcal{P}_{\mathrm{rev}}\!\rightarrow\!\mathcal{P}_{\mathrm{obs}})}\right),
\end{aligned}
\label{eq:mem_ratios}
\end{equation}
and the scene-memory score combines them as
\begin{equation}
S_{\mathrm{scene}} = \frac{2\,S_{\mathrm{ret}}\,S_{\mathrm{anti}}}{S_{\mathrm{ret}} + S_{\mathrm{anti}}}.
\label{eq:scene_memory}
\end{equation}

These two directed scores play roles analogous to recall and precision. Treating $\mathcal{P}_{\mathrm{obs}}$ as the geometry that ought to be reproduced and $\mathcal{P}_{\mathrm{rev}}$ as what the model actually produces, $S_{\mathrm{ret}}$ measures how closely the observed geometry is recovered on revisit and therefore plays the role of \emph{recall}, whereas $S_{\mathrm{anti}}$ measures how consistently the revisit geometry is supported by something previously observed and therefore plays the role of \emph{precision}. The two capture different forms of geometric inconsistency: missing previously observed structure can lower retention, whereas additional unsupported structure can lower anti-hallucination. Because $\delta$ is directed, a single symmetric Chamfer distance would average the two directions and conceal exactly this asymmetry; we therefore combine them with the harmonic mean of \cref{eq:scene_memory}, which is high only when \emph{both} terms are high and thus requires a model to avoid forgetting and hallucination simultaneously.

The clamp in \cref{eq:mem_ratios} should be read as saturation rather than truncation: $S_{\mathrm{ret}}=1$ does not assert that the two clouds are identical, only that their discrepancy has fallen below the resolution at which our reconstruction can distinguish them. Empirically $20.2\%$ of first-person cases saturate, i.e.\ they reach this floor. Appendix~\ref{sec:appendix_scene_memory_details} gives the full transition-localization, point-cloud filtering, registration, and scoring details, together with additional implementation notes; Appendix~\ref{sec:frame_pair_memory} analyzes why frame-pair alternatives remain unreliable.

\subsubsection{Subject Memory}
\label{sec:subject_memory}

Subject memory captures a complementary failure mode in TPV rollouts: the protagonist may preserve the scene trajectory while drifting in identity, body structure, or appearance. Because the subject is a small, articulated region and cannot be reliably evaluated by scene-level point clouds, we use a tracking-and-VLM protocol instantiated with \vlmjudge{}. We first apply a controllability gate that removes cases with insufficient subject-tracking continuity. Crucially, this gate is defined on tracking \emph{continuity} rather than on displacement from the initial pose: a correctly controlled protagonist is expected to walk away from where it started, so we compare each sampled frame against the previous sampled frame, and penalize only discontinuities such as disappearance, teleportation, tracking loss, and abrupt scale change. For the cases that pass, SAM\,2 propagates an annotated first-frame subject mask through the rollout, each sampled frame is reduced to a background-suppressed crop of the subject, and the entire temporally ordered crop sequence is submitted to the \vlmjudge{} in a single holistic judgment rather than as pairwise comparisons. Its rubric explicitly declares viewpoint and illumination change benign while penalizing identity change, structural collapse, appearance rewriting, disappearance, and progressive quality degradation. The judge produces an integer subject memory score $S_{\mathrm{VLM}}\in\{1,\ldots,10\}$, which we normalize as
\begin{equation}
S_{\mathrm{sub}} = \frac{S_{\mathrm{VLM}} - 1}{9}.
\end{equation}
This holistic score avoids over-penalizing benign viewpoint or lighting changes while still detecting identity changes, structural collapse, disappearance, and appearance drift. Appendix~\ref{sec:appendix_subject_memory_details} describes the controllability gate, tracking procedure, failure taxonomy, normalization, and prompt.

\subsubsection{Memory Score}
\label{sec:memory_total_score}

The final memory score depends on the evaluated view. For FPV, there is no visible protagonist, so memory is the scene-memory score:
\begin{equation}
S_{\mathrm{memory}}^{\mathrm{FPV}} = S_{\mathrm{scene}}.
\end{equation}
For TPV, memory combines scene preservation and protagonist preservation:
\begin{equation}
S_{\mathrm{memory}}^{\mathrm{TPV}}
=
\frac{1}{2}\left(S_{\mathrm{scene}} + S_{\mathrm{sub}}\right).
\end{equation}
This formulation keeps the geometric revisit criterion used in FPV while additionally requiring TPV models to maintain a stable protagonist. Both TPV terms are averaged only over the cases that pass the controllability gate, and we report the overall TPV control rate separately (\cref{tab:tpv_control_rate}).

\subsection{Scoring and Leaderboard}
\label{sec:scoring}

The overall \ours{} score is a weighted composite:
\begin{equation}
\begin{aligned}
    S_{\text{ROAM}} ={}& w_1 \cdot S_{\text{action}} + w_2 \cdot S_{\text{visual}}\\
    &+ w_3 \cdot S_{\text{physics}} + w_4 \cdot S_{\text{memory}} .
\end{aligned}
\end{equation}
Results are reported with equal default weights $w_1{=}w_2{=}w_3{=}w_4{=}0.25$, together with per-dimension rankings.
For TPV, the leaderboard overall score additionally accounts for whether a model can maintain a continuously trackable third-person subject:
\begin{equation}
    S_{\text{ROAM}}^{\mathrm{TPV}} = R_{\text{ctrl}} \cdot S_{\text{ROAM}},
\end{equation}
where $R_{\text{ctrl}}$ is the fraction of evaluated TPV cases passing the controllability gate (Appendix~\ref{sec:appendix_validity_details}). We report $R_{\text{ctrl}}$ separately so that the dimension scores can be interpreted alongside the controllability gate.

\section{Test Suite}
\label{sec:testsuite}

The \ours{} test suite provides first-frame images and keystroke action sequences for all four evaluation dimensions.
Representative cases from the complete test suite are displayed in Appendix~\ref{sec:appendix_gallery}.
We use a unified construction methodology while tailoring specific cases to each dimension's requirements.
The image suite and action suite are described separately to clarify both shared design principles and dimension-specific choices.

\subsection{Image Suite}
\label{sec:image_suite}

\paragraph{Unified construction pipeline.}
First-frame images spanning first-person and third-person views are collected from two complementary sources:
(1)~the open Internet, providing diverse real-world scenes, and
(2)~existing world model benchmarks such as WorldScore~\cite{worldscore2025}, providing standardized evaluation imagery.
These raw images are then refined with GPT-Image-2 through targeted edits: improving resolution and clarity, adjusting object placement and layout, converting between first-person and third-person views, and captioning followed by regeneration to produce visually similar but distinct variants.
This refinement pipeline serves two purposes: it ensures high image quality and broad scene diversity, while reducing the likelihood that the resulting first frames overlap with images seen during world model training, thereby mitigating potential data contamination that could inflate benchmark scores.

\paragraph{Dimension-specific image selection.}
Although all dimensions draw from the same image library, the selection criteria differ according to each dimension's evaluation goal.
For the \emph{action dimension}, spacious, unobstructed first-frame images (\emph{e.g.}, open fields, wide corridors, empty plazas) are prioritized to facilitate translation and rotation with fewer obstacle-induced confounds. This supports a cleaner measurement of per-keystroke accuracy.
For the \emph{physics dimension}, scenes contain specific physical affordances: environments with walls, pillars,
or furniture for clipping testing; elevated platforms or staircases for terrain-following evaluation; and long unobstructed corridors for 3D consistency measurement.
For the \emph{memory dimension}, images span a spectrum of scene complexity, from simple scenes with few distinctive landmarks (\emph{e.g.}, a single building on an open plain) to complex scenes with rich geometric detail (\emph{e.g.}, a cluttered indoor room or a dense urban street). This range is paired with both simple and multi-step action sequences to evaluate memory across different revisit difficulties.

\subsection{Action Suite}
\label{sec:action_suite}

\paragraph{Dimension-specific action design.}
Action sequences are designed with varying horizons and tailored to each dimension. Across the benchmark we use one horizon definition based on commanded rollout duration: \emph{Short} ($t\leq 15$\,s), \emph{Mid} ($15<t\leq 30$\,s), and \emph{Long} ($30<t\leq60$\,s).
For the \emph{action dimension}, we cross three rollout durations (15/30/60\,s) with 0/1/2 action switches in a balanced $3\!\times\!3$ design, holding a single key within each segment. This supports separate analyses of rollout duration and action switching. Sequences are adapted to scene layout to reduce collision-related confounds. When a sequence combines translation and rotation, translation precedes rotation, avoiding subsequent movement along model-dependent post-rotation headings. \Cref{sec:appendix_action_design} details the construction.
For the \emph{memory dimension}, sequences follow symmetric traversal patterns such as A$\rightarrow$B$\rightarrow\mathrm{inv}$(B)$\rightarrow\mathrm{inv}$(A), where $\mathrm{inv}(\cdot)$ denotes the inverse action (\emph{e.g.}, forward--backward, left--right loops), to enable revisit evaluation. As with the action dimension, paths are intentionally routed through open space to minimize collisions or other physics-driven phenomena, reducing the influence of physics-induced trajectory deviations on revisit comparisons.
For the \emph{physics dimension}, sequences are crafted to \emph{trigger} physical interactions: paths are directed toward walls (collision), along edges of elevated surfaces (gravity), and through long corridors at a constant input rate (3D consistency). Each sequence targets a specific physical phenomenon for focused evaluation.

\paragraph{Avoiding confounding physics in non-physics dimensions.}
A key cross-cutting design principle is that action sequences in the action and memory suites should avoid physical phenomena (\emph{e.g.}, collisions, gravitational effects) that could confound the target evaluation.
For instance, if a forward-walking sequence causes the model to collide with a wall, the resulting deceleration would be incorrectly penalized as an action-following error.
Combining spacious first-frame environments (\cref{sec:image_suite}) with obstacle-free paths helps reduce physics-related confounds in action and memory evaluation.

\subsection{Test Suite Statistics}
\label{sec:testsuite_stats}

\begin{figure}[t]
\centering
\includegraphics[width=\columnwidth]{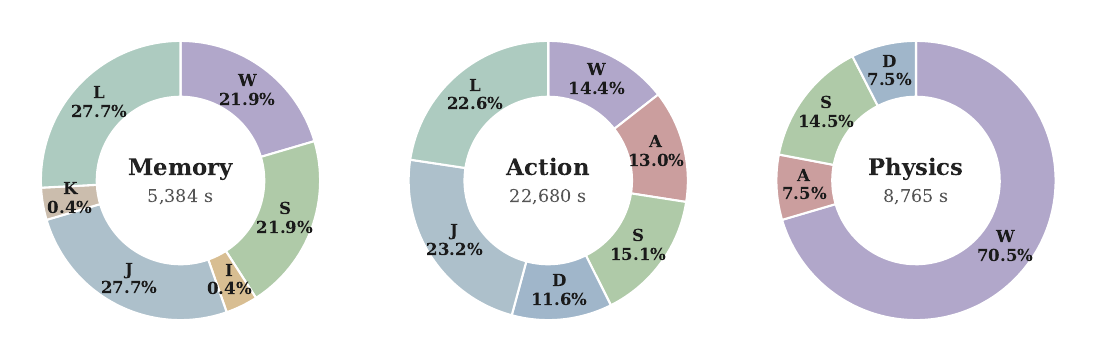}
\caption{\textbf{Action key distribution} across the three evaluation dimensions. Each donut shows the proportion and total duration (s) of commanded keys over all test cases in that dimension.}
\label{fig:action_key_dist}
\end{figure}

\Cref{fig:action_key_dist} shows the action-key distribution across the three evaluation dimensions, measured directly from the commanded duration in seconds. We use \emph{key-seconds}: holding one key for one second contributes one key-second, and simultaneous keys each accumulate their own duration.
Memory is dominated by two symmetric pairs: J/L yaw commands each account for 27.7\% of commanded key-seconds, while W/S translation commands each account for 21.9\%, reflecting the revisit-oriented design in which every outbound command must have an exact inverse.
Action covers six keys and is deliberately the most balanced dimension, led by J (23.2\%) and L (22.6\%), followed by S (15.1\%), W (14.4\%), A (13.0\%), and D (11.6\%), so that no single motion axis dominates the per-keystroke accuracy measurement.
Physics is strongly skewed toward W (70.5\%), as most physics test cases involve sustained forward motion toward obstacles, terrain changes, or long corridors, with smaller S (14.5\%) and A/D (7.5\% each) components used for backward or lateral interactions.

\begin{figure}[t]
\centering
\includegraphics[width=\columnwidth]{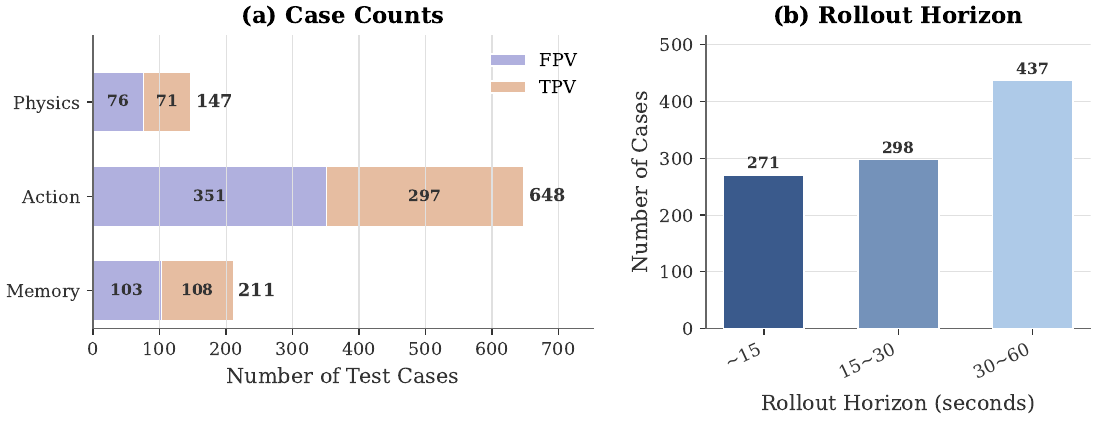}
\caption{\textbf{Test case count and rollout-horizon distribution.} (a) Test cases by dimension, split by FPV and TPV. (b) Distribution across commanded rollout horizons, measured in seconds.}
\label{fig:case_counts}
\end{figure}

\begin{figure}[!t]
\centering
\includegraphics[width=\columnwidth]{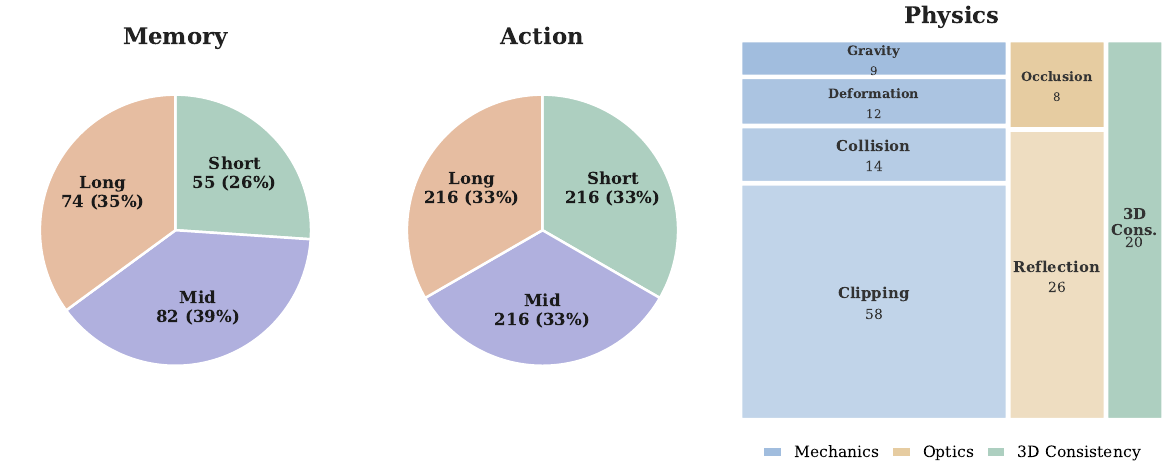}
\caption{\textbf{Rollout-horizon and physics-protocol distribution.} Memory and Action are shown over Short, Mid, and Long horizon splits, using the exact evaluation counts. The right subfigure reports all 147 physics cases by protocol.}
\label{fig:horizon_dist}
\end{figure}

\begin{table*}[!t]
\centering
\small
\caption[Evaluated interactive world models.]{Evaluated interactive world models. In the \textit{Input} column, color-coded icons denote input modalities: \Act\,=\,Action (keyboard WASD/IJKL), \Pose\,=\,Pose (continuous 6-DoF camera trajectory), and \Txt\,=\,Text (caption/instruction). The arrow ($\to$) shows the conversion applied to the keyboard actions---e.g., \Act$\to$\Pose means keyboard actions are converted to camera poses before being fed to the model. Models with only \Act accept keyboard input natively with no conversion. \textit{FPS} reports the native output frame rate of each model. \textit{Chunk (frames)} is the number of frames produced per autoregressive chunk, and \textit{Inf. Speed (Hz)} is the average generation throughput computed as chunk frames divided by wall-clock chunk time, measured on a single NVIDIA~H20 GPU and averaged over 20 runs. \textit{MPPS} (million pixels generated per second) is computed as resolution width $\times$ height $\times$ inference speed divided by $10^6$. SANA-WM is a bidirectional model, so we evaluate it with 56 frames per chunk.}
\label{tab:models}
\setlength{\tabcolsep}{4.5pt}%
\resizebox{\textwidth}{!}{%
\begin{tabular}{lcccccccccc l}
\toprule
\textbf{Model} & \textbf{Type} & \textbf{Params} & \textbf{Year} & \textbf{Views} & \textbf{Input} & \textbf{Resolution} & \textbf{FPS} & \textbf{Chunk (frames)} & \textbf{Inf. Speed (Hz)} & \textbf{MPPS} & \textbf{Source} \\
\midrule
\cellcolor{closedColB} Genie~3~\cite{genie3_2025} & \cellcolor{closedColB} Closed & \cellcolor{closedColB} --- & \cellcolor{closedColB} 2025 & \cellcolor{closedColB} FPV/TPV & \cellcolor{closedColB} \Act & \cellcolor{closedColB} 1280$\times$704 & \cellcolor{closedColB} 20 & \cellcolor{closedColB} --- & \cellcolor{closedColB} --- & \cellcolor{closedColB} --- & \cellcolor{closedColB} \makebox[1.5em][c]{\raisebox{-0.2\height}{\includegraphics[height=1em]{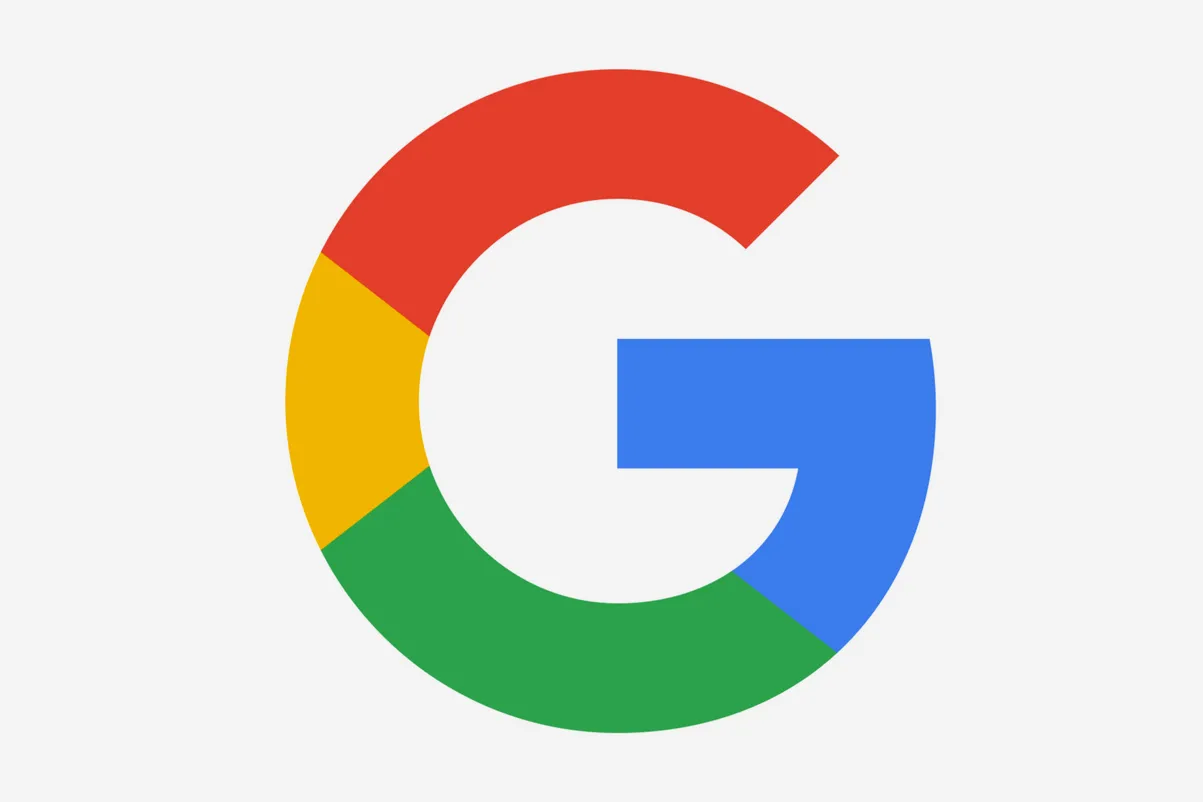}}}\,DeepMind \\[-1pt]
\cellcolor{closedColB} Happy Oyster~\cite{happyoyster2026} & \cellcolor{closedColB} Closed & \cellcolor{closedColB} --- & \cellcolor{closedColB} 2026 & \cellcolor{closedColB} FPV/TPV & \cellcolor{closedColB} \Act & \cellcolor{closedColB} 1296$\times$720 & \cellcolor{closedColB} 24 & \cellcolor{closedColB} --- & \cellcolor{closedColB} --- & \cellcolor{closedColB} --- & \cellcolor{closedColB} \makebox[1.5em][c]{\raisebox{-0.2\height}{\includegraphics[height=1em]{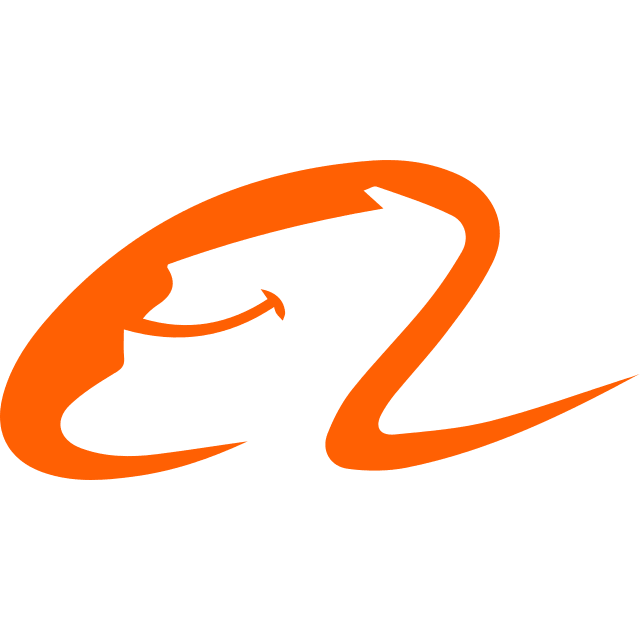}}}\,Alibaba \\
\midrule
LingBot-World~\cite{lingbotworld} & Open & 14B & 2026 & FPV/TPV & \Act$\to$\Pose & 832$\times$480 & 16 & 12 & 1.79 & 0.72 & \makebox[1.5em][c]{\raisebox{-0.2\height}{\includegraphics[height=1em]{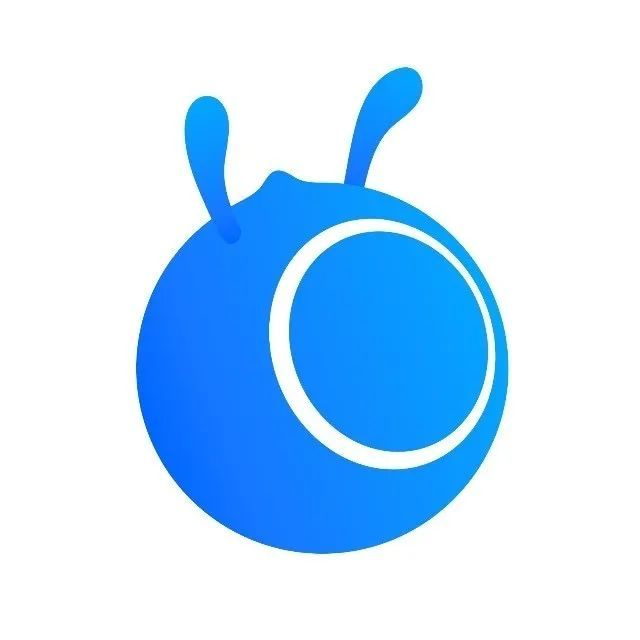}}}\,Ant Group \\
\rowcolor{metricSub} LingBot-World~2.0~\cite{lingbot-world-v2} & Open & 14B & 2026 & FPV/TPV & \Act$\to$\Pose & 832$\times$480 & 16 & 16 & 1.75 & 0.70 & \makebox[1.5em][c]{\raisebox{-0.2\height}{\includegraphics[height=1em]{figures/logos/ant.png}}}\,Ant Group \\
HY-World~1.5~\cite{hyworld15} & Open & 8B & 2025 & FPV/TPV & \Act$\to$\Act{+}\Pose & 832$\times$480 & 24 & 16 & 1.30 & 0.52 & \makebox[1.5em][c]{\raisebox{-0.2\height}{\includegraphics[height=1em]{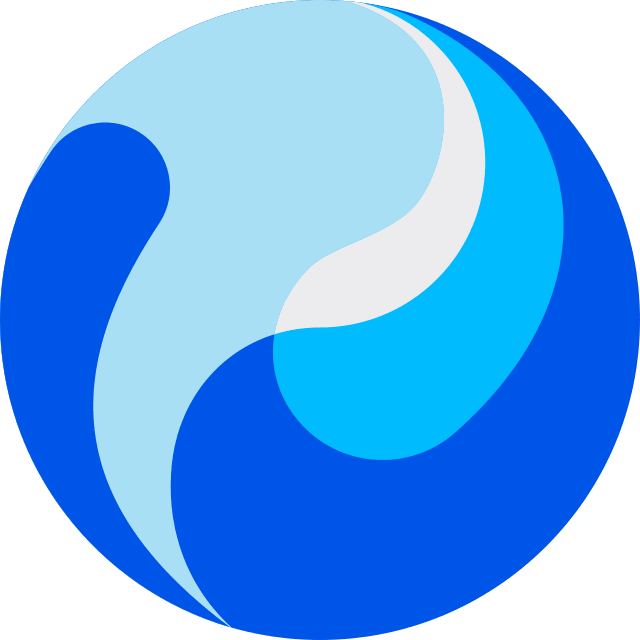}}}\,Tencent \\
\rowcolor{metricSub} Lyra~2.0~\cite{lyra2_2026} & Open & 14B & 2026 & FPV & \Act$\to$\Pose & 832$\times$480 & 16 & 80 & 2.31 & 0.92 & \makebox[1.5em][c]{\raisebox{-0.2\height}{\includegraphics[height=1em]{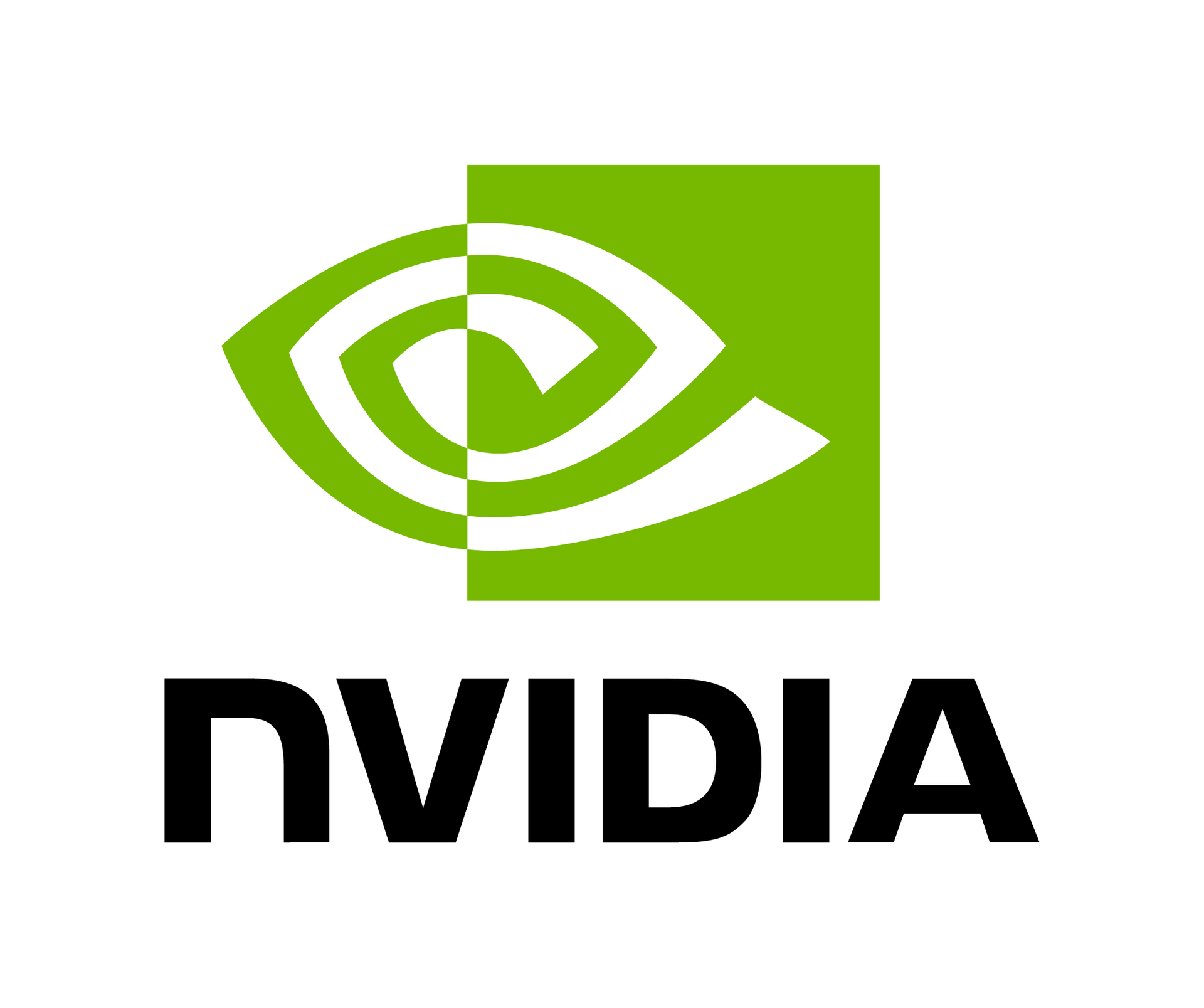}}}\,NVIDIA \\
SANA-WM~\cite{sanawm} & Open & 2.6B & 2026 & FPV & \Act$\to$\Pose & 1280$\times$704 & 16 & 56 & 0.81 & 0.73 & \makebox[1.5em][c]{\raisebox{-0.2\height}{\includegraphics[height=1em]{figures/logos/nvidia.png}}}\,NVIDIA \\
\rowcolor{metricSub} SANA-WM-AR~\cite{sanawm} & Open & 2.6B & 2026 & FPV & \Act$\to$\Pose & 1280$\times$704 & 16 & 24 & 0.93 & 0.84 & \makebox[1.5em][c]{\raisebox{-0.2\height}{\includegraphics[height=1em]{figures/logos/nvidia.png}}}\,NVIDIA \\
Matrix-Game~3.0~\cite{matrixgame3} & Open & 5B & 2026 & FPV & \Act$\to$\Act{+}\Pose & 1280$\times$704 & 17 & 40 & 5.34 & 4.82 & \makebox[1.5em][c]{\raisebox{-0.2\height}{\includegraphics[height=1em]{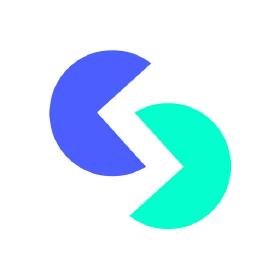}}}\,Skywork \\
\rowcolor{metricSub} Matrix-Game~2.0~\cite{matrixgame2} & Open & 1.8B & 2025 & FPV & \Act & 640$\times$352 & 12 & 12 & 6.26 & 1.41 & \makebox[1.5em][c]{\raisebox{-0.2\height}{\includegraphics[height=1em]{figures/logos/matrix.png}}}\,Skywork \\
Yume~1.5~\cite{yume15} & Open & 5B & 2025 & FPV & \Act$\to$\Txt & 1280$\times$704 & 16 & 32 & 6.49 & 5.85 & \makebox[1.5em][c]{\raisebox{-0.2\height}{\includegraphics[height=1em]{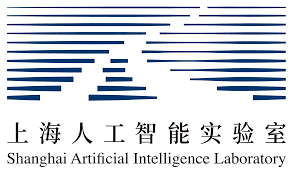}}}\,Shanghai AI Lab \\
\rowcolor{metricSub} minWM~\cite{minwm2026} & Open & 8B & 2026 & FPV & \Act$\to$\Pose & 832$\times$480 & 16 & 16 & 3.97 & 1.58 & \makebox[1.5em][c]{\raisebox{-0.2\height}{\includegraphics[height=1em]{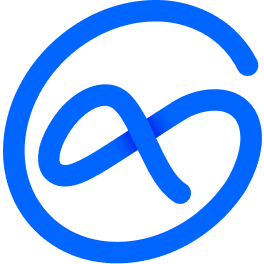}}}\,ShengShu \\
\bottomrule
\end{tabular}%
}
\end{table*}

\begin{table*}[!t]
\centering
\caption{\textbf{Unified quantitative results on \ours{}.} All values $\times 100$. Metrics are rows and models are columns, organized into an FPV table, a TPV table, and an Overall-score ranking leaderboard. Gray columns denote closed-source models. \textbf{Bold}/\underline{underline}\,=\,best/2nd-best within each metric. $\uparrow$\,=\,higher is better; $\downarrow$\,=\,lower is better. For both FPV and TPV, Visual Score\,=\,mean(Aesthetic, $1-$Aesthetic Drift, Imaging Quality, $1-$Imaging Drift) and Overall Score\,=\,mean(Action, Visual, Physics, Memory). The view-specific terms are Action Score\,=\,mean(Strict Acc, Partial Acc, Traj), Physics Score\,=\,mean(Mech, Opt, 3D) and Memory Score\,=\,$F_1$ for FPV; Action Score\,=\,mean(Partial Acc, Traj), Physics Score\,=\,mean(Mech, Opt) (no 3D data) and Memory Score\,=\,mean($F_1$, Subject) for TPV. TPV Overall Score is further weighted by Control Rate.}
\label{tab:main_results}
\setlength{\tabcolsep}{3.2pt}
\scriptsize
\renewcommand{\arraystretch}{0.9}
\textbf{(a) First-person View (FPV)}\\[0.12em]
\resizebox{\textwidth}{!}{%
\begin{tabular}{l|*{2}{>{\columncolor{closedCol}}c}cccccccccc}
\toprule
\textbf{Metric}
& \modelhead{figures/logos/google.png}{Genie 3}
& \modelhead{figures/logos/alibaba.png}{Happy Oyster}
& \modelhead{figures/logos/nvidia.png}{Lyra 2.0}
& \modelhead{figures/logos/hy.png}{HY-World 1.5}
& \modelhead{figures/logos/ant.png}{LingBot-World~2.0}
& \modelhead{figures/logos/matrix.png}{Matrix-Game 3.0}
& \modelhead{figures/logos/nvidia.png}{SANA-WM-AR}
& \modelhead{figures/logos/ant.png}{LingBot-World}
& \modelhead{figures/logos/nvidia.png}{SANA-WM}
& \modelhead{figures/logos/matrix.png}{Matrix-Game 2.0}
& \modelhead{figures/logos/yume.png}{Yume 1.5}
& \modelhead{figures/logos/shengshu.png}{minWM} \\
\midrule
\rowcolor{metricTotal}
\textbf{Overall Score} $\uparrow$ & \textbf{70.32} & 68.53 & \underline{69.61} & 63.97 & 63.90 & 60.34 & 59.37 & 58.77 & 54.61 & 51.92 & 47.06 & 41.88 \\
\midrule
\rowcolor{metricAction}
\textbf{Action Score} $\uparrow$ & 76.31 & 88.22 & \underline{90.82} & 87.38 & 90.74 & \textbf{91.34} & 75.73 & 90.78 & 70.42 & 89.67 & 71.77 & 60.79 \\
\rowcolor{metricAction}
Strict Accuracy $\uparrow$ & 62.04 & 85.90 & \underline{90.85} & 83.21 & 88.95 & 88.77 & 71.56 & \textbf{91.17} & 61.91 & 88.11 & 58.45 & 46.15 \\
\rowcolor{metricAction}
Partial Accuracy $\uparrow$ & 78.06 & 89.96 & 93.82 & 88.49 & \textbf{96.30} & \underline{94.45} & 76.66 & 94.38 & 69.77 & 92.79 & 76.41 & 53.18 \\
\rowcolor{metricAction}
Trajectory $\uparrow$ & 88.83 & 88.80 & 87.78 & \underline{90.44} & 86.98 & \textbf{90.79} & 78.96 & 86.80 & 79.59 & 88.10 & 80.45 & 83.04 \\
\rowcolor{metricAction}
nATE$_t$ $\downarrow$ & 11.48 & 11.02 & 9.61 & \underline{8.63} & \textbf{6.85} & 9.24 & 23.23 & 9.92 & 22.60 & 11.57 & 18.91 & 13.63 \\
\rowcolor{metricAction}
nATE$_r$ $\downarrow$ & 10.85 & 11.37 & 14.83 & \underline{10.49} & 19.18 & \textbf{9.17} & 18.86 & 16.47 & 18.22 & 12.23 & 20.18 & 20.28 \\
\midrule
\rowcolor{metricVisual}
\textbf{Visual Score} $\uparrow$ & 66.21 & 66.49 & 69.61 & 66.06 & 68.55 & 68.70 & \textbf{74.40} & 69.28 & \underline{71.42} & 60.83 & 69.07 & 58.21 \\
\rowcolor{metricVisual}
Aesthetic Score $\uparrow$ & 53.41 & 55.93 & 59.62 & 58.22 & 59.93 & 56.50 & \underline{60.08} & \textbf{60.81} & 57.17 & 53.69 & 57.97 & 48.93 \\
\rowcolor{metricVisual}
Aesthetic Drift $\downarrow$ & \textbf{17.22} & 19.15 & 22.63 & 27.06 & 24.89 & 26.25 & \underline{18.36} & 21.20 & 21.83 & 32.43 & 23.91 & 43.50 \\
\rowcolor{metricVisual}
Imaging Quality $\uparrow$ & 54.53 & 49.04 & 60.82 & 55.11 & 61.57 & 64.77 & \textbf{69.08} & 59.96 & \underline{67.80} & 48.52 & 64.64 & 51.59 \\
\rowcolor{metricVisual}
Imaging Drift $\downarrow$ & 25.87 & 19.87 & 19.36 & 22.04 & 22.43 & 20.24 & \textbf{13.20} & 22.47 & \underline{17.48} & 26.46 & 22.42 & 24.20 \\
\midrule
\rowcolor{metricPhysics}
\textbf{Physics Score} $\uparrow$ & \textbf{66.67} & \underline{61.40} & 51.52 & 40.43 & 43.52 & 25.56 & 22.88 & 31.08 & 17.38 & 12.21 & 14.33 & 7.12 \\
\rowcolor{metricPhysics}
Mechanics $\uparrow$ & \textbf{61.50} & \underline{45.50} & 4.80 & 0.00 & 3.80 & 0.00 & 6.40 & 3.80 & 3.20 & 0.00 & 7.40 & 0.00 \\
\rowcolor{metricPhysics}
Optics $\uparrow$ & 60.60 & \underline{67.50} & \textbf{69.10} & 48.80 & 56.60 & 17.20 & 41.20 & 51.20 & 44.70 & 9.70 & 10.30 & 4.40 \\
\rowcolor{metricPhysics}
3D Consistency $\uparrow$ & \underline{77.90} & 71.21 & \textbf{80.65} & 72.50 & 70.17 & 59.49 & 21.05 & 38.25 & 4.23 & 26.92 & 25.30 & 16.97 \\
\midrule
\rowcolor{metricMemory}
\textbf{Memory Score} $\uparrow$ & \textbf{72.09} & 58.00 & \underline{66.49} & 62.02 & 52.80 & 55.76 & 64.48 & 43.93 & 59.23 & 44.96 & 33.05 & 41.41 \\
\rowcolor{metricMemory}
Retention $\uparrow$ & \textbf{73.93} & 55.11 & 62.25 & 59.34 & 53.12 & 59.24 & \underline{62.79} & 38.77 & 60.09 & 45.29 & 30.67 & 41.55 \\
\rowcolor{metricMemory}
Anti-Hallucination $\uparrow$ & \underline{77.44} & 71.29 & \textbf{79.45} & 71.26 & 60.01 & 59.78 & \underline{77.44} & 62.16 & 72.51 & 54.66 & 48.72 & 59.34 \\
\bottomrule
\end{tabular}%
}

\vspace{0.62em}
\noindent
\begin{minipage}[t]{0.47\textwidth}
\vspace{0pt}
\centering
\textbf{(b) Third-person View (TPV)}\\[0.14em]
\resizebox{\linewidth}{!}{%
\begin{tabular}{l|*{2}{>{\columncolor{closedCol}}c}ccc}
\toprule
\textbf{Metric}
& \modelhead{figures/logos/alibaba.png}{Happy Oyster}
& \modelhead{figures/logos/google.png}{Genie 3}
& \modelhead{figures/logos/ant.png}{LingBot-World~2.0}
& \modelhead{figures/logos/hy.png}{HY-World 1.5}
& \modelhead{figures/logos/ant.png}{LingBot-World} \\
\midrule
\rowcolor{metricTotal}
\textbf{Overall Score} $\uparrow$ & \textbf{70.66} & \underline{61.86} & 38.31 & 28.07 & 23.73 \\
\rowcolor{metricTotal}
Control Rate $\uparrow$ & \textbf{96.43} & \underline{83.82} & 53.78 & 39.71 & 35.50 \\
\midrule
\rowcolor{metricAction}
\textbf{Action Score} $\uparrow$ & 84.81 & 84.76 & \underline{88.61} & \textbf{91.46} & 82.29 \\
\rowcolor{metricAction}
Partial Accuracy $\uparrow$ & 81.16 & 79.83 & \underline{90.29} & \textbf{91.88} & 81.28 \\
\rowcolor{metricAction}
Trajectory $\uparrow$ & 88.45 & \underline{89.68} & 86.92 & \textbf{91.03} & 83.30 \\
\rowcolor{metricAction}
nATE$_t$ $\downarrow$ & 13.39 & \underline{10.78} & 12.57 & \textbf{9.14} & 19.55 \\
\rowcolor{metricAction}
nATE$_r$ $\downarrow$ & \underline{9.71} & 9.86 & 13.60 & \textbf{8.80} & 13.85 \\
\midrule
\rowcolor{metricVisual}
\textbf{Visual Score} $\uparrow$ & 69.84 & 72.03 & \textbf{74.46} & 71.52 & \underline{73.47} \\
\rowcolor{metricVisual}
Aesthetic Score $\uparrow$ & 61.18 & 60.65 & \textbf{64.62} & \underline{63.25} & 62.25 \\
\rowcolor{metricVisual}
Aesthetic Drift $\downarrow$ & \underline{14.45} & \textbf{12.21} & 15.38 & 17.83 & 14.49 \\
\rowcolor{metricVisual}
Imaging Quality $\uparrow$ & 49.10 & 57.60 & \textbf{62.96} & 56.29 & \underline{62.37} \\
\rowcolor{metricVisual}
Imaging Drift $\downarrow$ & 16.47 & 17.92 & \textbf{14.35} & \underline{15.65} & 16.27 \\
\midrule
\rowcolor{metricPhysics}
\textbf{Physics Score} $\uparrow$ & \textbf{68.90} & \underline{67.15} & 58.00 & 51.35 & 53.85 \\
\rowcolor{metricPhysics}
Mechanics $\uparrow$ & \underline{52.10} & \textbf{70.70} & 41.00 & 41.20 & 35.00 \\
\rowcolor{metricPhysics}
Optics $\uparrow$ & \textbf{85.70} & 63.60 & \underline{75.00} & 61.50 & 72.70 \\
\midrule
\rowcolor{metricMemory}
\textbf{Memory Score} $\uparrow$ & \underline{69.55} & \textbf{71.27} & 63.88 & 68.41 & 57.75 \\
\rowcolor{metricMemory}
Retention $\uparrow$ & 59.76 & \underline{66.56} & 59.46 & \textbf{72.86} & 49.52 \\
\rowcolor{metricMemory}
Subject Memory $\uparrow$ & \textbf{81.09} & \underline{77.99} & 73.88 & 67.52 & 62.22 \\
\rowcolor{metricMemory}
Anti-Hallucination $\uparrow$ & 69.14 & \underline{72.19} & 58.17 & \textbf{74.90} & 70.02 \\
\bottomrule
\end{tabular}%
}
\end{minipage}\hfill
\begin{minipage}[t]{0.50\textwidth}
\vspace{0pt}
\centering
\textbf{(c) Overall-score Ranking}\\[0.14em]
\resizebox{\linewidth}{!}{%
\begin{tikzpicture}[x=1cm,y=1cm,font=\scriptsize]
  \node[anchor=west,font=\bfseries] at (0,6.24) {FPV};
  \node[anchor=west,text=black!60] at (0.55,6.24) {medals mark top-3};

  \node[circle,fill=medalgold,text=white,inner sep=0pt,minimum size=0.24cm,font=\tiny\bfseries] at (0.16,5.86) {1};
  \node[anchor=west] at (0.38,5.86) {Genie 3};
  \fill[barblue] (2.85,5.78) rectangle (6.12,5.94);
  \node[anchor=west] at (6.65,5.86) {70.32};

  \node[circle,fill=medalsilver,text=white,inner sep=0pt,minimum size=0.24cm,font=\tiny\bfseries] at (0.16,5.60) {2};
  \node[anchor=west] at (0.38,5.60) {Lyra 2.0};
  \fill[bargold] (2.85,5.52) rectangle (6.08,5.68);
  \node[anchor=west] at (6.65,5.60) {69.61};

  \node[circle,fill=medalbronze,text=white,inner sep=0pt,minimum size=0.24cm,font=\tiny\bfseries] at (0.16,5.34) {3};
  \node[anchor=west] at (0.38,5.34) {Happy Oyster};
  \fill[bargreen] (2.85,5.26) rectangle (6.03,5.42);
  \node[anchor=west] at (6.65,5.34) {68.53};

  \node[anchor=center,text=black!60,font=\tiny\bfseries] at (0.16,5.08) {4};
  \node[anchor=west] at (0.38,5.08) {HY-World 1.5};
  \fill[barmauve] (2.85,5.00) rectangle (5.82,5.16);
  \node[anchor=west] at (6.65,5.08) {63.97};

  \node[anchor=center,text=black!60,font=\tiny\bfseries] at (0.16,4.82) {5};
  \node[anchor=west] at (0.38,4.82) {LingBot-World~2.0};
  \fill[barslate] (2.85,4.74) rectangle (5.82,4.90);
  \node[anchor=west] at (6.65,4.82) {63.90};

  \node[anchor=center,text=black!60,font=\tiny\bfseries] at (0.16,4.56) {6};
  \node[anchor=west] at (0.38,4.56) {Matrix-Game 3.0};
  \fill[barrose] (2.85,4.48) rectangle (5.65,4.64);
  \node[anchor=west] at (6.65,4.56) {60.34};

  \node[anchor=center,text=black!60,font=\tiny\bfseries] at (0.16,4.30) {7};
  \node[anchor=west] at (0.38,4.30) {SANA-WM-AR};
  \fill[barteal] (2.85,4.22) rectangle (5.61,4.38);
  \node[anchor=west] at (6.65,4.30) {59.37};

  \node[anchor=center,text=black!60,font=\tiny\bfseries] at (0.16,4.04) {8};
  \node[anchor=west] at (0.38,4.04) {LingBot-World};
  \fill[barolive] (2.85,3.96) rectangle (5.58,4.12);
  \node[anchor=west] at (6.65,4.04) {58.77};

  \node[anchor=center,text=black!60,font=\tiny\bfseries] at (0.16,3.78) {9};
  \node[anchor=west] at (0.38,3.78) {SANA-WM};
  \fill[barviolet] (2.85,3.70) rectangle (5.39,3.86);
  \node[anchor=west] at (6.65,3.78) {54.61};

  \node[anchor=center,text=black!60,font=\tiny\bfseries] at (0.16,3.52) {10};
  \node[anchor=west] at (0.38,3.52) {Matrix-Game 2.0};
  \fill[barpurple] (2.85,3.44) rectangle (5.26,3.60);
  \node[anchor=west] at (6.65,3.52) {51.92};

  \node[anchor=center,text=black!60,font=\tiny\bfseries] at (0.16,3.26) {11};
  \node[anchor=west] at (0.38,3.26) {Yume 1.5};
  \fill[bargray] (2.85,3.18) rectangle (5.04,3.34);
  \node[anchor=west] at (6.65,3.26) {47.06};

  \node[anchor=center,text=black!60,font=\tiny\bfseries] at (0.16,3.00) {12};
  \node[anchor=west] at (0.38,3.00) {minWM};
  \fill[barsage] (2.85,2.92) rectangle (4.80,3.08);
  \node[anchor=west] at (6.65,3.00) {41.88};

  \node[anchor=west,font=\bfseries] at (0,2.48) {TPV};

  \node[circle,fill=medalgold,text=white,inner sep=0pt,minimum size=0.24cm,font=\tiny\bfseries] at (0.16,2.10) {1};
  \node[anchor=west] at (0.38,2.10) {Happy Oyster};
  \fill[bargreen] (2.85,2.02) rectangle (6.13,2.18);
  \node[anchor=west] at (6.65,2.10) {70.66};

  \node[circle,fill=medalsilver,text=white,inner sep=0pt,minimum size=0.24cm,font=\tiny\bfseries] at (0.16,1.84) {2};
  \node[anchor=west] at (0.38,1.84) {Genie 3};
  \fill[barblue] (2.85,1.76) rectangle (5.72,1.92);
  \node[anchor=west] at (6.65,1.84) {61.86};

  \node[circle,fill=medalbronze,text=white,inner sep=0pt,minimum size=0.24cm,font=\tiny\bfseries] at (0.16,1.58) {3};
  \node[anchor=west] at (0.38,1.58) {LingBot-World~2.0};
  \fill[barslate] (2.85,1.50) rectangle (4.63,1.66);
  \node[anchor=west] at (6.65,1.58) {38.31};

  \node[anchor=center,text=black!60,font=\tiny\bfseries] at (0.16,1.32) {4};
  \node[anchor=west] at (0.38,1.32) {HY-World 1.5};
  \fill[barmauve] (2.85,1.24) rectangle (4.15,1.40);
  \node[anchor=west] at (6.65,1.32) {28.07};

  \node[anchor=center,text=black!60,font=\tiny\bfseries] at (0.16,1.06) {5};
  \node[anchor=west] at (0.38,1.06) {LingBot-World};
  \fill[barolive] (2.85,0.98) rectangle (3.95,1.14);
  \node[anchor=west] at (6.65,1.06) {23.73};
\end{tikzpicture}%
}
\end{minipage}
\end{table*}

\Cref{fig:case_counts} summarizes the number of cases per dimension and their distribution across rollout horizons.
Overall, the benchmark contains 1{,}006 test cases: 648 action cases (351 FPV, 297 TPV), 211 memory cases (103 FPV, 108 TPV), and 147 physics cases (76 FPV, 71 TPV).
Rollout horizons span 10--60\,s. Following the evaluation split, 271 cases are Short-horizon ($t\leq 15$\,s), 298 are Mid-horizon ($15<t\leq 30$\,s), and 437 are Long-horizon ($30<t\leq60$\,s). The Long-horizon group is the largest because it includes every physics case and the longest action and memory rollouts.

\Cref{fig:horizon_dist} shows the rollout-horizon distributions of the Memory and Action suites together with the protocol composition of the Physics suite.
The resulting Action distribution is exactly balanced at 216/216/216 cases, while Memory contains 55/82/74 cases across the three horizon splits.
All Physics rollouts last 60 seconds. Horizon-specific evaluations therefore reuse the single full-suite Physics score as a fixed term. Mechanics comprises clipping (58), collision (14), deformation (12), and gravity (9); Optics comprises reflection (26) and occlusion (8); and 3D Consistency contributes 20 cases, covering all 147 physics cases.

\section{Experiments}
\label{sec:experiments}

\subsection{Evaluated Models}
\label{sec:evaluated_models}

We evaluate representative frontier interactive world models under a unified image-and-action interface.
The benchmark includes two closed-source products, Genie~3 and Happy Oyster, which represent strong publicly accessible interactive world models, and ten publicly released or open-source models covering the dominant design choices in video world models.
These models differ substantially in their control interfaces: some natively accept keyboard actions, some expose pose- or trajectory-conditioned generation, and others use text or learned action tokens.
To ensure comparability, every \ours{} test case is converted into the closest supported control format for each model while preserving the same first-frame image, action order, and evaluation pipeline.

This model set is intentionally heterogeneous and spans four control-interface families.
Genie~3, Happy Oyster, and Matrix-Game~2.0 natively accept keyboard input with no conversion and are closest to the target WASD/IJKL interaction protocol.
LingBot-World, LingBot-World~2.0, Lyra~2.0, SANA-WM, SANA-WM-AR, and minWM are conditioned on continuous camera-pose sequences, testing whether pose-driven models can follow discrete keystroke programs after action-to-pose conversion.
HY-World~1.5 and Matrix-Game~3.0 are conditioned on both discrete keyboard actions and converted camera poses, evaluating a hybrid action-plus-pose interface.
Yume~1.5 represents a caption- or instruction-driven interface, with benchmark keyboard commands aggregated over 32-frame windows and converted into motion captions.
\Cref{tab:models} summarizes the evaluated model set, and Appendix~\ref{sec:appendix_models} gives model-by-model details and the adapter used for each interface.

\Cref{tab:models} also reports inference speed and throughput.
Open-source models vary widely in efficiency:
Yume~1.5 is fastest at 6.49~Hz, followed by Matrix-Game~2.0 (6.26~Hz) and Matrix-Game~3.0 (5.34~Hz), while SANA-WM is slowest at 0.81~Hz, with HY-World~1.5 (1.30~Hz) and LingBot-World (1.79~Hz) similarly throughput-limited.
Normalized by resolution, Yume~1.5 and Matrix-Game~3.0 still lead in MPPS (5.85 and 4.82), reflecting efficient high-resolution generation, whereas HY-World~1.5, LingBot-World, and SANA-WM stay below 1~MPPS (0.52, 0.72, 0.73).
These gaps highlight the trade-off between scale, visual fidelity, and real-time deployability in interactive world generation.

\subsection{Main Results}

\paragraph{Overall ranking.}
\Cref{tab:main_results} reports the first-person and third-person quantitative results, with metrics as rows and models as columns.
Genie~3 achieves the best overall performance in first-person view (70.32), narrowly ahead of Lyra~2.0 (69.61) and Happy Oyster (68.53); Happy Oyster ranks highest in third-person view (70.66) after the TPV overall score is weighted by control rate.
Across both views, the rankings show that \textbf{no model dominates all dimensions}.
Closed-source models are generally stronger in physics and memory, whereas frontier open-source models are clearly stronger in action following and visual quality.
In third-person evaluation, Happy Oyster benefits from the strongest control rate (96.43) and subject memory (81.09), while HY-World~1.5 and LingBot-World~2.0 are stronger in action following and visual quality but control the third-person subject far less reliably. 

\paragraph{Action following.}
Open-source models achieve the strongest action scores in first-person view.
Matrix-Game~3.0 (91.34), Lyra~2.0 (90.82), LingBot-World (90.78), and LingBot-World~2.0 (90.74) are within a point of one another, and this advantage largely carries over to third-person evaluation for the models that support it.
Notably, the two closed-source models are \emph{not} the leading action models: Genie~3 reaches only 76.31 and Happy Oyster 88.22, and Genie~3's per-frame strict accuracy (62.04) is among the lowest in the leaderboard.
High action scores therefore do not imply uniformly strong performance in other dimensions, and conversely the strongest overall models can trade keystroke fidelity for other capabilities.

\paragraph{Visual quality.}
Visual quality is comparatively clustered: ten of twelve models fall within 8.3 points of one another, from SANA-WM-AR (74.40) and SANA-WM (71.42) to Genie~3 (66.21) and HY-World~1.5 (66.06), with only Matrix-Game~2.0 (60.83) and minWM (58.21) clearly behind.
Both SANA-WM variants lead first-person visual quality despite ranking seventh and ninth overall.

\paragraph{Physics.}
In contrast, the physics dimension exposes a far larger gap: Genie~3 (66.67) leads the best open-source model, Lyra~2.0 (51.52), by 15.15 points, while Happy Oyster (61.40) leads it by 9.88 points, and the weakest models score in the single digits.
The mechanics sub-score is the most discriminative of all metrics in the benchmark, with Genie~3 at 61.50 and Happy Oyster at 45.50 while every open-source model stays below 8.
In third-person view, Happy Oyster is strongest overall in physics (68.90), narrowly ahead of Genie~3 (67.15).
These results suggest that current open-source models can follow camera commands well, but still struggle to consistently enforce collision, clipping, terrain, and optical constraints during interaction.

\paragraph{Memory.}
Memory scores reveal a further distinction between controllability and persistence.
Genie~3 shows the strongest first-person memory performance (72.09), followed by Lyra~2.0 (66.49) and SANA-WM-AR (64.48), while the strongest action-following models are markedly weaker: LingBot-World reaches only 43.93 despite a 90.78 action score.
In third-person view Genie~3 leads overall TPV memory (71.27) and Happy Oyster leads subject memory (81.09).
This supports our observation that Happy Oyster has particularly strong third-person subject memory, even though its first-person action-following score is lower than the strongest open-source models.
Memory is also the dimension that degrades most sharply with rollout horizon, which we analyze in \cref{sec:memory_decay}.

\section{Discussion}
\label{sec:discussion}

\subsection{Key Findings}
\label{sec:key_findings}

The evaluation reveals four key findings that challenge common assumptions about interactive world models.

\paragraph{A high trajectory score does not imply high per-frame correctness.}
A high trajectory score does not guarantee correct per-frame actions: models with strong trajectory alignment (trajectory score above 85) can still fall below 65\% per-frame strict action accuracy, revealing a hidden failure mode in which a model eventually drifts along the right path while missing individual keystrokes.
Trajectory-level metrics aggregate motion over long segments, so delayed responses, short-term stalls, and compensatory over-corrections can cancel out geometrically even though the interactive experience remains poor.
This supports the need for keystroke-level evaluation alongside trajectory scoring.

\paragraph{Low visual quality does not necessarily imply poor controllability.}
Visual quality alone is not a reliable indicator of action-following performance: a model with relatively low visual quality can still respond faithfully to commands.
Matrix-Game~2.0, for example, sits near the bottom of first-person visual quality yet ranks fifth on action following, whereas the two SANA-WM variants lead visual quality but fall to ninth and eleventh on action following.
Weak textures, unstable lighting, or plain composition therefore do not preclude faithful responses to user commands, just as high perceptual quality does not guarantee them.

\paragraph{Stricter physics adherence may compromise GT action following.}
Models that avoid clipping, respect collisions, or follow terrain constraints often deviate from predefined ground-truth trajectories, revealing a trade-off between physical plausibility and instruction fidelity.
For example, a model that correctly slows down or stops at a wall will accumulate trajectory error relative to an unconstrained path, even though the behavior is physically desirable.
This motivates separating action-following metrics from interaction-physics metrics rather than collapsing them into a single trajectory score.

\paragraph{Memory evaluation is confounded by action-following imprecision.}
Current models rarely reach the prescribed turning point or revisit location exactly, so symmetric frame pairs often depict different spatial positions.
As a result, frame-pair metrics conflate action error with memory degradation: a model may appear to forget a scene simply because it failed to return to the same viewpoint.
Action-aware transition localization and scene-level 3D point-cloud reconstruction \emph{reduce the sensitivity} of memory scoring to these factors by comparing aggregated geometry across the observation and revisit segments, mitigating the coupling between action-following error and apparent memory loss.
As detailed in Appendix~\ref{sec:frame_pair_memory}, this trajectory-aware protocol is substantially more robust than GT-indexed frame matching under imperfect control.

\subsection{Performance Across Horizons}
\label{sec:horizon_difficulty}

We compare first-person performance over the Short-, Mid-, and Long-horizon splits defined in \cref{sec:testsuite_stats}. For each metric $d\in\{\mathrm{Overall},\mathrm{Vision},\mathrm{Action},\mathrm{Memory}\}$, we define its Short-to-Long decay rate as
\begin{equation}
\rho_d = 1 - \frac{S_d^{\mathrm{Long}}}{S_d^{\mathrm{Short}}}.
\label{eq:horizon_decay}
\end{equation}
A larger $\rho_d$ indicates a greater Short-to-Long score decrease, while $\rho_d<0$ indicates a score increase. The same full-suite Physics score enters every horizon split; these decay rates therefore characterize horizon sensitivity of Action, Vision, Memory, and the Overall composite containing the fixed Physics term, not of Physics itself.

\begin{table*}[t]
\centering
\fontsize{6}{7}\selectfont
\caption{\textbf{First-person scores across horizons.} Each model is evaluated on the Short-, Mid-, and Long-horizon splits for Overall, Vision, Action, and Memory; each $\rho$ column reports the corresponding Short-to-Long decay rate from \cref{eq:horizon_decay}. Rows are ordered by Short-horizon Overall score; gray rows denote closed-source models. \textbf{Bold}/\underline{underline}\,=\,best/second-best within each metric and horizon. Score columns (Short/Mid/Long) are higher-is-better ($\uparrow$); decay-rate columns $\rho$ are lower-is-better ($\downarrow$). Scores are reported $\times100$.}
\label{tab:horizon_scores}
\setlength{\tabcolsep}{4.5pt}
\begin{tabular}{l|cccc|cccc|cccc|cccc}
\toprule
\textbf{Model} & \multicolumn{4}{c|}{\textbf{Overall}} & \multicolumn{4}{c|}{\textbf{Vision}} & \multicolumn{4}{c|}{\textbf{Action}} & \multicolumn{4}{c}{\textbf{Memory}} \\
 & \textbf{Short}\,$\uparrow$ & \textbf{Mid}\,$\uparrow$ & \textbf{Long}\,$\uparrow$ & $\boldsymbol{\rho}\!\downarrow$ & \textbf{Short}\,$\uparrow$ & \textbf{Mid}\,$\uparrow$ & \textbf{Long}\,$\uparrow$ & $\boldsymbol{\rho}\!\downarrow$ & \textbf{Short}\,$\uparrow$ & \textbf{Mid}\,$\uparrow$ & \textbf{Long}\,$\uparrow$ & $\boldsymbol{\rho}\!\downarrow$ & \textbf{Short}\,$\uparrow$ & \textbf{Mid}\,$\uparrow$ & \textbf{Long}\,$\uparrow$ & $\boldsymbol{\rho}\!\downarrow$ \\
\midrule
\rowcolor{closedCol} Genie 3 & \textbf{73.03} & 69.85 & \textbf{67.55} & 7.5\% & 68.04 & 66.13 & 64.46 & 5.3\% & 77.31 & 74.41 & 77.22 & 0.1\% & \textbf{80.10} & \textbf{72.21} & \textbf{61.84} & 22.8\% \\
Lyra 2.0 & \underline{72.80} & \textbf{70.63} & \underline{64.82} & 11.0\% & 73.80 & 70.47 & 64.58 & 12.5\% & 92.16 & 91.02 & 89.26 & 3.1\% & 73.72 & \underline{69.52} & 53.91 & 26.9\% \\
\rowcolor{closedCol} Happy Oyster & 71.60 & \underline{69.99} & 63.28 & 11.6\% & 69.21 & 66.90 & 63.36 & 8.5\% & 88.71 & 88.94 & 87.02 & 1.9\% & 67.06 & 62.72 & 41.32 & 38.4\% \\
HY-World 1.5 & 71.00 & 66.85 & 53.28 & 25.0\% & 72.64 & 68.32 & 57.22 & 21.2\% & \textbf{93.69} & 91.08 & 77.38 & 17.4\% & \underline{77.23} & 67.58 & 38.10 & 50.7\% \\
LingBot-World~2.0 & 69.52 & 63.18 & 58.30 & 16.1\% & 72.39 & 69.16 & 64.09 & 11.5\% & 90.42 & \underline{91.39} & \textbf{90.43} & 0.0\% & 71.77 & 48.65 & 35.15 & 51.0\% \\
Matrix-Game 3.0 & 65.91 & 61.77 & 52.17 & 20.8\% & 72.79 & 69.93 & 63.36 & 13.0\% & \underline{92.31} & \textbf{91.85} & 89.85 & 2.7\% & 72.96 & 59.75 & 29.91 & 59.0\% \\
LingBot-World & 63.36 & 58.94 & 53.10 & 16.2\% & 72.95 & 69.87 & 65.01 & 10.9\% & 91.55 & 90.83 & \underline{89.98} & 1.7\% & 57.84 & 43.96 & 26.34 & 54.5\% \\
SANA-WM-AR & 60.36 & 61.54 & 55.92 & 7.4\% & \textbf{77.40} & \textbf{74.48} & \textbf{71.33} & 7.8\% & 70.16 & 84.74 & 72.29 & -3.0\% & 70.99 & 64.05 & 57.18 & 19.5\% \\
SANA-WM & 57.61 & 54.13 & 52.25 & 9.3\% & 74.39 & \underline{73.40} & \underline{66.47} & 10.6\% & 67.86 & 79.27 & 64.14 & 5.5\% & 70.83 & 46.48 & \underline{61.03} & 13.8\% \\
Matrix-Game 2.0 & 56.30 & 51.75 & 46.90 & 16.7\% & 67.31 & 60.91 & 54.27 & 19.4\% & 90.32 & 89.85 & 88.83 & 1.6\% & 55.37 & 44.04 & 32.30 & 41.7\% \\
Yume 1.5 & 51.14 & 46.39 & 43.22 & 15.5\% & \underline{74.53} & 68.98 & 63.70 & 14.5\% & 71.53 & 69.97 & 73.81 & -3.2\% & 44.17 & 32.28 & 21.02 & 52.4\% \\
minWM & 45.45 & 42.63 & 37.15 & 18.3\% & 66.84 & 58.92 & 48.86 & 26.9\% & 64.96 & 62.11 & 55.31 & 14.9\% & 42.86 & 42.36 & 37.31 & 12.9\% \\
\bottomrule
\end{tabular}%
\end{table*}

\begin{figure*}[t]
\centering
\includegraphics[width=0.98\textwidth]{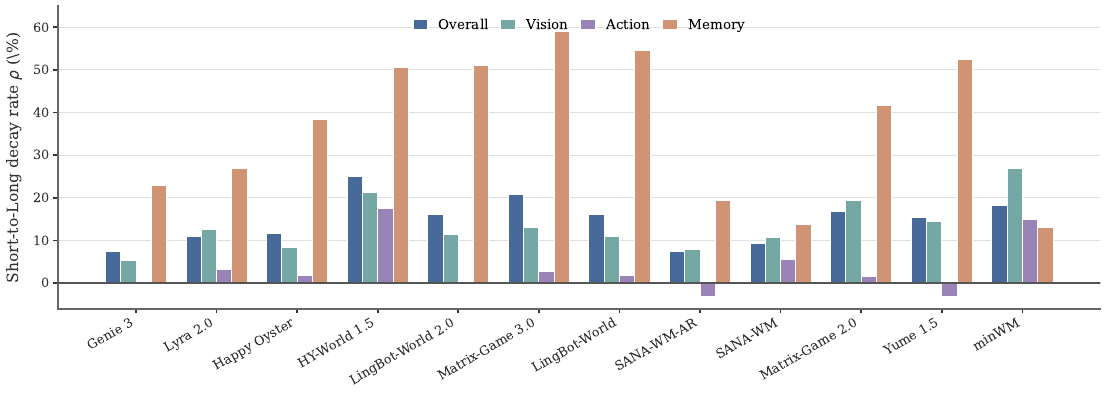}
\caption{\textbf{Short-to-Long horizon decay rates.} Decay rates $\rho_d$ from \cref{eq:horizon_decay} for Overall, Vision, Action, and Memory. Positive values denote degradation; negative Action values denote models that improve at Long horizon.}
\label{fig:horizon_decay}
\end{figure*}

\subsubsection{Overall Performance}
The mean Overall score decreases from 63.17 at Short horizon to 59.80 at Mid horizon and 54.00 at Long horizon, corresponding to 14.6\% average decay (\cref{tab:horizon_scores,fig:horizon_decay}). All twelve models decline, but by markedly different amounts: SANA-WM-AR and Genie~3 lose only 7.4\% and 7.5\%, whereas HY-World~1.5 and Matrix-Game~3.0 lose 25.0\% and 20.8\%. Horizon robustness is therefore a complementary axis that a single aggregate score conceals.

Long-horizon evaluation is also more discriminative among the leaders. The top-four spread grows from only 2.03 points at Short horizon (73.03--71.00) to 9.25 points at Long horizon (67.55--58.30), a $4.6\times$ increase, while the full-leaderboard spread grows only modestly from 27.58 to 30.40 points. Longer interaction therefore separates models that appear nearly tied on short clips rather than shifting all scores uniformly.

The podium remains stable---Genie~3, Lyra~2.0, and Happy Oyster hold the top three positions at every horizon---but the mid-field reshuffles. SANA-WM-AR rises from 8th to 5th and LingBot-World~2.0 from 5th to 4th, while HY-World~1.5 falls from 4th to 6th and Matrix-Game~3.0 from 6th to 9th. Five models change rank, producing 10 positions of total absolute rank migration.

\subsubsection{Action Following}
Action is substantially more stable: the mean score changes from 82.58 to 83.79 and 79.63, yielding only 3.6\% average decay. LingBot-World~2.0 is effectively horizon-invariant and rises from 5th to 1st at Long horizon, while Yume~1.5 and SANA-WM-AR improve relative to Short horizon ($\rho=-3.2\%$ and $-3.0\%$). HY-World~1.5 and minWM are the main exceptions, decaying by 17.4\% and 14.9\%; HY-World~1.5 consequently falls from 1st to 7th.

\begin{figure*}[t]
\centering
\includegraphics[width=0.98\textwidth]{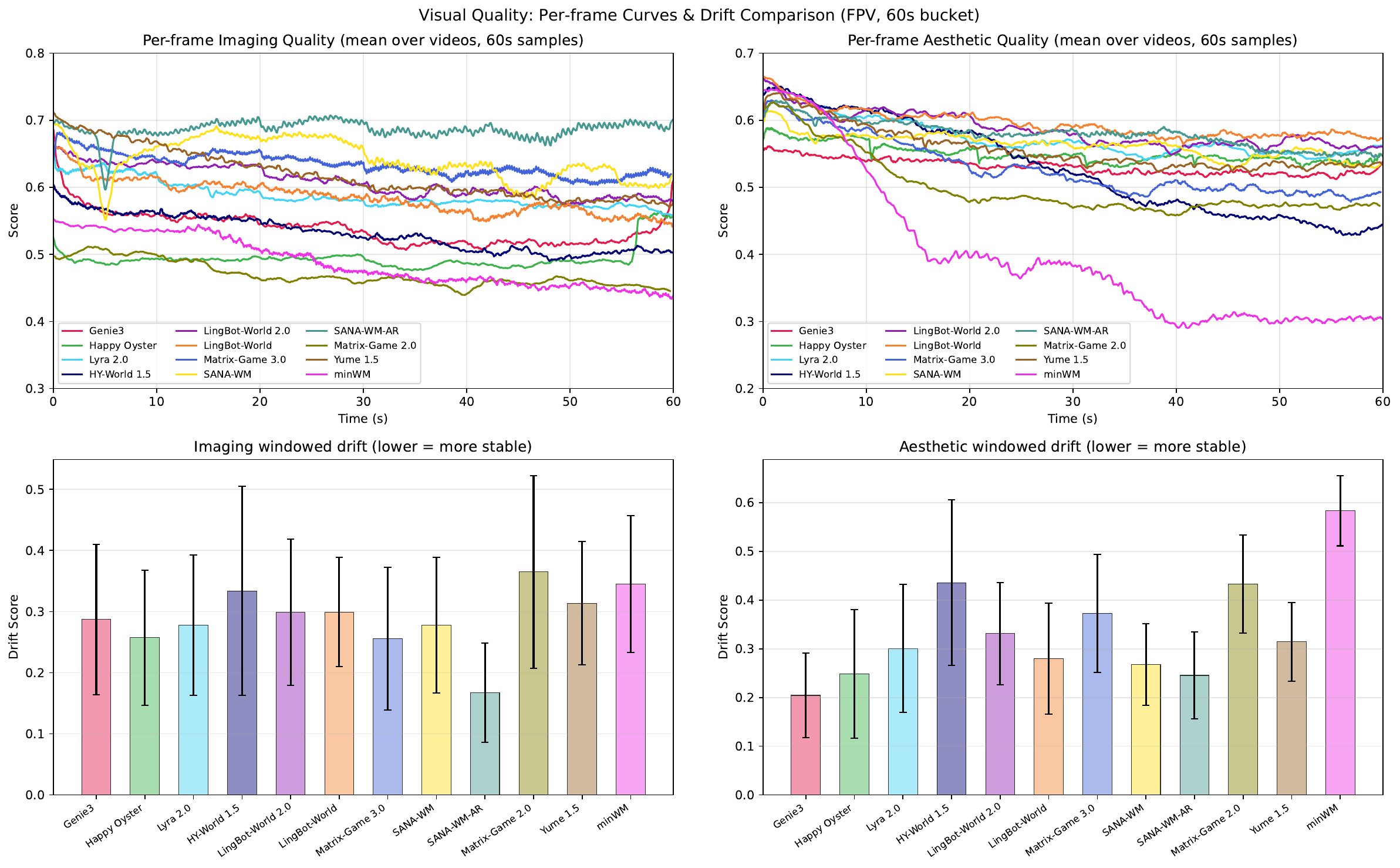}
\caption{\textbf{Temporal visual quality and windowed drift (first-person, 60\,s).} Top: mean imaging and aesthetic scores as functions of elapsed time over 117 cases per model; timestamps are derived using each model's native output frame rate (12--24\,fps), and each time point averages videos that reach that elapsed time. Bottom: mean per-video relative drop from the best to worst 3\,s window (1\,s stride; lower is more stable); black error bars denote $\pm1$\,std across videos.}
\label{fig:visual_curves_drift}
\end{figure*}

\subsubsection{Visual Quality}
Mean Visual Score decreases monotonically from 71.86 to 68.12 and 62.23, with 13.5\% average decay. Because Visual Score includes range-based drift, cross-horizon differences also depend on the number of evaluated windows and do not isolate temporal degradation. SANA-WM-AR remains the visual leader (77.40$\to$71.33), while Genie~3 has the lowest decay (5.3\%) and rises from 10th to 5th. minWM and HY-World~1.5 show the largest relative score decreases (26.9\% and 21.2\%), and Yume~1.5 falls from 2nd to 7th.

\subsubsection{Memory}
\label{sec:memory_decay}
Memory is the most horizon-sensitive dimension: its mean score falls from 65.41 at Short horizon to 54.47 at Mid horizon and 41.28 at Long horizon, corresponding to 37.0\% average decay. A model can therefore retain action responsiveness while progressively losing previously observed scene content. At Long horizon, only Genie~3 (61.84) and SANA-WM (61.03) remain above 60, while seven models fall below 40.

Model-wise decay varies from 12.9\% (minWM) to 59.0\% (Matrix-Game~3.0), defining fast ($\rho\geq50\%$), moderate ($30\%\leq\rho<50\%$), and slow ($\rho<30\%$) regimes. Its rank correlation with Short-horizon memory is only $+0.06$, so strong initial memory does not imply robustness: Matrix-Game~3.0 falls from 4th to 10th, whereas SANA-WM rises from 7th to 2nd.

Memory rankings are substantially more volatile than Overall rankings. From Short to Long horizon, 11 of the 12 
   models change rank in Memory, compared with only 5 in Overall. Genie~3 is the only model that retains first place at all three horizons; the largest movements are Matrix-Game~3.0 ($4\to10$), minWM ($12\to7$), SANA-WM ($7\to2$), and HY-World~1.5 ($2\to6$).

Across metrics, horizon sensitivity follows a clear ordering: Memory Score declines most rapidly, Visual Score and Overall Score show moderate decay, and Action Score remains comparatively stable.

\begin{figure*}[t]
\centering
\includegraphics[width=0.98\textwidth]{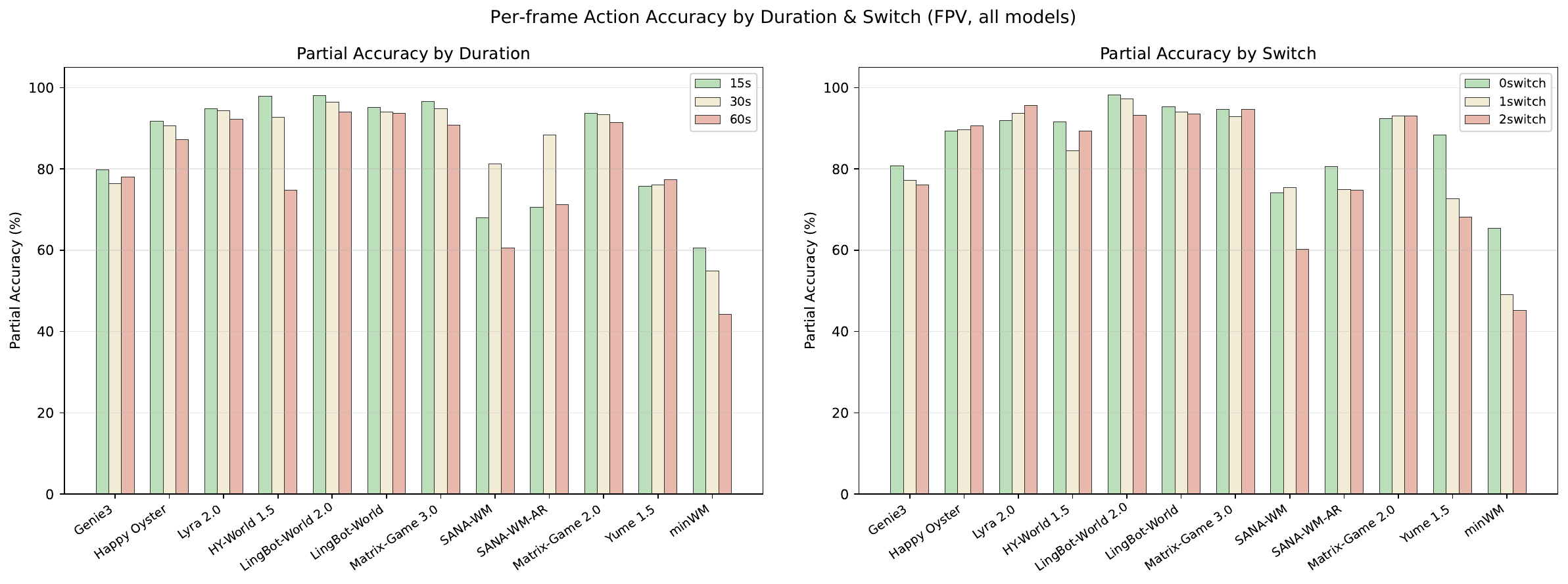}
\caption{\textbf{Per-frame partial action accuracy by rollout duration and action-switch count (first-person).} Bars are grouped by rollout duration (left: 15/30/60\,s) and number of ground-truth action switches (right: 0/1/2). The balanced $3\!\times\!3$ design has 39 cases per duration--switch cell and 351 per model; each bar is the unweighted mean over 117 cases. Camera poses use the MegaSaM\,+\,DA-v3 pipeline. A frame is a partial match when the predicted and ground-truth atomic-action sets intersect or are both empty.}
\label{fig:action_accuracy_difficulty}
\end{figure*}

\subsection{Other Observations in World Models}
\label{sec:observations_wm}

\paragraph{Visual quality degradation exhibits model-specific temporal patterns.}
\Cref{fig:visual_curves_drift} shows model-specific and axis-dependent degradation over 60\,s.
SANA-WM-AR is the most stable on imaging (16.73 windowed drift; $0.691\!\to\!0.692$ from the first to last 3\,s), while Genie~3 is the most stable on aesthetics (20.45). Matrix-Game~3.0 and Happy Oyster follow on imaging (25.57/25.72), and SANA-WM-AR and Happy Oyster on aesthetics (24.55/24.89).
At the other extreme, minWM exhibits selective collapse: its aesthetic score falls from 0.643 to 0.310, mostly within the first 40\,s, producing the largest aesthetic drift (58.36), while its imaging drift is 34.50. Matrix-Game~3.0 shows a milder version of this asymmetry (25.57 imaging vs.\ 37.27 aesthetics), confirming that the axes capture distinct failure modes. HY-World~1.5 and Matrix-Game~2.0 instead degrade progressively on both axes (33.38/43.60 and 36.50/43.32); they have the largest aesthetic drifts after minWM, and Matrix-Game~2.0 also has the largest imaging drift.
Happy Oyster and SANA-WM exhibit transient dips around 5\,s followed by partial recovery, illustrating why best-to-worst windowed drift captures non-monotonic failures that start-to-end comparisons may miss.
Absolute quality and temporal stability are also distinct: Matrix-Game~2.0 has the lowest full-rollout imaging score (0.470), followed by minWM (0.487) and Happy Oyster (0.490), although Happy Oyster has the third-lowest imaging drift. Conversely, Yume~1.5 starts with the highest imaging score (0.696) but reaches 31.34 imaging drift. Thus, time-averaged quality alone does not capture temporal stability.

\paragraph{Longer rollouts and action switching expose model-dependent control failures.}
\Cref{fig:action_accuracy_difficulty} compares partial action accuracy across Short, Mid, and Long rollouts (15/30/60\,s) and 0/1/2 action switches. Action proportions vary across conditions, so these comparisons characterize performance under the benchmark design rather than isolated causal effects. Mean accuracy is 85.2\%, 86.1\%, and 79.6\% across the three horizons, with a 5.6\,pp decrease from Short to Long. Ten of twelve models decline, most notably HY-World~1.5 ($97.9\%\!\to\!74.7\%$) and minWM ($60.6\%\!\to\!44.2\%$). LingBot-World~2.0 maintains the highest accuracy at every horizon (98.1/96.4/94.1\%). SANA-WM and SANA-WM-AR peak at Mid horizon, while SANA-WM-AR and Yume~1.5 score slightly higher at Long than at Short, showing that longer rollouts do not uniformly reduce action accuracy.

Across 0/1/2 switches, mean accuracy is 86.9\%, 82.9\%, and 81.2\%. Eight models decline from zero to two switches, but only six decline monotonically. The largest decreases occur for minWM ($20.2$\,pp), Yume~1.5 ($20.1$\,pp), and SANA-WM ($13.9$\,pp), whereas Lyra~2.0 improves by $3.7$\,pp and the Matrix-Game models remain nearly flat. Yume~1.5 improves slightly with duration yet drops sharply with more switches, illustrating why robustness to longer rollouts does not necessarily imply robustness to action switching.

Partial accuracy credits shared primitives without requiring an exact action match. For example, Yume~1.5 achieves 88.3\% partial versus 71.4\% strict accuracy with zero switches, while LingBot-World~2.0 achieves 97.2\% versus 80.6\% with one switch. These gaps show that high partial accuracy can coexist with substantial errors in the complete action label.

\paragraph{Third-person controllability varies dramatically across models.}
Before the metrics are evaluated, each model is tested for controllable third-person generation.
A test case is counted as controllable if its generated rollout passes the TPV controllability gate (Appendix~\ref{sec:appendix_validity_details}).
Pooled over the 476 third-person test cases that enter the action (297), physics (71), and memory (108) dimensions, Happy Oyster achieves the highest control rate (96.43), followed by Genie~3 (83.82), LingBot-World~2.0 (53.78), HY-World~1.5 (39.71), and LingBot-World (35.50), matching the TPV control-rate row in \cref{tab:main_results} and summarized in \cref{tab:tpv_control_rate}.
The resulting gap is substantial: Happy Oyster's control rate is roughly $2.7\times$ that of LingBot-World and $2.4\times$ that of HY-World~1.5.

\begin{table}[t]
\centering
\small
\caption{\textbf{Third-person controllability.} Control rate is the fraction of the 476 TPV test cases (297 action, 71 physics, 108 memory) that pass the controllability gate. Values are reported $\times 100$ to match \cref{tab:main_results}.}
\label{tab:tpv_control_rate}
\setlength{\tabcolsep}{4pt}
\resizebox{\columnwidth}{!}{%
\begin{tabular}{lccccc}
\toprule
\textbf{Model} & \textbf{Happy~Oyster} & \textbf{Genie~3} & \textbf{LingBot-World~2.0} & \textbf{HY-World~1.5} & \textbf{LingBot-World} \\
\midrule
\textbf{Control Rate} & \textbf{96.43} & \underline{83.82} & 53.78 & 39.71 & 35.50 \\
\bottomrule
\end{tabular}%
}
\end{table}

Qualitatively, Genie~3 and Happy Oyster exhibit substantially stronger third-person control than the open-source models.
Both models maintain coherent third-person interaction across realistic and game-like scenes, with Happy Oyster performing particularly strongly; this pattern is consistent with its high third-person subject memory score in \cref{tab:main_results}.
One plausible explanation is that Happy Oyster includes specialized optimization for third-person data, avatar-centric interaction, or architectures that explicitly support third-person control.
By contrast, HY-World~1.5 exhibits third-person controllability mainly in particular game-style domains, such as Unity-like scenes, first-person-shooter-like scenes, or Genshin-style scenes.
LingBot-World, LingBot-World~2.0 and HY-World~1.5 do not expose an explicit hyperparameter for switching between first-person and third-person generation, suggesting that their third-person abilities emerge from training data and the base model rather than from an explicit mechanism for selecting FPV or TPV, whereas the strongest closed-source models may benefit from more deliberate third-person design.

\paragraph{First-person physics depends on whether the camera is embodied by a held object.}
In the physics dimension, the same first-person scene can yield different physical behavior depending on whether the view contains a visible held object.
For example, first-person views containing a weapon in an FPS-style scene, a cup in hand, or another foreground object behave differently from clean camera-only first-person views with no visible body or object.
Within the same scene, Genie~3 is less likely to produce clipping violations when the first-person view is visually grounded by a held object, whereas clean first-person camera views more often pass through walls or other solid geometry.
Textual prompts that describe the first-person subject as human do not fully remove this gap.
This suggests that even models with strong overall physics performance, such as Genie~3 and Happy Oyster, may rely partly on data-derived visual priors for physical compliance rather than fully internalized physical understanding.

\subsection{Limitations and Future Work}

Although \ours{} provides a comprehensive evaluation of action following, visual quality, interaction physics, and memory, several limitations remain. First, the current benchmark covers rollouts of up to 60 seconds and a finite set of scene categories, interaction patterns, and physical phenomena. Its action space is also primarily restricted to camera movement and navigation commands. As interactive world models increasingly support richer actions, such as attacking, jumping, object manipulation, and environment interaction, the current suite does not fully capture their emerging capabilities. Second, several evaluation components rely on estimated camera poses, monocular depth, semantic segmentation, and VLM judgments; errors in these intermediate components may propagate into the final scores. Third, differences in native resolution, frame rate, control interface, and access conditions---particularly for closed-source systems---prevent perfectly standardized inference across all evaluated models. Fourth, because all physics rollouts last 60 seconds, the physics dimension is not stratified by rollout horizon: horizon-specific evaluations reuse the single full-suite physics score as a fixed term, so the current analysis cannot characterize how physical plausibility evolves with interaction duration.

Future extensions should evaluate substantially longer interaction horizons and broaden the diversity of open-world environments, visual styles, weather conditions, and interaction contexts. The action suite should be expanded beyond navigation to cover richer embodied interactions and compositional action sequences. The physics suite can likewise be extended beyond mechanics and optics to include electricity, thermodynamics, and other forms of material or energy transfer. In particular, constructing physics cases at multiple rollout durations would enable a duration ablation of the physics dimension and reveal whether adherence to physical laws degrades over longer horizons. Future work should also improve the robustness, calibration, and uncertainty estimation of the evaluation pipeline, while developing more unified inference and control protocols for cross-model comparison.

\section{Conclusion}
\label{sec:conclusion}

\ours{} benchmarks long-horizon interactive world models under continuous WASD/IJKL control.
By combining keystroke-level action following, visual stability, interaction physics, and scene/subject-level memory evaluation, \ours{} exposes failure modes that are largely hidden by short clips, trajectory-only metrics, or frame-pair memory tests.
Evaluation of twelve open- and closed-source models shows that no model dominates all dimensions: strong visual quality or trajectory alignment does not guarantee per-frame controllability, physical plausibility, or reliable memory.
Stratifying action following, visual stability, and memory by rollout horizon, while holding Physics at its full-suite value, further shows that the leaderboard obtained from short clips is not the leaderboard that matters for sustained interaction.
These results highlight the need for future world models to integrate action responsiveness, physical grounding, and persistent scene and subject memory.
\ours{} provides a diagnostic foundation for measuring progress toward genuinely interactive and stable world simulation.

{\small
\bibliographystyle{ieee_fullname}

}

\clearpage
\appendix
\crefalias{section}{appendix}
\crefalias{subsection}{appendix}
\crefname{algorithm}{Algorithm}{Algorithms}
\Crefname{algorithm}{Algorithm}{Algorithms}

\begin{strip}
\centering
\includegraphics[width=\textwidth]{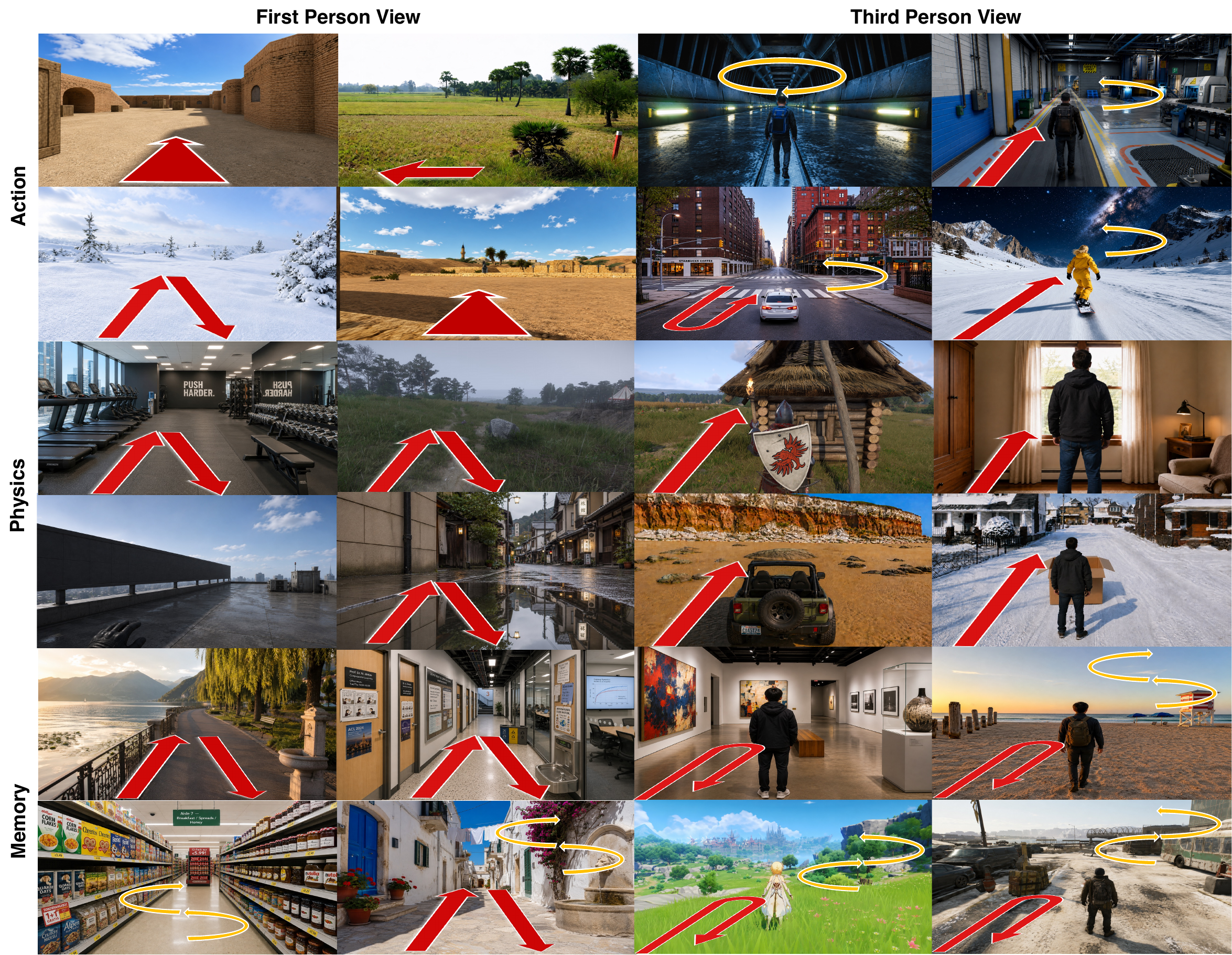}
\captionof{figure}{\textbf{Test suite gallery.} Each subfigure shows a representative test case with the first frame, trajectory schematic, and action sequence. The gallery is organized by scene category from top to bottom (Indoor, Urban, Nature) and by view from left to right (FPV, TPV).}
\label{fig:test_suite_gallery}
\end{strip}

\begin{figure}[t]
\centering
\includegraphics[width=\columnwidth]{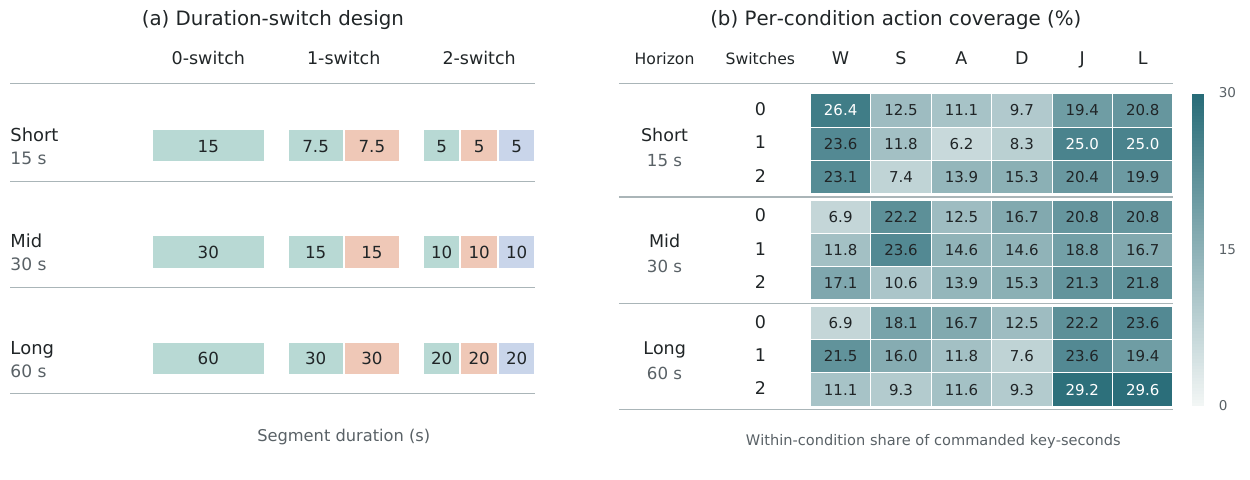}
\caption{\textbf{Action-control test design and coverage.} (a) Nine duration--switch combinations per initial image, with segment durations in seconds. Colors indicate segment positions, not fixed actions. (b) Within-condition action shares (\%) measured in commanded key-seconds. All six keys occur in every condition.}
\label{fig:action_design_coverage}
\end{figure}

\section{Test Suite Details}
\label{sec:appendix_gallery}

To provide a qualitative overview of the visual and action coverage in \ours{}, we include a gallery of representative test cases in \Cref{fig:test_suite_gallery}.
Each subfigure shows the first-frame image together with an action schematic, illustrating both the scene context and the intended WASD/IJKL trajectory.
The gallery is organized along two axes: scene category and camera view.
Rows correspond to the three major scene families used throughout the benchmark---Indoor, Urban, and Nature---while columns separate first-person and third-person views.
This layout highlights that each semantic environment type is evaluated under both egocentric exploration and avatar-centered interaction, enabling the benchmark to probe camera control, physical interaction, scene memory, and subject memory across diverse visual domains.

\subsection{Action-Control Test Design}
\label{sec:appendix_action_design}

\paragraph{Rollout duration and action switches.}
Each first-frame image is paired with nine action sequences, crossing three rollout durations (15/30/60\,s) with three action-switch counts (0/1/2), as illustrated in \Cref{fig:action_design_coverage}(a). A rollout contains one, two, or three equal-duration segments, respectively; each segment sustains a single key, and adjacent segments use different keys. The 39 FPV and 33 TPV images yield 351 and 297 cases, respectively. Equal case counts across the nine conditions support comparisons across horizons at a fixed switch count and across switch counts at a fixed horizon.

\paragraph{Scene-aware sequence construction.}
Initial images are inspected for forward and lateral clearance, and action identities and templates are adapted to scene layout and rollout duration. Lateral motion is restricted to scenes with sufficient lateral clearance and excluded for TPV vehicle subjects. In sequences containing both translation and yaw rotation, all translation segments precede the rotations. Models can produce different turning angles under the same yaw command; this ordering avoids subsequent translation along model-dependent headings, reducing the risk that collision responses are mistaken for action-following errors.

\paragraph{Representative patterns and action coverage.}
Templates include translation followed by opposite yaw commands (\texttt{W}$\rightarrow$\texttt{J}$\rightarrow$\texttt{L}), same-axis reversals (\texttt{A}$\rightarrow$\texttt{D}$\rightarrow$\texttt{A}), and alternating yaw commands (\texttt{J}$\rightarrow$\texttt{L}$\rightarrow$\texttt{J}). These cover transitions between motion modalities and direction reversals without placing translation after rotation. The suite uses four translation keys (\texttt{W/S/A/D}) and two yaw keys (\texttt{J/L}); left/right variants are alternated across samples to improve directional balance subject to scene and subject constraints. \Cref{fig:action_design_coverage}(b) reports the within-condition action shares, computed from commanded key-seconds. All six action primitives are represented in every duration--switch condition, supporting per-action evaluation across horizons and switch counts. \Cref{fig:action_key_dist} reports the aggregate distribution.

\section{Evaluated Model Details}
\label{sec:appendix_models}

\ours{} evaluates models as black-box or gray-box interactive generators: each model receives the same initial frame and an action program derived from the WASD/IJKL vocabulary, then generates a video rollout scored by the same action, visual-quality, physics, and memory pipelines.
No model is fine-tuned on \ours{}.
For models with non-keyboard interfaces, deterministic adapters translate the benchmark action program into each model's supported control format. Scores therefore characterize performance under the evaluated configurations and adapters.
\Cref{tab:model_details_appendix} summarizes the adapter used for each evaluated model.

\begin{table*}[t]
\centering
\small
\caption{\textbf{Model evaluation adapters.} Each model receives the benchmark's WASD/IJKL actions through a deterministic adapter that converts them to the model's native input format.}
\label{tab:model_details_appendix}
\setlength{\tabcolsep}{8pt}
\begin{tabular}{@{}lp{12cm}@{}}
\toprule
\textbf{Model} & \textbf{Evaluation Adapter} \\
\midrule
Genie~3~\cite{genie3_2025} & Native WASD/IJKL support. Actions are injected via frontend automation at 1/20~s per action, synchronized with the model's 20~fps output. \\
Happy Oyster~\cite{happyoyster2026} & Native WASD/IJKL support. Actions are injected via frontend automation at 1/24~s per action, synchronized with the model's 24~fps output. \\
LingBot-World~\cite{lingbotworld} & Autoregressive (Fast) checkpoint; benchmark actions are supplied through camera-pose conditioning. Actions are converted to camera poses at a translation speed of 0.02 per frame and a rotation speed of $0.75^{\circ}$ per frame. \\
LingBot-World~2.0~\cite{lingbot-world-v2} & Benchmark WASD/IJKL commands are converted into camera-pose sequences for the evaluated configuration. \\
HY-World~1.5~\cite{hyworld15} & Native keyboard support, but the model accepts one action per latent frame. Four consecutive frame-level actions are merged into a single latent-frame action before being fed to the model. \\
Lyra~2.0~\cite{lyra2_2026} & Actions are converted to camera poses at a translation speed of 0.02 per frame and a rotation speed of $0.75^{\circ}$ per frame, preserving action order. \\
SANA-WM~\cite{sanawm} & WASD/IJKL are mapped to local camera increments: A/D for translation and J/L for rotation. \\
SANA-WM-AR~\cite{sanawm} & Chunk-causal configuration; benchmark WASD/IJKL commands are converted into camera-pose conditioning, with 24 generated video frames per chunk. \\
Matrix-Game~2.0~\cite{matrixgame2} & Native per-frame action support; each benchmark action is passed directly as a per-frame action token. \\
Matrix-Game~3.0~\cite{matrixgame3} & Native per-frame action support; an external adapter converts benchmark keyboard actions into per-frame action conditions and camera-pose conditioning. \\
Yume~1.5~\cite{yume15} & In the evaluated configuration, motion control is expressed through text rather than pose or action tensors; the initial image is also provided. Every 32 frame-level actions are aggregated by taking the dominant WASD and IJKL keys within the segment, then rendered into a fixed-template motion caption (e.g., ``Person moves forward; Camera turns left''). Each caption controls one autoregressive chunk of 32 pixel frames. \\
minWM~\cite{minwm2026} & Benchmark actions are converted into camera-pose sequences and supplied to the model together with the initial image. \\
\bottomrule
\end{tabular}
\end{table*}

\subsection{Model Descriptions}

\paragraph{Genie~3~\cite{genie3_2025}.}
Genie~3 is Google DeepMind's closed-source interactive world model.
It generates subsequent video frames in response to user actions, enabling interactive exploration of generated environments.
Public demonstrations also show the persistence of previously observed scene content and changes to the environment through text prompts.
We evaluate Genie~3 through its web interface, using browser automation to issue the benchmark keyboard commands after providing the initial image and associated scene and character descriptions.
We treat the model as a black box, without access to its internal architecture or sampling configuration.
Genie~3 ranks first in first-person Overall, Physics, and Memory Scores, but eighth in Action Score and ninth in Visual Score.

\paragraph{Happy Oyster~\cite{happyoyster2026}.}
Happy Oyster is Alibaba's closed-source interactive world model.
It supports exploration of generated environments and scene editing through text, voice, and image inputs.
We evaluate its responses to the benchmark keyboard commands through its web interface using browser automation.
As with Genie~3, we treat the model as a black box.
Happy Oyster ranks first in third-person Overall and Physics Scores and in the Subject Memory submetric, but third in Action Score and last in Visual Score among the five third-person models.

\paragraph{LingBot-World~\cite{lingbotworld}.}
LingBot-World is Ant Group's open-source world-model framework built on a Wan2.2 image-to-video backbone.
Its training introduces action conditioning and progressively increases the video duration to learn control responses and long-term temporal dependencies.
The Fast variant is obtained through block-causal adaptation and few-step distillation of the bidirectional model, allowing video chunks to be generated sequentially from past visual context and control inputs.
We evaluate the 14B \texttt{LingBot-World-Fast} checkpoint, converting benchmark navigation and view-control commands into camera-pose sequences.
Its scene memory is learned from video context rather than maintained in an explicit 3D scene representation.
LingBot-World ranks third in first-person Action Score but tenth in Memory Score.

\paragraph{LingBot-World~2.0~\cite{lingbot-world-v2}.}
LingBot-World~2.0 is Ant Group's open-source interactive world model with camera-pose and text conditioning.
Compared with LingBot-World, LingBot-World~2.0 introduces causal world-model pretraining before distillation, learning to generate subsequent video chunks from historical visual context and control inputs.
Camera poses control viewpoint motion, while per-chunk text prompts specify character actions and environmental events.
Post-training combines distillation with optimization over long rollouts to improve generation efficiency and reduce accumulated errors.
We evaluate the \texttt{lingbot-world-v2-14b-causal-fast} checkpoint with 16 generated video frames per chunk, converting benchmark navigation and view-control commands into camera-pose sequences.
LingBot-World~2.0 ranks first in third-person Visual Score but fourth out of five models in third-person Memory Score.

\paragraph{HY-World~1.5~\cite{hyworld15}.}
HY-World~1.5 is Tencent Hunyuan's open-source interactive world model built on HunyuanVideo~1.5.
It uses a dual action representation that combines discrete action signals with continuous camera information.
Its reconstituted context memory retrieves relevant past frames and adjusts their temporal encoding to retain access to observations from earlier parts of the rollout.
We evaluate the 8B configuration, mapping benchmark navigation and view-control commands to the model's action-conditioning inputs.
HY-World~1.5 ranks first in third-person Action Score but last in third-person Physics Score among the five evaluated models.

\paragraph{Lyra~2.0~\cite{lyra2_2026}.}
Lyra~2.0 is NVIDIA's open-source framework for generating explorable 3D environments from camera-controlled video.
Its video generator uses per-frame geometry to retrieve relevant past frames and establish dense correspondences with target viewpoints.
The retrieved information guides appearance synthesis by the generative model.
During training, self-augmented histories expose the model to its own degraded outputs so that it learns to correct rather than propagate generation errors.
We evaluate the 14B video generator using continuous camera-pose sequences derived from benchmark navigation and view-control commands.
The evaluation scores the generated video, not the framework's downstream 3D reconstruction output.
Lyra~2.0 ranks second in first-person Overall Score and first in the first-person Optics and 3D Consistency submetrics, but ranks fifth in Mechanics.

\paragraph{SANA-WM~\cite{sanawm}.}
SANA-WM is NVIDIA's open-source framework for camera-controlled video generation.
Its hybrid attention architecture combines Gated DeltaNet with softmax attention to model long video sequences, while a dual-branch conditioning mechanism incorporates 6-DoF camera poses.
We evaluate its 2.6B-parameter bidirectional configuration, whose attention allows information exchange across temporal positions within the generation window.
Benchmark navigation and view-control commands are converted into camera-pose sequences and supplied as conditioning inputs.
We report this configuration separately from SANA-WM-AR because the two use different temporal attention structures.
SANA-WM ranks second in first-person Visual Score but eleventh in Action Score and ninth in Physics Score.

\paragraph{SANA-WM-AR~\cite{sanawm}.}
SANA-WM-AR denotes the chunk-causal SANA-WM configuration evaluated in \ours{}.
Its generative backbone processes video in temporal chunks, with each chunk conditioned on preceding visual context rather than later chunks.
This differs from the bidirectional temporal attention used by SANA-WM above.
Our evaluation uses a 2.6B-parameter generative model with 24 generated video frames per chunk.
Benchmark navigation and view-control commands are converted into camera-pose sequences and supplied as conditioning inputs.
SANA-WM-AR ranks first in first-person Visual Score and third in Memory Score, but ninth in Action Score and eighth in Physics Score.

\paragraph{Matrix-Game~2.0~\cite{matrixgame2}.}
Matrix-Game~2.0 is Skywork's open-source 1.8B interactive world model for streaming video generation.
Its action-injection module incorporates keyboard and mouse signals into a causal video generator, and few-step distillation reduces the number of denoising steps required to generate each video chunk.
For \ours{}, benchmark WASD/IJKL commands are converted into the model's action-conditioning inputs.
Matrix-Game~2.0 ranks fifth in first-person Action Score but eleventh in both Visual and Physics Scores.

\paragraph{Matrix-Game~3.0~\cite{matrixgame3}.}
Matrix-Game~3.0 is Skywork's open-source interactive world model with a memory mechanism for long video rollouts.
It uses camera information to retrieve and incorporate relevant historical observations, while training on imperfect generated histories teaches the model to correct accumulated errors.
Multi-segment autoregressive distillation reduces the number of denoising steps required for streaming generation.
We evaluate the 5B configuration, mapping benchmark commands to its action-conditioning inputs, which support discrete action and continuous camera-pose representations.
Matrix-Game~3.0 ranks first in first-person Action Score but seventh in both Physics and Memory Scores.

\begin{figure*}[t]
\centering
\resizebox{0.75\textwidth}{!}{%
\begin{tikzpicture}[
  node distance=1.2cm and 1.5cm,
  every node/.style={font=\footnotesize},
  input/.style={
    rectangle, draw=blue!70, fill=blue!10, rounded corners=3pt,
    minimum width=2.2cm, minimum height=0.7cm, align=center, font=\footnotesize,
    inner sep=4pt
  },
  process/.style={
    rectangle, draw=black!60, fill=gray!12, rounded corners=3pt,
    minimum width=2.2cm, minimum height=0.7cm, align=center, font=\footnotesize,
    inner sep=4pt
  },
  browser/.style={
    rectangle, draw=orange!80, fill=orange!10, rounded corners=3pt,
    minimum width=2.6cm, minimum height=0.7cm, align=center, font=\footnotesize,
    inner sep=4pt
  },
  backend/.style={
    rectangle, draw=purple!70, fill=purple!10, rounded corners=3pt,
    minimum width=2.4cm, minimum height=0.7cm, align=center, font=\footnotesize,
    inner sep=4pt
  },
  output/.style={
    rectangle, draw=green!70, fill=green!10, rounded corners=3pt,
    minimum width=2.2cm, minimum height=0.7cm, align=center, font=\footnotesize,
    inner sep=4pt
  },
  arr/.style={-Stealth, thick, color=black!70},
  label/.style={font=\scriptsize, color=black!80, align=center, fill=white, inner sep=2pt},
  group/.style={
    rectangle, draw=#1!40, fill=#1!4, rounded corners=6pt,
    dashed, thick, inner sep=10pt
  },
]

\node[input] (action) {action sequence\\(WASD/IJKL)};
\node[input, below=1.2cm of action] (image) {initial image};
\node[input, below=1.2cm of image] (prompt) {scene description};

\node[browser, right=2.5cm of image, minimum width=3.0cm, minimum height=1.0cm] (pw) {\textbf{Playwright}\\(Open and Operate\\Happy Oyster, Genie~3)};

\node[browser, below=1.2cm of pw, minimum width=3.0cm] (js) {\textbf{JS Executor}\\(Action Injection;\\dispatch actions at fixed fps)};

\node[backend, right=2.0cm of pw, minimum width=2.6cm] (ho) {\textbf{Backend}\\(DataChannel $\rightarrow$ Model)};

\node[process, below=1.2cm of ho, minimum width=2.6cm] (download) {\textbf{Status Poll}\\+ Download};

\node[output, below=1.2cm of download, minimum width=2.6cm] (video) {generated video};

\begin{scope}[on background layer]
  \node[group=orange, fit=(pw)(js), inner sep=14pt,
    label={[font=\small\bfseries, text=orange!80, anchor=south west]north west:Browser Automation}] {};
  \node[group=purple, fit=(ho)(download)(video), inner sep=14pt,
    label={[font=\small\bfseries, text=purple!70, anchor=south west]north west:Server Side}] {};
\end{scope}

\draw[arr] (action) -- (pw);
\draw[arr] (image) -- (pw);
\draw[arr] (prompt) -- (pw);

\draw[arr] (pw) -- node[left, label, pos=0.5] {KeyEvent dispatch} (js);

\draw[arr] (pw) -- node[above, label, pos=0.5] {create world} (ho);

\draw[arr] (js.east) -- node[below, label, pos=0.5] {DataChannel} (ho.south west);

\draw[arr] (ho) -- node[right, label, pos=0.5] {video URL} (download);

\draw[arr] (download) -- (video);

\node[font=\scriptsize, color=purple!70!black, align=center, text width=3.2cm,
  draw=purple!30, fill=purple!3, rounded corners=2pt, inner sep=3pt]
  at ($(ho) + (0, 1.0)$) {
  Closed-source; no public API;\\video available after session
};

\end{tikzpicture}%
}
\caption{Closed-source automated interaction pipeline for Happy Oyster and Genie~3. The system reads benchmark test cases (action sequence, first frame, and scene description), then uses Playwright to drive the target platform's web interface. The entire action list is injected into the browser's V8 engine via a single \texttt{page.evaluate()} call, where actions are dispatched at intervals matching the model's native frame rate (e.g., $\frac{1}{24}$~s for Happy Oyster, $\frac{1}{20}$~s for Genie~3). The frontend encodes the keyboard commands and sends them to the backend via a WebRTC DataChannel. After the session completes, the backend-generated video is downloaded via status API polling.}
\label{fig:happy_oyster_pipeline}
\end{figure*}

\paragraph{Yume~1.5~\cite{yume15}.}
Yume~1.5 is Shanghai AI Lab's open-source 5B world-generation model with text-based control.
It combines context compression and linear attention to process video history, and uses bidirectional attention distillation to accelerate streaming generation.
In our evaluated configuration, the model receives the benchmark initial image and motion instructions expressed in natural language.
The adapter summarizes the dominant navigation and view-control commands in each 32-frame window using a fixed text template.
The video is evaluated using the same action-following metrics as the other models.
This temporal aggregation may omit brief actions or action transitions within a window.
Yume~1.5 ranks fifth in first-person Visual Score, but tenth in Action Score and last in Memory Score.

\paragraph{minWM~\cite{minwm2026}.}
minWM is ShengShu's open-source framework for converting pretrained video diffusion models into camera-controlled autoregressive generators.
Its training pipeline first adds camera conditioning to a bidirectional backbone, then applies autoregressive training and distillation to obtain a generator that requires few denoising steps.
We use the 8B \texttt{HY15/Action2V/dmd} checkpoint downloaded from the official Hugging Face repository (\texttt{MIN-Lab/minWM}).
For \ours{}, benchmark actions are converted into camera-pose sequences and supplied to the model together with the initial image.
In our evaluation, minWM ranks last in first-person Overall, Action, Visual, and Physics Scores.

\subsection{Closed-Source Model Evaluation}

\subsubsection{Closed-Source Model Data Acquisition}

Evaluating closed-source interactive world models such as Genie~3 and Happy Oyster poses a practical challenge: neither downloadable model weights nor programmatic APIs for batch inference were available to us during the evaluation. We therefore access these models through their product interfaces using browser automation.
A frontend automation pipeline drives the interactive session by injecting WASD/IJKL keystrokes at the model's native frame rate.
For Genie~3, which outputs video at 20~fps, each benchmark action is held for $\frac{1}{20}$~s before the next action is issued; for Happy Oyster, which runs at 24~fps, each action is held for $\frac{1}{24}$~s.
The automation layer injects keyboard events according to a predefined command sequence and timing schedule.
\cref{fig:happy_oyster_pipeline} illustrates the end-to-end pipeline: benchmark test cases are fed into a Playwright-driven browser session, where the full action list is injected and dispatched at the model's native frame rate, and the backend-generated video is retrieved once the session completes.

\subsubsection{RAFT Alignment for Closed-Source Videos}

Unlike open-source models with end-to-end control over the generation pipeline and known frame-to-action correspondence, closed-source models return a video file downloaded from the backend after the interactive session completes.
The downloaded video often contains extra frames before the first effective action takes effect (e.g., loading screens, transition animations, or buffering delays), and may have a different total frame count than the benchmark action sequence.

To align the downloaded video with the ground-truth action program, we use RAFT (Recurrent All-Pairs Field Transforms)~\cite{teed2020raft} optical flow to detect the onset of motion.
Specifically, we compute the mean optical flow magnitude between consecutive frames. The first frame whose optical-flow magnitude exceeds the threshold provides an initial estimate of the effective start frame. This estimate is then manually reviewed to identify the onset of the response to the user's command.
From this start frame, we crop the video to the required number of frames that matches the benchmark action schedule. Manual review also verifies that the video boundaries and duration match the prescribed evaluation window. The verified videos are then used to evaluate action following, visual quality, physics, and memory over matched interaction windows for closed- and open-source models.

\section{Pose Estimation Pipeline Comparison}
\label{sec:appendix_pose_pipelines}

\paragraph{Comparison protocol.}
We compare pose pipelines by their suitability for action decoding. For each video, the \emph{oracle ceiling} is the maximum strict accuracy over a grid of $31$ translation and $8$ rotation thresholds. This reduces dependence on a particular operating point, but measures action recoverability rather than pose error against ground-truth geometry. The evaluation covers $12$ models and $351$ cases per model. This full-grid diagnostic is distinct from the four-candidate selection used in the benchmark.

We compare six variants. Three use ViPE~\cite{vipe} with a UniDepth-L~\cite{unidepth}, Depth-Anything-v3~\cite{depthanythingv3} (DA-v3), or Pi3X~\cite{pi3x}$+$MoGe-2~\cite{moge2} depth prior; two use MegaSaM~\cite{megasam} with a DA-v1$+$UniDepth-L or DA-v3 prior; the sixth uses DA-v3 as a standalone feed-forward pose estimator. The ViPE$+$Pi3X$+$MoGe-2 and MegaSaM$+$DA-v3 variants are configurations that we assembled and implemented on top of the original ViPE and MegaSaM frameworks, respectively.

\paragraph{Configuration differences.}
The five SLAM-based variants share a DROID-SLAM~\cite{droidslam} backbone but differ in their use of depth and dynamic-scene cues. MegaSaM initializes disparity from aligned monocular depth, retains a depth prior during bundle adjustment, and weights flow confidence using a learned motion probability. ViPE uses constant disparity initialization, depth regularization, and instance masks to suppress dynamic regions.

The comparison also varies focal handling and depth-model capacity. The upstream MegaSaM configuration refines a shared focal length when calibration is well-conditioned; our DA-v3 variant fixes the DA-v3 field-of-view estimate. ViPE uses DA-v3 Metric-L, whereas MegaSaM$+$DA-v3 and the standalone baseline use DA-v3 Giant. The latter predicts poses directly without bundle adjustment. ViPE's depth-only post-processing leaves camera poses unchanged and is excluded from this comparison. These configurations compare complete pipelines rather than isolate the effects of individual components.

\begin{table}[h]
\centering
\small
\setlength{\tabcolsep}{4pt}
\caption{Six pose-estimation pipelines, grouped by SLAM framework. ``BA'': global bundle adjustment. ``Focal'': bundle-adjusted focal refinement. ``Ceil.'': mean per-sample oracle ceiling accuracy. $\dagger$ marks configurations assembled and implemented by us on top of the corresponding original framework.}
\label{tab:pose_pipelines}
\begin{tabular*}{\columnwidth}{@{\extracolsep{\fill}}llccc@{}}
\toprule
Tag & Depth prior & BA & Focal & Ceil. \\
\midrule
\multicolumn{5}{@{}l}{\emph{MegaSaM (DROID-SLAM based)}} \\
\texttt{megasam\_dav3}$^\dagger$ & DA-v3 Giant & \cmark & \xmark & \textbf{0.795} \\
\texttt{megasam} & DA-v1$+$UniDepth-L & \cmark & \cmark & 0.770 \\
\addlinespace[2pt]
\multicolumn{5}{@{}l}{\emph{ViPE (DROID-SLAM based)}} \\
\texttt{default} & UniDepth-L & \cmark & \cmark & 0.760 \\
\texttt{dav3} & DA-v3 Metric-L & \cmark & \cmark & 0.759 \\
\texttt{pi3x\_moge2}$^\dagger$ & Pi3X$+$MoGe-2 & \cmark & \cmark & 0.752 \\
\addlinespace[2pt]
\multicolumn{5}{@{}l}{\emph{Feed-forward pose}} \\
\texttt{da3\_direct} & DA-v3 Giant & \xmark & \xmark & 0.689 \\
\bottomrule
\end{tabular*}
\end{table}

\begin{figure}[t]
\centering
\includegraphics[width=\columnwidth]{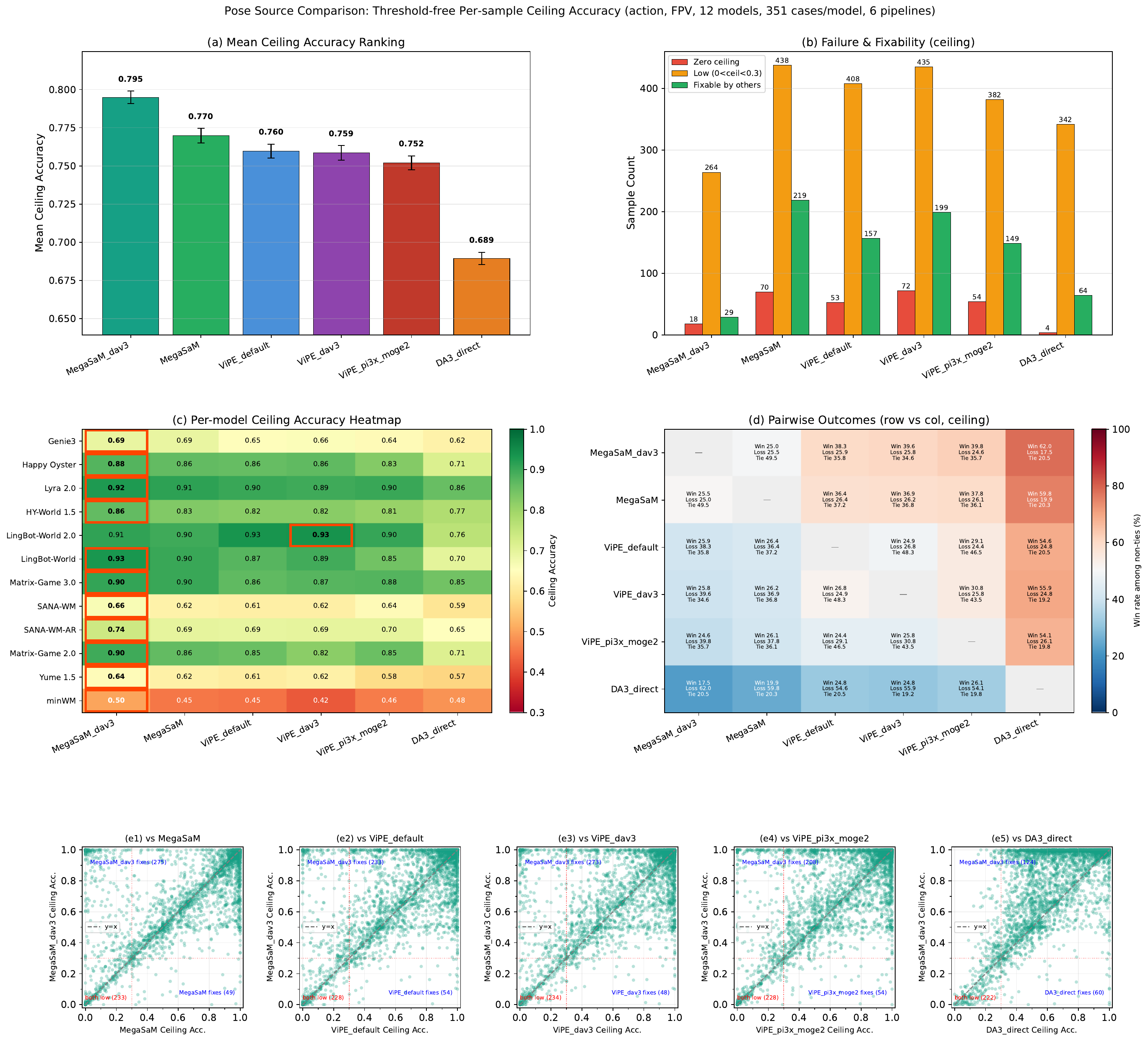}
\caption{\textbf{Action-decoding ceiling across six pose pipelines.} Mean ceiling accuracy and low-accuracy sample rates (top), breakdowns by model and action (middle), and pairwise comparisons (bottom). In the scatter plots, points above the diagonal favor the $y$-axis pipeline.}
\label{fig:pose_ceiling}
\end{figure}

\paragraph{Results and default choice.}
The mean ceiling accuracies follow the ranking $\texttt{megasam\_dav3} > \texttt{megasam} > \texttt{default} \approx \texttt{dav3} > \texttt{pi3x\_moge2} > \texttt{da3\_direct}$ (\Cref{tab:pose_pipelines,fig:pose_ceiling}). \texttt{megasam\_dav3} achieves the highest mean ceiling accuracy ($0.795$) and leads on $11$ of $12$ models, with \texttt{dav3} leading on LingBot-World~2.0. Its advantage therefore extends across most evaluated models.

Relative to upstream MegaSaM, its sample-level win and loss rates are nearly balanced ($25.0\%$ wins, $25.5\%$ losses, and $49.5\%$ ties), yet its mean ceiling is higher by $2.5$ percentage points. This higher mean is accompanied by fewer low-accuracy samples: $6.7\%$ have ceiling accuracy below $0.3$, compared with $12.1\%$ for upstream MegaSaM. The ViPE variants span $0.752$--$0.760$, while standalone DA-v3 reaches $0.689$. We therefore adopt MegaSaM$+$DA-v3 as the default action-evaluation backbone.

\section{Latent-Stride Action Discretization Details}
\label{sec:appendix_compass}

This appendix specifies the windowed pose-to-action decoder, threshold selection, and stride comparison in \Cref{sec:per_frame_action}. The decoder extends CompassReward~\cite{worldcompass2026} with motion aggregation over $K$ consecutive action intervals.

\subsection{Motivation and Windowed Decoding}
\label{sec:appendix_stride_ablation}
Chunk-wise world models such as Matrix-Game~2.0~\cite{matrixgame2} jointly denoise blocks of latent frames and use a video VAE with a temporal compression factor of four. Our benchmark constructs action sequences to maintain a shared command within four-frame groups. Together, these motivate decoding one action over $K$ frames. Aggregating motion over this window can strengthen the motion signal relative to pose-estimation noise, whereas per-frame discretization can produce spurious labels when small inter-frame displacements are comparable to the noise floor.

For each complete window $c=0,\ldots,\lfloor T/K\rfloor-1$, we compute the relative pose as
\begin{equation}
    \Delta P_c=\hat{P}_{cK}^{-1}\hat{P}_{(c+1)K}.
\end{equation}
Its translation is expressed in the camera coordinates at the start of the window. The decoded action is assigned to steps $cK+1,\ldots,(c+1)K$ and compared with their original frame-level commands.

\begin{figure}[H]
\centering
\includegraphics[width=\columnwidth]{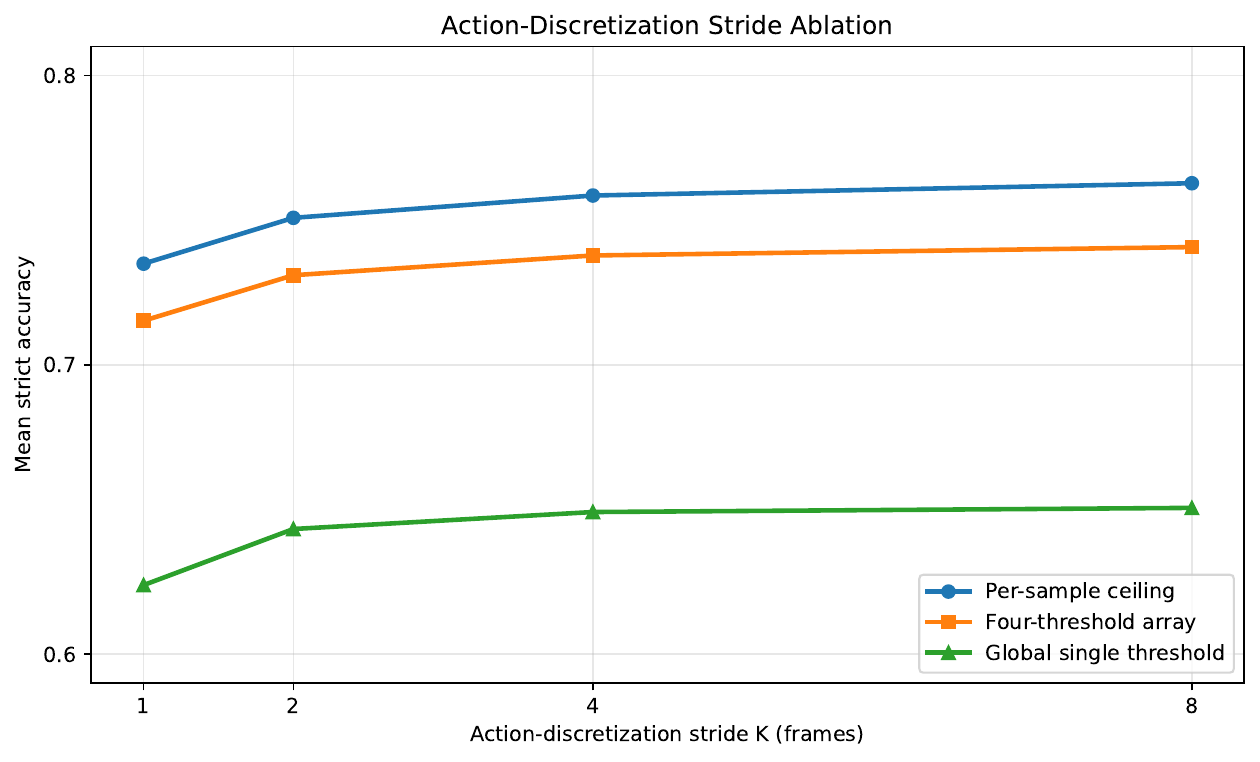}
\caption{Action-discretization stride ablation.}
\label{fig:stride_ablation_curve}
\end{figure}

To assess the choice of window size, we compare $K\in\{1,2,4,8\}$ on all $4{,}212$ first-person action samples from 12 models. Translation and rotation thresholds are recalibrated independently for each $K$. \Cref{fig:stride_ablation_curve} reports three strict-accuracy diagnostics. ``Per-sample ceiling'' is the per-sample oracle over the full threshold grid; ``Four-threshold array'' uses four global translation candidates and one rotation threshold, with the best candidate selected per sample; ``Global single threshold'' uses the best global threshold pair on the same samples.

Across all three strategies, mean strict accuracy improves as $K$ increases from 1 to 4, with diminishing gains beyond $K{=}4$. This trend is consistent with temporal motion aggregation improving robustness to inter-frame pose noise. Increasing the window to $K{=}8$ yields only marginal gains while halving temporal resolution; assigning a shared label to a longer window can introduce errors over multiple frames near action switches. We therefore choose $K{=}4$ to balance decoding accuracy and temporal resolution.

\subsection{Threshold Scaling and Calibration}
The same action can produce different translation magnitudes across models and scenes, making a single motion-detection threshold unsuitable for all rollouts. Following the multi-threshold design of CompassReward~\cite{worldcompass2026}, we calibrate a fixed set of four candidates using statistics across the evaluated models and scenes. Calibration is performed separately for each pose source to account for differences in coordinate scale.

For the final $K{=}4$ decoder, the effective rotation threshold is $\tau_\mathrm{rot}=1.0^\circ$, corresponding to a base threshold of $0.25^\circ$ per action interval. \Cref{tab:compass_thresholds} lists effective translation candidates in each pose source's reconstructed coordinate units; they are used directly without further stride scaling. The stride ablation independently recalibrates both threshold types for each $K$.

The candidate set $\Theta$ is fixed for each pose source. Let $\hat{a}_t^{(\tau)}$ be the label at step $t$ decoded using translation threshold $\tau$ and the fixed rotation threshold. For each video, we select the candidate with the highest strict accuracy against its ground-truth commands:
\begin{equation}
    \tau^*=\operatorname*{arg\,max}_{\tau\in\Theta}
    \frac{1}{T}\sum_{t=1}^{T}\llbracket\hat{a}_t^{(\tau)}=a_t\rrbracket.
\end{equation}
Strict, partial, move, and look accuracy all use the selected sequence $\hat{\mathcal{A}}^{(\tau^*)}$.

\begin{table}[h]
\centering
\small
\setlength{\tabcolsep}{5pt}
\caption{Effective translation thresholds (already scaled by $K{=}4$), calibrated per pose source. The rotation threshold is $0.25^\circ$ before scaling, i.e., $\tau_\mathrm{rot}{=}1.0^\circ$, for every source.}
\label{tab:compass_thresholds}
\begin{tabular*}{\columnwidth}{@{\extracolsep{\fill}}lcccc@{}}
\toprule
Pose source & \multicolumn{4}{c}{Candidate $\tau_\mathrm{move}$} \\
\midrule
\texttt{megasam\_dav3} & 0.0001 & 0.015 & 0.05 & 0.3 \\
\texttt{megasam} & 0.0001 & 0.008 & 0.03 & 0.3 \\
\texttt{default} & 0.0001 & 0.04 & 0.1 & 0.3 \\
\texttt{dav3} & 0.0001 & 0.03 & 0.1 & 0.3 \\
\texttt{pi3x\_moge2} & 0.0001 & 0.04 & 0.12 & 0.3 \\
\texttt{da3\_direct} & 0.0001 & 0.006 & 0.015 & 0.3 \\
\bottomrule
\end{tabular*}
\end{table}

\subsection{Discretization Algorithm}
\label{sec:appendix_discretize_algo}

\Cref{alg:discretize_latent} decodes a window into a label from the command space $\mathbb{A}$ defined in \Cref{sec:per_frame_action}, using its translation magnitude and direction, with signed pitch and yaw. Translation primitives are activated only above $\tau_\mathrm{move}$ and within $60^\circ$ of the corresponding signed camera axis; multiple active primitives form a compound label. Look primitives are activated when the associated angle exceeds $\tau_\mathrm{rot}$ in magnitude. If no primitive is active, the label is \texttt{no\_op}.

For rotation matrix $R$, the decoder uses $s_y=\sqrt{R_{00}^2+R_{10}^2}$, yaw $\psi=\operatorname{atan2}(-R_{20},s_y)$, and pitch $\phi=\operatorname{atan2}(R_{21},R_{22})$. When $s_y\leq10^{-6}$, pitch instead uses $\operatorname{atan2}(-R_{12},R_{11})$. Angles are converted to degrees; positive yaw denotes \texttt{look\_right} and positive pitch denotes \texttt{look\_up}.

\Cref{alg:predict_sequence} combines windowed decoding with threshold selection to produce the frame-level prediction sequence. For \Cref{eq:partial_accuracy}, we use $\mathcal{P}(\texttt{no\_op})=\{\texttt{no\_op}\}$ so that two \texttt{no\_op} labels also count as a partial match.

\begin{algorithm}[t]
\caption{Window-level pose-to-action decoding.}
\label{alg:discretize_latent}
\centering
\small
\fbox{\parbox{0.95\columnwidth}{%
\textbf{Input:} $\Delta P \in \mathrm{SE}(3)$ \hfill$\triangleright$ relative camera pose\\
\quad $\tau_\mathrm{move} \in \mathbb{R}^+$ \hfill$\triangleright$ overall translation threshold\\
\quad $\tau_\mathrm{rot} \in \mathbb{R}^+$ \hfill$\triangleright$ rotation threshold (degrees)\\
\textbf{Output:} $\hat{a} \in \mathbb{A}$ \hfill$\triangleright$ predicted action label\\[-2pt]
\rule{\linewidth}{0.4pt}\\[2pt]
1:\; $\Delta\mathbf{t} \gets \Delta P[:3, 3]$;\quad $\nu \gets \|\Delta\mathbf{t}\|_2$\\
2:\; $\mathbf{o}_t \gets (0,0,0,0)$ \hfill$\triangleright$ (fwd, bwd, right, left)\\
3:\; \textbf{if} $\nu > \tau_\mathrm{move}$ \textbf{then}\\
4:\; \quad $\hat{\mathbf{n}} \gets \mathrm{clip}(\Delta\mathbf{t}/\nu,\, {-}1,\, 1)$\\
5:\; \quad $\boldsymbol{\theta} \gets \arccos(\hat{\mathbf{n}}) \cdot 180/\pi$ \hfill$\triangleright$ angle to each axis\\
6:\; \quad \textbf{if} $\theta_z < 60^\circ$: $\mathbf{o}_t[1] \gets 1$ \hfill$\triangleright$ forward\\
7:\; \quad \textbf{if} $\theta_z > 120^\circ$: $\mathbf{o}_t[2] \gets 1$ \hfill$\triangleright$ backward\\
8:\; \quad \textbf{if} $\theta_x < 60^\circ$: $\mathbf{o}_t[3] \gets 1$ \hfill$\triangleright$ right\\
9:\; \quad \textbf{if} $\theta_x > 120^\circ$: $\mathbf{o}_t[4] \gets 1$ \hfill$\triangleright$ left\\
10: \textbf{end if}\\
11: $R \gets \Delta P[:3, :3]$;\quad $s_y \gets \sqrt{R_{00}^2 + R_{10}^2}$\\
12: $(\phi, \psi) \gets \mathrm{EulerPitchYaw}(R, s_y)$ \hfill$\triangleright$ gimbal-safe\\
13: $\mathbf{o}_r \gets (0,0,0,0)$ \hfill$\triangleright$ (yaw+, yaw$-$, pitch+, pitch$-$)\\
14: \textbf{if} $\psi > +\tau_\mathrm{rot}$: $\mathbf{o}_r[1] \gets 1$ \quad\textbf{else if} $\psi < -\tau_\mathrm{rot}$: $\mathbf{o}_r[2] \gets 1$\\
15: \textbf{if} $\phi > +\tau_\mathrm{rot}$: $\mathbf{o}_r[3] \gets 1$ \quad\textbf{else if} $\phi < -\tau_\mathrm{rot}$: $\mathbf{o}_r[4] \gets 1$\\
16: $\ell_t \gets \mathrm{OneHotToLabel}(\mathbf{o}_t)$ \hfill$\triangleright$ translation label\\
17: $\ell_r \gets \mathrm{OneHotToLabel}(\mathbf{o}_r)$ \hfill$\triangleright$ rotation label\\
18: $\mathcal{S} \gets \mathrm{TransParts}(\ell_t) \cup \mathrm{RotateParts}(\ell_r)$\\
19: $\hat{a} \gets \bigoplus_{p \in \mathrm{sort}(\mathcal{S})} p$ \;($\oplus$ = ``+'');\; $\hat{a} \gets \texttt{no\_op}$ if $\mathcal{S} = \emptyset$\\
20: \textbf{return} $\hat{a}$
}}
\end{algorithm}

\begin{algorithm}[t]
\caption{Video-level action decoding and threshold selection.}
\label{alg:predict_sequence}
\centering
\small
\fbox{\parbox{0.95\columnwidth}{%
\textbf{Input:} $\hat{\mathcal{T}} = (\hat{P}_0, \ldots, \hat{P}_T)$ \hfill$\triangleright$ $T{+}1$ aligned camera poses\\
\quad $\mathcal{A} = (a_1, \ldots, a_T)$ \hfill$\triangleright$ $T$ GT action commands\\
\quad $K = 4$ \hfill$\triangleright$ action-window size\\
\quad $\Theta$ \hfill$\triangleright$ 4 fixed effective move candidates\\
\quad $\tau_\mathrm{rot} = 1.0^\circ$ \hfill$\triangleright$ effective rotation threshold\\
\textbf{Output:} $\hat{\mathcal{A}} = (\hat{a}_1, \ldots, \hat{a}_T)$ \hfill$\triangleright$ per-action predictions\\
\quad $\tau^*$, $\pi$ \hfill$\triangleright$ selected threshold \& sweep table\\[-2pt]
\rule{\linewidth}{0.4pt}\\[2pt]
1:\; \textbf{if} $T < K$ \textbf{then return} $\langle\texttt{no\_op}\rangle^T,\, \varnothing,\, \varnothing$\\
2:\; $C \gets \lfloor T/K\rfloor$ \hfill$\triangleright$ number of complete windows\\
3:\; \textbf{for} $c = 0, \ldots, C{-}1$ \textbf{do}\\
4:\; \quad $s \gets c \cdot K$;\; $e \gets s{+}K$\\
5:\; \quad $\Delta P_c \gets \hat{P}_s^{-1} \cdot \hat{P}_e$ \hfill$\triangleright$ relative pose across window $c$\\
6:\; \textbf{end for}\\
7:\; $(\hat{\mathcal{A}}^*,\, \alpha^*) \gets (\emptyset,\, -\infty)$\\
8:\; \textbf{for each} $\tau_\mathrm{move} \in \Theta$ \textbf{do}\\
9:\; \quad \textbf{for} $c = 0, \ldots, C{-}1$ \textbf{do}\\
10: \quad\quad $\hat{b}_c \gets \textsc{DiscretizeWindow}(\Delta P_c,\, \tau_\mathrm{move},\, \tau_\mathrm{rot})$\\
11: \quad \textbf{end for}\\
12: \quad $\hat{\mathcal{A}} \gets \mathrm{RepeatEach}(\langle \hat{b}_c \rangle,\, K)$\\
13: \quad \textbf{if} $|\hat{\mathcal{A}}| < T$ \textbf{then} $\hat{\mathcal{A}} \gets \hat{\mathcal{A}} \,\|\, \langle\hat{b}_{C-1}\rangle^{T-|\hat{\mathcal{A}}|}$\\
14: \quad $\alpha \gets \frac{1}{T}\sum_{i=1}^{T} \llbracket \hat{a}_i = a_i \rrbracket$ \hfill$\triangleright$ exact-match accuracy\\
15: \quad $\pi[\tau_\mathrm{move}] \gets \alpha$\\
16: \quad \textbf{if} $\alpha > \alpha^*$: $(\hat{\mathcal{A}}^*,\, \alpha^*,\, \tau^*) \gets (\hat{\mathcal{A}},\, \alpha,\, \tau_\mathrm{move})$\\
17: \textbf{end for}\\
18: $\mathrm{Acc}_\mathrm{strict} \gets \alpha^*$\\
19: $\mathrm{Acc}_\mathrm{partial} \gets \frac{1}{T}\sum_{i=1}^{T} \llbracket \mathcal{P}(\hat{a}_i^*) \cap \mathcal{P}(a_i) \neq \emptyset \rrbracket$\\
20: \textbf{return} $\hat{\mathcal{A}}^*,\, \tau^*,\, \pi$
}}
\end{algorithm}

\section{TrajScore: Detailed Computation}
\label{sec:appendix_trajscore}

\paragraph{Core idea.}
Above minimum-motion safeguards, the reference retains the model's displacement or rotation magnitude while imposing the ideal direction and shape prescribed by the action. It therefore answers: \emph{``Given that the model moved this far, what is the ideal trajectory shape?''} For a \texttt{forward} action, the reference is a straight line in the commanded direction with the same endpoint displacement magnitude as the prediction, but no lateral drift. This evaluates geometric deviations without requiring a fixed movement distance or turning angle across models.

\subsection{Adaptive Reference Construction}

Action changes partition $\mathcal{A}$ into constant-command segments. The trajectory evaluator uses half-open slices: commands $a_{s+1},\ldots,a_e$ define a segment whose scored poses are $\hat{P}_s,\ldots,\hat{P}_{e-1}$. Write $\hat{P}_j=(\hat{R}_j,\hat{\mathbf{p}}_j)$ for camera-to-world rotation and position.

\paragraph{Translation.}
For a pure movement command, let $\mathbf{u}_a$ be its unit direction in local camera coordinates and $R_\mathrm{world}$ the running reference orientation, initialized from the first scored pose. The reference is a straight line with direction $\mathbf{d}_a=R_\mathrm{world}\mathbf{u}_a$ and displacement
\begin{equation}
    d_\mathrm{ref}=
    \begin{cases}
        \|\hat{\mathbf{p}}_{e-1}-\hat{\mathbf{p}}_s\|_2,
        & \|\hat{\mathbf{p}}_{e-1}-\hat{\mathbf{p}}_s\|_2>0.1,\\
        1.0, & \text{otherwise}.
    \end{cases}
\end{equation}
For an isolated segment, its reference positions are
\begin{equation}
    \mathbf{p}^{\mathrm{ref}}_j
    =\hat{\mathbf{p}}_s+
    \frac{j-s}{\max(e-s-1,1)}\,d_\mathrm{ref}\mathbf{d}_a,
    \quad j=s,\ldots,e-1.
\end{equation}
The orientation remains fixed. The fallback displacement prevents a nearly static prediction from defining a zero-length target.

\paragraph{Rotation and compound actions.}
Look commands define in-place rotation in FPV and an orbit about the subject in TPV; compound commands combine this rotation with the prescribed movement. Reference rotation has the commanded direction and uses the predicted magnitude with a minimum of $30^\circ$. For TPV, subject depth determines the orbit radius: it is estimated when predicted accumulated rotation is at least $5^\circ$, otherwise set to $1.0$, with estimated depths also bounded below by $1.0$.

Rotating reference segments are aligned to the initial predicted heading through
$R_\mathrm{align}=\hat{R}_s(R^\mathrm{ref}_s)^\top$.
Their extent is matched using the ratio of mean radial extents,
$\bar{r}_\mathrm{pred}/\bar{r}_\mathrm{ref}$; this scale is reset to $1$ when predicted displacement or rotation falls below the minimum-motion thresholds.

\paragraph{Segment composition.}
Reference segments are translated to begin at the preceding reference endpoint. The running reference orientation is advanced after segments containing rotation. This constructs a continuous multi-action reference while preserving the commanded sequence.

\subsection{Segment-wise Alignment}

Two paths with the same geometry may traverse it at different speeds, producing apparent errors under frame-wise comparison. For segments with nonzero translational path length, we therefore match samples at equal fractions of the predicted and reference path lengths. Positions are linearly interpolated along cumulative arc length and orientations use spherical linear interpolation (Slerp).

For zero-length paths, including pure in-place rotation, we instead sample at uniformly spaced frame indices, retaining rotational timing.

We target $60$ evaluation pairs across $N_\mathrm{seg}$ scored segments. Each segment with more than two poses receives
\begin{equation}
    n_\mathrm{per}=\max\!\left(
    \left\lfloor\frac{60}{N_\mathrm{seg}}\right\rfloor,4\right)
\end{equation}
paired samples; shorter segments retain their available poses. This allocation normally yields $M=N_\mathrm{seg}n_\mathrm{per}$ pairs with equal segment weighting.

\subsection{Geometric Errors and Normalization}

For paired samples $(\hat{\mathbf{p}}_i,\hat{R}_i)$ and
$(\mathbf{p}^{\mathrm{ref}}_i,R^{\mathrm{ref}}_i)$, define
\begin{align}
    e_t^{(i)}&=\|\hat{\mathbf{p}}_i-\mathbf{p}^{\mathrm{ref}}_i\|_2,\\
    e_r^{(i)}&=\angle(\hat{R}_i,R^{\mathrm{ref}}_i).
\end{align}
The angle is the geodesic rotation difference in degrees. Averaging over all $M$ pairs gives $\mathrm{ATE}_t$ and $\mathrm{ATE}_r$.

For the same absolute error, a longer predicted path or greater accumulated rotation can yield a smaller normalized error. We therefore normalize by the predicted total path length $L_\mathrm{path}$ and accumulated rotation $\Phi_\mathrm{total}$, as in the main text, with denominator floors $\ell_\mathrm{min}=0.5$ and $\phi_\mathrm{min}=10^\circ$. Each normalized error is clipped to $[0,1]$ before their average is subtracted from $1$.

The denominator floors stabilize low-motion normalization, while reference-motion safeguards prevent trivial targets.

\subsection{Numerical Example}

Consider $300$ consecutive \texttt{forward} action intervals, forming one segment. The path advances mainly along $z$, with lateral and vertical drift. Its start and end positions are
\begin{align}
    \hat{\mathbf{p}}_\mathrm{start}&=(-0.001,\,-0.000,\,-0.000),\\
    \hat{\mathbf{p}}_\mathrm{end}&=(0.002,\,-0.245,\,6.509).
\end{align}
The endpoint displacement is $6.513$, the path length is $6.603$, and accumulated rotation is $2.72^\circ$. Since the displacement exceeds $0.1$, the reference is a straight line of length $6.513$ along the initial forward direction, approximately $(0,0,1)^\top$.

Resampling produces $60$ pairs, with mean errors $\mathrm{ATE}_t=0.090$ and $\mathrm{ATE}_r=0.41^\circ$. Therefore,
\begin{align}
    \mathrm{nATE}_t&=\frac{0.090}{\max(6.603,0.5)}\approx0.014,\\
    \mathrm{nATE}_r&=\frac{0.41}{\max(2.72,10)}=0.041,\\
    \mathrm{TrajScore}&=1-\frac{0.014+0.041}{2}\approx0.973.
\end{align}
The high score reflects a nearly straight forward path with small position and heading deviations. It does not imply exact timing or speed agreement. Temporal command agreement is assessed separately by the discrete action metrics.

\section{Visual Drift: Window and Stride Selection}
\label{sec:appendix_drift_sweep}

The drift metric of \cref{eq:drift} has two free parameters, the window duration $W$ and the stride $R$. This appendix reports the sweep that fixes $W = 3\,$s and $R = 1\,$s. All numbers are measured on $9{,}460$ evaluated rollouts covering the full duration range of our internal runs (from $15\,$s up to $3600\,$s, i.e.\ including rollouts far longer than the released $15$/$30$/$60\,$s buckets), so that both ends of the temporal scale are stressed. As a sensitivity reference we use the Spearman rank correlation $\eta$ between the windowed drift and the earlier fixed-$10$-segment variant on the $15\,$s bucket, where the two definitions are expected to agree most closely. With ten equal segments, segment duration increases from $1.5\,$s for a $15\,$s rollout to $6\,$s for a $60\,$s rollout, potentially averaging away brief collapses in longer rollouts. On the $15\,$s bucket, a window near $1.5\,$s closely reproduces the segment-based ranking; this correlation measures agreement between the two definitions.

\paragraph{Window duration.}
Sweeping $W \in \{1.5, 2, 3, 5\}\,$s gives $\eta = 0.992$, $0.991$, $0.977$ and $0.932$ respectively on the $15\,$s bucket. A $5\,$s window can dilute $1$--$2\,$s collapses through temporal averaging and yields lower rank agreement with the segment-based reference. At $W = 3\,$s, the correlation is $0.977$ and the mean drift is $0.142$, compared with $0.165$ for the reference. At the long end the choice matters much less---on the $3600\,$s bucket $W = 3\,$s yields $0.587$ against $0.554$ for $W = 5\,$s, a ${\sim}6\%$ difference with no qualitative change---so the decision is driven by the short-clip regime. We prefer $3\,$s over $2\,$s despite the latter's higher correlation ($\eta = 0.991$): at $16\,$fps, a $3\,$s window averages $48$ frames, reducing the contribution of isolated frame-level fluctuations while emphasizing \emph{sustained} degradation. $W = 3\,$s is the compromise between temporal resolution and robustness.

\paragraph{Stride.}
With $W$ at $3\,$s, disjoint windows ($R = W = 3\,$s; $5$ windows on a $15\,$s rollout) and sliding windows ($R = 1\,$s) agree closely in aggregate: mean drift differs by only $3$--$7\%$ and $\eta$ between the two is $0.987$--$0.991$. Despite this aggregate agreement, overlapping windows reduce sensitivity to window placement by mitigating the dilution of collapses split across disjoint-window boundaries. For example, on the most divergent sample the disjoint variant reports $0.468$ while the sliding variant reports $0.674$, a gap of $0.21$ caused purely by a collapse straddling a window boundary and being split across two neighbouring means. The effect is stronger for larger windows---in an earlier comparison at $W = 5\,$s, switching to a $1\,$s stride raised per-model mean drift by $6$--$27\%$ and lifted one boundary-straddling case from $0.0038$ to $0.0853$. Since the sliding window set is a superset of the disjoint one, the sliding drift is by construction never smaller. The cost is negligible: a $3600\,$s rollout produces ${\sim}3600$ windows, each a mean over a list of scalars, evaluated in milliseconds. We therefore keep both choices, as they address orthogonal problems: reducing $W$ from $5\,$s to $3\,$s fixes the temporal scale being too coarse for short rollouts, while overlapping windows prevent boundary-aligned collapses from being diluted in individual cases.


\section{Interaction Physics Evaluation Details}
\label{sec:appendix_physics_details}

\begin{figure}[t]
\centering
\includegraphics[width=\columnwidth]{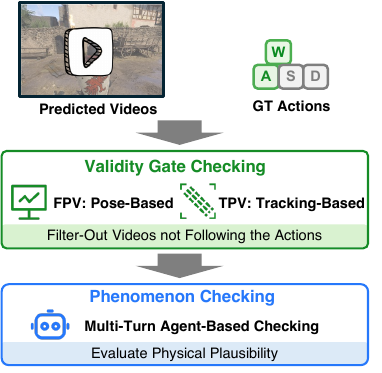}
\caption{\textbf{Detailed interaction-physics evaluation pipeline.} The full pipeline includes action-responsiveness validation, mechanics protocols, optics protocols, 3D-consistency evaluation, \vlmjudge{}-based VLM scoring, fallback queries, and domain-level aggregation.}
\label{fig:physics_pipeline_details}
\end{figure}

This appendix provides the implementation details omitted from the main interaction-physics section. The high-level goal remains to evaluate whether a generated rollout produces physically plausible consequences under user control. We first apply a validity gate to remove static or uncontrolled generations, and then evaluate valid rollouts along three complementary domains: mechanics, optics, and 3D consistency. Unless otherwise noted, all VLM queries in this appendix are instantiated with \vlmjudge{}. \Cref{fig:physics_pipeline_details} shows the complete implementation pipeline.

\subsection{Validity Gate Details}
\label{sec:appendix_validity_details}

\paragraph{Motivation.}
Before physics is evaluated, a validity gate checks whether the rollout provides usable motion evidence for the corresponding view. Invalid cases receive a score of $-1$ and are excluded from aggregate physics statistics.

\paragraph{FPV validity.}
For FPV videos, we use a pose-based pre-filter. The camera trajectory is reconstructed with the benchmark pose pipeline, and for each prescribed translation segment we measure the forward displacement of the camera in its local frame; a case is marked invalid when this displacement falls below a per-pose-source calibrated threshold, i.e., the camera fails to respond to the commanded motion. Compared with an optical-flow direction check, the pose-based gate remains reliable in texture-poor scenes where flow estimates degrade.

\paragraph{TPV validity.}
For TPV videos, optical flow cannot reliably distinguish camera motion from subject motion. We therefore adopt a subject-centric validity check based on SAM\,2~\cite{ravi2024sam2}, defined on tracking \emph{continuity} rather than on displacement from the initial pose: a correctly controlled protagonist is expected to move away from where it started, so we penalize only discontinuities such as disappearance, teleportation, tracking loss, and abrupt scale change. This tracking-continuity criterion provides usable subject evidence rather than, by itself, certifying action fidelity. It is also used for subject memory (Appendix~\ref{sec:appendix_subject_memory_details}), and the resulting video-level decisions are aggregated into $R_{\mathrm{ctrl}}$ for TPV leaderboard weighting (\cref{sec:scoring}).

Concretely, we track and segment the subject every $\Delta_{\mathrm{ctrl}}=20$ frames and extract an axis-aligned bounding box $B_{t_i}$ for each sampled frame. For each pair of adjacent sampled frames, we compute
\begin{equation}
\mathrm{IoU}_{t_i}
=
\frac{|B_{t_{i-1}} \cap B_{t_i}|}{|B_{t_{i-1}} \cup B_{t_i}|},
\qquad
i = 2,\ldots,N.
\end{equation}
A sampled frame in which no subject mask is found contributes zero IoU. For a video $V$, the tracking-continuity score $c(V)$ is the fraction of adjacent sampled-frame transitions satisfying $\mathrm{IoU}_{t_i}\geq\tau_{\mathrm{iou}}$:
\begin{equation}
c(V)
=
\frac{1}{N-1}
\sum_{i=2}^{N}
\mathbf{1}[\mathrm{IoU}_{t_i} \geq \tau_{\mathrm{iou}}].
\end{equation}
A video passes validity when $c(V) \geq \tau_{\mathrm{ctrl}}$. We set $\tau_{\mathrm{iou}}=0.4$ and $\tau_{\mathrm{ctrl}}=0.9$ throughout. In principle, continuity and commanded motion need not coincide; however, a per-segment speed audit with the pose-pipeline maintainer found no such divergence among the evaluated rollouts. When masks are unavailable, a lightweight \vlmjudge{} query over three frames checks the expected motion direction as a substitute action check, rather than as a continuity estimate.

\subsection{Physics Protocol Details}
\label{sec:appendix_physics_protocols}

\subsubsection{Mechanics Details}
\label{sec:appendix_mechanics_details}

Mechanics is evaluated through four protocol families: collision, clipping, deformation, and gravity. Ground-contact adherence on slopes and steps (terrain following) is scored within the gravity family but retains its own rubric (\Cref{sec:appendix_terrain_details}), matching the protocol taxonomy of \Cref{fig:horizon_dist}. Unless otherwise specified, each protocol returns a binary score, where $1$ indicates a physically plausible outcome and $0$ indicates failure.

\paragraph{Two evidence regimes.}
FPV and TPV rollouts expose different evidence. In FPV, the reconstructed camera trajectory is the primary signal: after the validity pre-filter, a bird's-eye-view protocol checks whether the trajectory exhibits the environmental influence expected under the prescribed action---lateral deviation toward the expected side, contact with a lateral barrier, or disturbance on either side---and pass candidates are re-checked by \vlmjudge{} for scene degradation before being accepted. In TPV, trajectory signals cannot localize subject--object events, so each criterion is judged by an evidence-gathering agent loop.

\paragraph{TPV agent loop.}
The agent receives the case objective (criterion plus scene description) together with a coarse overview timeline of the rollout. It then iterates for at most four rounds: localize the critical event window, call an inspection tool, and either gather more evidence or emit a structured verdict. Two tools are exposed: \texttt{inspect\_interval(start, end, fps, crop)}, which densely samples a time window at $1$--$8$\,fps, and \texttt{inspect\_frames(timestamps, crop)}, which retrieves exact frames; the crop argument (\texttt{full}, \texttt{center}, \texttt{lower}, \texttt{upper}) zooms into the interaction region at up to $768$\,px. The verdict is a structured object containing the label, a confidence, the event interval, and timestamped evidence observations, which makes every score auditable.

\paragraph{Per-criterion rubrics.}
Each criterion instantiates the same loop with a one-line evidence requirement: collision compares the target's position and pose before and after first contact; clipping tests whether the subject ever traverses a solid obstacle (body--obstacle interpenetration, the obstacle vanishing at the subject's position, or side-switching without a passage that visibly pre-exists contact); deformation compares the contacted surface before and after passage (persistent traces) or during contact (real-time ripples and crop parting); gravity requires a continuous descent or submersion response at edges and water entries without demanding free-fall signatures, and, for its terrain-adherence subfamily, continuous height and foot contact over slopes and steps.

\paragraph{In-loop evidence audit.}
Before a verdict is committed, the loop applies a final audit step: clipping pass verdicts are re-examined by an independent crossing reviewer that checks obstacle-topology continuity around the contact window, and fail verdicts on interaction criteria receive one high-recall positive-evidence pass that searches for subtle missed responses (faint traces, small splashes, partial displacement). The audit is part of the loop rather than a separate pipeline stage, so its evidence enters the same structured verdict.

\subsubsection{Optics Details}
\label{sec:appendix_optics_details}

Optics is evaluated through two protocols: reflection and occlusion, where occlusion is instantiated as shadow correctness. These protocols differ in score granularity.

\paragraph{Reflection.}
Reflection uses a two-tier evaluation. Tier~1 is a coarse global check. We uniformly sample $N=12$ frames at 384\,px resolution and ask the \vlmjudge{} VLM whether reflections are globally plausible: consistent in shape, approximately symmetric about the reflecting surface, and changing naturally with camera or subject motion. A failure at Tier~1 yields a score of $0$.

If Tier~1 passes, Tier~2 performs fine-grained per-frame scoring. We sample $N=8$ frames and pair each sampled frame with the first frame from the test case as reference. The \vlmjudge{} VLM assigns each frame an integer score:
\begin{equation}
s_t \in \{-1,\;0,\;1,\;2,\;\mathrm{N/A}\},
\end{equation}
where $2$ denotes a physically correct reflection, $1$ indicates partial errors, $0$ means the reflecting surface is visible but the reflection is missing or wrong, $-1$ marks a collapsed scene, and N/A indicates that the reflecting surface has left the field of view. Collapsed ($-1$) frames are excluded; N/A frames are instead scored by a surface-agnostic optical-quality query on the same $0$--$2$ scale (normal $=2$, degraded $=0$) and enter $\mathcal{V}$, so surface absence alone is not treated as a reflection failure. The reflection score is
\begin{equation}
S_{\mathrm{refl}}
=
\frac{1}{2|\mathcal{V}|}
\sum_{t\in\mathcal{V}} s_t,
\qquad
\mathcal{V}=\{t:s_t\geq 0\},
\end{equation}
yielding a continuous score in $[0,1]$. A per-case degradation re-check samples start, mid, and end frames and asks \vlmjudge{} whether the rollout has visually collapsed; a degraded verdict overrides the reflection score to $0$ irrespective of the tier outcomes.

\paragraph{Occlusion and shadow correctness.}
The occlusion protocol evaluates whether shadows appear and behave consistently with the scene lighting. All reported shadow cases are TPV and use $N=10$ samples from the 10--100\% range; the FPV configuration samples six frames from 40--100\% and prepends the first frame as a reference. The \vlmjudge{} VLM judges whether the expected shadow is present and whether its direction and shape are physically consistent with the light source. Scoring is binary.

\subsubsection{3D Consistency Details}
\label{sec:appendix_3d_consistency_details}

3D consistency evaluates whether the generated scene preserves rigid spatial structure under controlled camera motion. In the current benchmark, these cases are primarily instantiated in FPV settings with long, unobstructed trajectories and a constant input rate. This design makes geometric distortions, inconsistent depth ordering, and non-rigid scene deformation easier to detect with reduced confounding effects from collisions, deformation, or terrain changes. In implementation, propagated landmarks define per-region interest masks on reference crops; each sampled predicted frame is compared with its matched reference by a perceptual (LPIPS) distance within each region. A two-threshold calibration then maps per-region distances to scores: distances below a rejection threshold indicate retrievable mild distortion and are stretched into the upper score range, while distances in an intermediate collapse band score $0$. Regions above the upper threshold are excluded when correspondence to the reference is unreliable (e.g., reference content is no longer visible or appearance divergence is extreme), reducing the frame denominator rather than contributing zero. This exclusion is heuristic: high perceptual distance may also reflect disappearance or severe distortion, and validating it against ground-truth visibility remains future work. Matching uses rendered landmark crops and perceptual similarity rather than model trajectory scale; views too far from every reference enter the unreliable-correspondence regime. The resulting per-case scores are treated as a third physics domain when available.

\subsection{Prompt Design Principles}
\label{sec:appendix_physics_prompt_design}

Across all protocols, prompt construction follows four design principles derived from iterative calibration against human judgments.

\paragraph{Scene-grounded context.}
Each agent run begins from the scene description in the test-case metadata, giving the judge the expected scenario without imposing rigid subject assumptions (the subject may be a person, vehicle, or animal).

\paragraph{Evidence before verdict.}
Verdicts must cite timestamped observations collected through the inspection tools; a pass without positive visual evidence, or a fail without a violated-constraint observation, is not accepted. This replaces one-shot whole-video questions, which are sensitive to temporal aliasing in sparse uniform samples.

\paragraph{Adaptive resolution and windowing.}
The agent chooses both the temporal window and the spatial crop per query: coarse full-frame overviews for localization, dense cropped samples (up to $768$\,px) for the interaction region. Fine-grained criteria such as deformation therefore receive the resolution their evidence requires without paying it globally.

\paragraph{In-loop audit instead of retries.}
Rather than re-issuing the same query on a different frame range, the loop ends with a targeted audit (crossing review for clipping passes, high-recall positive-evidence search for interaction fails). The audit consumes the same structured evidence and its outcome is recorded in the verdict trace.

\subsection{Physics Score Details}
\label{sec:appendix_physics_score_details}

Per-case scores are aggregated into a physics composite. Because validity filtering is model dependent, the resulting domain scores are conditional on different accepted subsets; TPV Control Rate is a complementary overall coverage indicator, not a replacement for identical per-domain coverage. Let $\mathcal{C}_{\mathrm{mech}}$, $\mathcal{C}_{\mathrm{opt}}$, and $\mathcal{C}_{\mathrm{3d}}$ denote the sets of valid mechanics, optics, and 3D-consistency cases, respectively. The domain-level scores are
\begin{equation}
S_{\mathrm{mech}}
=
\frac{1}{|\mathcal{C}_{\mathrm{mech}}|}
\sum_{c\in\mathcal{C}_{\mathrm{mech}}} s_c,
\qquad
S_{\mathrm{opt}}
=
\frac{1}{|\mathcal{C}_{\mathrm{opt}}|}
\sum_{c\in\mathcal{C}_{\mathrm{opt}}} s_c,
\end{equation}
with $S_{\mathrm{3d}}$ defined analogously when $\mathcal{C}_{\mathrm{3d}}$ is available. Here, $s_c\in[0,1]$ is the per-case score. The overall physics score averages all available physics domains:
\begin{equation}
S_{\mathrm{physics}}
=
\frac{1}{|\mathcal{D}|}
\sum_{d\in\mathcal{D}} S_d,
\qquad
\mathcal{D}\subseteq\{\mathrm{mech},\mathrm{opt},\mathrm{3d}\}.
\end{equation}
The available domains are weighted equally because they test complementary aspects of physical plausibility: mechanics evaluates interaction-level consequences, optics evaluates light-level consistency, and 3D consistency evaluates spatial coherence.

\subsection{Collision Evaluation Details}
\label{sec:appendix_collision_details}

\paragraph{Evidence-localized verification.}
The agent localizes first contact between the subject and the target object, then compares the target's position and pose before and after that moment. Target identity and expected response come from the test-case metadata. The design deliberately avoids requiring the judge to witness the exact contact frame: displacement, tumbling, or deformation of the target after the localized contact window is sufficient evidence, which suppresses both false positives from rollouts that never reach the target and false negatives from brief responses.

\paragraph{Scoring.}
Binary: $\mathrm{score}{=}1$ if the target shows a physically plausible response and remains observable, else $0$; validity-gate failures are excluded with $\mathrm{score}{=}{-}1$.

\subsection{Collision Evaluation Prompt}
\label{sec:appendix_collision_prompt}

The collision agent instantiates the shared rubric, tool protocol, and structured verdict format of \Cref{fig:agent_prompts} with the criterion-specific evidence requirement described above.

\subsection{Clipping Evaluation Details}
\label{sec:appendix_clipping_details}

\paragraph{Crossing-centric detection.}
Clipping (pass-through of solid obstacles) is scored inversely: detecting a crossing yields $0$. The rubric defines a crossing as any of: body--obstacle interpenetration; the obstacle vanishing or being replaced at the subject's position; or the subject appearing on the far side of the obstacle without a passage that was visibly present before contact. Static subjects and subjects blocked at the obstacle are passes, which keeps the criterion purely geometric and independent of action responsiveness (handled by the validity gate).

\paragraph{In-loop crossing audit.}
Every pass verdict triggers the loop's final audit: an independent reviewer re-inspects the contact window and the post-contact segment for topology-change artifacts (obstacle disappearance, sudden side switches). Only audits that confirm a pre-existing passage or continuous same-side evidence keep the pass.

\paragraph{Scoring.}
$\mathrm{score}{=}0$ if a crossing is confirmed, else $1$; validity-gate failures yield $\mathrm{score}{=}{-}1$.

\subsection{Clipping Evaluation Prompt}
\label{sec:appendix_clipping_prompt}

The clipping agent instantiates the shared rubric, tool protocol, and structured verdict format of \Cref{fig:agent_prompts} with the criterion-specific evidence requirement described above.

\subsection{Deformation Evaluation Details}
\label{sec:appendix_deformation_details}

\paragraph{Before/after surface comparison.}
Deformation requires the right evidence window rather than a fixed sampling rule: the agent localizes when the subject contacts the deformable surface, then compares that surface before contact (reference frames) with frames after passage for persistent traces (footprints, tire tracks), or during contact for real-time responses (ripples, splashes, crop parting). Crops of the contact region at up to $768$\,px preserve the fine detail these effects live in.

\paragraph{Leniency with guards.}
Faint, partial, or short-lived responses count as passes, but illumination changes, background texture, and camera motion are explicitly excluded as evidence, which prevents texture hallucination from scoring as deformation.

\paragraph{Scoring.}
Binary: $\mathrm{score}{=}1$ if a trace or real-time response is evidenced, else $0$.

\subsection{Deformation Evaluation Prompt}
\label{sec:appendix_deformation_prompt}

The deformation agent instantiates the shared rubric, tool protocol, and structured verdict format of \Cref{fig:agent_prompts} with the criterion-specific evidence requirement described above.

\subsection{Gravity Evaluation Details}
\label{sec:appendix_gravity_details}

\paragraph{Descent-evidence rubric.}
The terrain-adherence subfamily (slopes, steps) is detailed in \Cref{sec:appendix_terrain_details}.

Gravity cases vary (falls from edges, sinking into water, projectiles), so the rubric states the expected descent signature per scenario and accepts continuous height decrease, progressive submersion, or arc-like motion as evidence; free-fall acceleration is not required, since third-person cameras rarely resolve it. Absence of any descent or submersion over the localized window yields failure.

\paragraph{Scoring.}
Binary: $\mathrm{score}{=}1$ if the descent response is evidenced, else $0$.

\subsection{Gravity Evaluation Prompt}
\label{sec:appendix_gravity_prompt}

The gravity agent instantiates the shared rubric, tool protocol, and structured verdict format of \Cref{fig:agent_prompts} with the criterion-specific evidence requirement described above.

\subsection{Terrain Following Evaluation Details}
\label{sec:appendix_terrain_details}

Terrain following is reported under the gravity family in \Cref{fig:horizon_dist}; we retain a separate rubric because its evidence (continuous ground contact over elevation changes) differs from free descent.

\paragraph{Ground-contact continuity.}
The agent localizes slope, step, or uneven segments and verifies that the subject's height and foot contact change continuously with the terrain, rather than floating above or sinking into the surface. Elevation changes themselves are natural and are not penalized; the failure signatures are hovering, ground-level camera collapse, and view-through-surface artifacts.

\paragraph{Scoring.}
Binary: $\mathrm{score}{=}1$ if ground contact remains plausible throughout the localized segment, else $0$.

\subsection{Terrain Following Evaluation Prompt}
\label{sec:appendix_terrain_prompt}

The terrain agent instantiates the shared rubric, tool protocol, and structured verdict format of \Cref{fig:agent_prompts} with the criterion-specific evidence requirement described above.

\subsection{Reflection Evaluation Details}
\label{sec:appendix_reflection_details}

\paragraph{Two-tier coarse-to-fine evaluation.}
Reflection is the only protocol producing a continuous score. The evaluation proceeds in two tiers:
\begin{itemize}
    \item \textbf{Tier 1 (coarse)}: 12 uniformly sampled frames are presented to the \vlmjudge{} VLM in a single call. It judges global reflection plausibility: are reflections consistent in shape, approximately symmetric about the reflecting surface, and changing naturally with camera movement? A ``no'' verdict short-circuits to a score of $0$.
    \item \textbf{Tier 2 (fine)}: 8 frames are sampled and each is scored independently (parallelized with 4 workers) against the first-frame reference. The \vlmjudge{} VLM assigns a per-frame score of $-1$ (invalid), $0$ (missing/wrong), $1$ (physics errors), $2$ (correct), or N/A (surface not visible).
\end{itemize}

\paragraph{Per-frame scoring rubric.}
The Tier~2 prompt assigns N/A when the reflecting surface has moved out of view, rather than $0$ for an incorrect reflection on a visible surface. Before aggregation, N/A frames receive surface-agnostic optical-quality scores on the same $0$--$2$ scale (normal $=2$, degraded $=0$) and are included in $\mathcal{V}$.

\paragraph{Aggregation.}
The final score is:
\begin{equation}
S_{\mathrm{refl}} = \frac{1}{2 \cdot |\mathcal{V}|} \sum_{i \in \mathcal{V}} s_i, \qquad \mathcal{V} = \{i : s_i \geq 0\},
\end{equation}
where $s_i \in \{0, 1, 2\}$ are per-frame scores; collapsed ($-1$) frames are excluded, while N/A frames contribute surface-agnostic optical-quality scores on the same scale. The result is a continuous value in $[0, 1]$.

\paragraph{Degradation re-check (v2).}
A per-case re-check samples start, mid, and end frames and asks \vlmjudge{} whether the rollout has visually collapsed (black frames, heavy artifacts, nonsensical imagery). A degraded verdict overrides the tiered reflection score to $0$, preventing collapsed rollouts from collecting partial credit on the few frames that remain interpretable.

\paragraph{Surface-aware prompting.}
The reflecting-surface type (\texttt{water\_puddle}, \texttt{water}, \texttt{canal}, \texttt{mirror}, \texttt{glass}) is injected into both tiers so that the judge applies the specular-reflection expectation appropriate to that surface.

\subsection{Reflection Evaluation Prompt}
\label{sec:appendix_reflection_prompt}

The reflection prompts implement the coarse global plausibility check and the per-frame scoring rubric; both are shown in \Cref{fig:reflection_prompts}.

\subsection{Occlusion / Shadow Evaluation Details}
\label{sec:appendix_occlusion_details}

\paragraph{Temporal sampling with reference frame.}
Shadow evaluation uses view-specific temporal ranges: FPV samples six frames from $40$--$100\%$ and prepends the first frame as a visual reference, whereas TPV samples ten frames from $10$--$100\%$. All shadow cases reported here are TPV and use the latter configuration.

\paragraph{Strict shadow recognition criteria.}
The prompt enforces strict criteria to distinguish real shadows from ambient darkening:
\begin{itemize}
    \item A real shadow must have a \emph{recognizable silhouette} matching the casting object (e.g., a person shadow looks like a person outline).
    \item The shadow must have a \emph{clear boundary} separating shadowed from lit regions---vague dark patches, ambient shading, surface texture, or lighting gradients do \emph{not} count.
    \item The shadow direction must be physically consistent with visible light sources.
\end{itemize}
The \vlmjudge{} VLM is instructed to answer ``yes'' \emph{only} if confident the shadow is real and correct. This deliberately strict, visibility-oriented rubric favors unambiguous attributable shadows; physically plausible soft shadows from extended or scattered illumination may not satisfy it, so these criteria are evaluation anchors rather than universal requirements on shadow appearance.

\paragraph{Expected-shadow injection.}
When \texttt{eval\_protocol\_args.expected\_shadow} is provided (e.g., ``wall shadow,'' ``person shadow on the ground''), it is injected into the prompt to specify what the \vlmjudge{} VLM should look for. Without it, a generic shadow detection prompt is used.

\paragraph{Scoring.}
Binary: $\mathrm{score}{=}1$ if the expected shadow is present and behaves correctly, else $0$.

\subsection{Occlusion / Shadow Evaluation Prompt}
\label{sec:appendix_occlusion_prompt}

The occlusion/shadow prompts include expected-shadow checks, third-person fallback checks, and a generic fallback for scenes without explicit shadow metadata, as shown in \Cref{fig:occlusion_shadow_prompts}.

\section{Memory Evaluation Details}
\label{sec:appendix_memory_details}

\begin{figure*}[t]
\centering
\includegraphics[width=\textwidth]{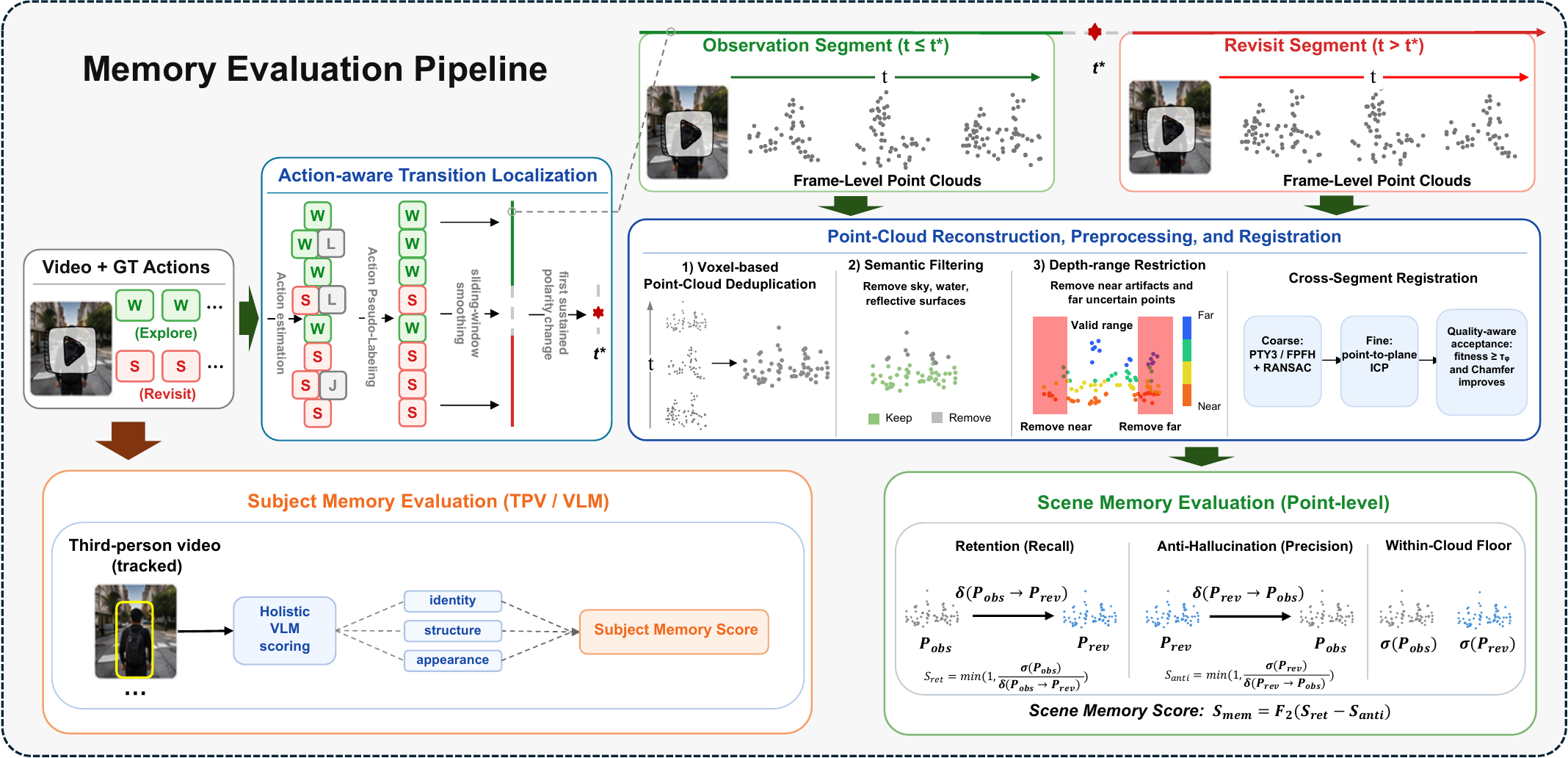}
\caption{\textbf{Detailed memory evaluation pipeline.} The full pipeline includes action-aware transition localization, point-cloud reconstruction, registration, filtering, scene-memory scoring, subject tracking, controllability gating, and \vlmjudge{}-based VLM subject-memory evaluation.}
\label{fig:memory_pipeline_details}
\end{figure*}

This appendix provides the implementation details omitted from the main memory section. Scene memory aggregates geometry over observation and revisit segments, while subject memory evaluates whether the protagonist is preserved in controllable TPV rollouts. \Cref{fig:memory_pipeline_details} shows the complete implementation pipeline.

\subsection{Scene Memory Details}
\label{sec:appendix_scene_memory_details}

\paragraph{Action-aware transition localization.}
We localize the observation--revisit transition from the generated video rather than using the symmetric GT index. Given the video and its GT action sequence, CompassReward first estimates a frame-level action for every generated frame. Since the estimator may output compound actions, such as \texttt{W+L}, we convert predictions into action pseudo-labels in the GT action space. Specifically, for each predicted action, we search the set of GT actions occurring anywhere in the rollout and assign the first GT action whose components are all contained in the prediction; unmatched predictions receive no pseudo-label. The resulting pseudo-label sequence is then temporally smoothed, and the transition point $t^{*}$ is identified as the first stable change point from the observation action to the revisit action. This procedure aligns the memory split with the model's executed action transition and reduces contamination from compound or noisy action predictions.

\paragraph{Point-cloud reconstruction and preprocessing.}
From the observation segment $\{I_t\}_{t \le t^{*}}$ and revisit segment $\{I_t\}_{t > t^{*}}$, subsampled at a fixed frame stride, we reconstruct point clouds $\mathcal{P}_{\mathrm{obs}}$ and $\mathcal{P}_{\mathrm{rev}}$ using monocular depth estimates and camera intrinsics. To reduce depth-estimation noise, particularly in distant and textureless regions, we apply the following preprocessing steps.

\noindent\emph{Voxel-based deduplication.} We average points within each voxel of an adaptive grid, reducing redundancy while preserving scene structure.

\noindent\emph{Semantic filtering.} We use a pretrained SegFormer to remove categories with unreliable depth, such as sky, water, and reflective surfaces.

\noindent\emph{Depth-range restriction.} We retain only points whose estimated depth falls inside a scene-dependent absolute range, discarding both near-field artifacts and distant regions with high depth uncertainty.

\noindent\emph{Subject removal (TPV only).} In third-person rollouts the protagonist is itself a large moving object, so its geometry would appear in one segment and not the other and be charged as scene inconsistency. We therefore propagate the annotated subject mask with SAM\,2 and exclude the tracked subject from both segment clouds, leaving only the static scene. This is the sole difference between the FPV and TPV scene-memory pipelines.

\paragraph{Cross-segment point-cloud registration.}
The coordinates of $\mathcal{P}_{\mathrm{obs}}$ and $\mathcal{P}_{\mathrm{rev}}$ are determined by the estimated camera-pose chain. Pose drift can introduce a rigid displacement unrelated to memory, artificially increasing both apparent forgetting and hallucination. We therefore register $\mathcal{P}_{\mathrm{rev}}$ to the stationary reference $\mathcal{P}_{\mathrm{obs}}$ using coarse-to-fine alignment. We estimate the transformation once from semantically filtered, scene-range-restricted, voxel-deduplicated clouds and reuse it for the scoring clouds.

\noindent \textit{Coarse registration.}\quad
When the initial misalignment, measured as the maximum of the inter-centroid displacement and symmetric Chamfer distance, exceeds $\tau_c$, we perform global registration. We compute per-point descriptors using Point Transformer V3~\cite{wu2024ptv3} features. Nearest-neighbor matching in descriptor space produces candidate correspondences, from which RANSAC estimates a geometrically consistent rigid transformation.

\noindent \textit{Fine registration.}\quad
Starting from the coarse estimate, or the identity transformation when global registration is unnecessary, we refine the alignment using point-to-plane Iterative Closest Point (ICP)~\cite{chen1992icp}.

\noindent \textit{Quality-aware acceptance.}\quad
To reject degenerate registrations, we apply a dual acceptance criterion. Let $T^{*}$ be the estimated transformation, $\phi(T)$ the ICP fitness, and $\mathrm{CD}(T)$ the symmetric Chamfer distance after applying $T$. We use
\begin{equation}
T =
\begin{cases}
T^{*}, & \text{if } \phi(T^{*}) \geq \tau_{\phi} \;\text{ and }\; \mathrm{CD}(T^{*}) \leq (1+\epsilon)\,\mathrm{CD}(I), \\[4pt]
I, & \text{otherwise},
\end{cases}
\end{equation}
where $\tau_{\phi}$ is the minimum acceptable fitness and $I$ is the $4{\times}4$ identity. Thus, registration is accepted only if it achieves sufficient fitness without increasing the Chamfer distance by more than a small relative tolerance $\epsilon$. All subsequent metrics are computed on the aligned pair $(\mathcal{P}_{\mathrm{obs}},\; T \circ \mathcal{P}_{\mathrm{rev}})$; for brevity, we use $\mathcal{P}_{\mathrm{rev}}$ to denote the aligned revisit cloud below.

\paragraph{Density-normalized retention and hallucination.}
We quantify geometric memory from nearest-neighbor distances between the aligned point clouds, normalized by the sampling resolution of the clouds themselves rather than by an absolute distance threshold. Writing $d(\mathbf{p},\mathcal{Y})$ for the distance from $\mathbf{p}$ to its nearest neighbor in $\mathcal{Y}$, the sampling floor $\sigma(\mathcal{X})$ of \cref{eq:sampling_floor} is estimated by drawing two \emph{disjoint} random subsets $\mathcal{X}_1,\mathcal{X}_2\subset\mathcal{X}$ of at most $N_{\mathrm{split}}$ points each under a fixed seed, and averaging $d(\mathbf{p},\mathcal{X}_2)$ over $\mathbf{p}\in\mathcal{X}_1$. Because the two halves are disjoint, no point can match itself, so $\sigma$ measures the characteristic spacing between independent samples of the same surface. The directed cross-segment term $\delta$ of \cref{eq:cross_discrepancy} is computed on subsamples of at most $N_{\max}$ points. The two scores are the clamped ratios of \cref{eq:mem_ratios}, combined by the harmonic mean of \cref{eq:scene_memory}. By analogy with precision and recall, the direction $\mathcal{P}_{\mathrm{obs}}\!\rightarrow\!\mathcal{P}_{\mathrm{rev}}$ asks how closely the geometry that should have been remembered is recovered and hence yields the recall term $S_{\mathrm{ret}}$, while the direction $\mathcal{P}_{\mathrm{rev}}\!\rightarrow\!\mathcal{P}_{\mathrm{obs}}$ asks how consistently the produced geometry is supported by prior observation and hence yields the precision term $S_{\mathrm{anti}}$. The results table reports $S_{\mathrm{anti}}$ directly, so that all three memory columns share the same orientation and the reported sub-scores are exactly the terms entering the harmonic mean. A clamped value of $1$ indicates that the cross-segment discrepancy has fallen to the sampling floor, i.e.\ the revisit is indistinguishable from a re-measurement of the same geometry at the achieved reconstruction resolution.

\subsection{Subject Memory Details}
\label{sec:appendix_subject_memory_details}

\paragraph{Third-person controllability filtering.}
Before evaluating subject memory, we filter out test cases with insufficient subject-tracking continuity. When control degenerates, the camera collapses toward a first-person view and produces characteristic artifacts: the subject grows or shrinks under translation actions (W/S) and disappears then reappears under yaw rotation actions (J/L).

To assess tracking continuity, we use SAM\,2~\cite{ravi2024sam2} to track and segment the subject every $\Delta_{\mathrm{ctrl}}=20$ frames and extract an axis-aligned bounding box $B_{t_i}$ for each sampled frame. For each pair of adjacent sampled frames, we compute
\begin{equation}
\mathrm{IoU}_{t_i} = \frac{|B_{t_{i-1}} \cap B_{t_i}|}{|B_{t_{i-1}} \cup B_{t_i}|}, \qquad i=2,\ldots,N.
\end{equation}
A sampled frame in which no subject mask is found contributes zero IoU. A rollout passes the gate when at least $90\%$ of the $N-1$ adjacent sampled-frame transitions satisfy $\mathrm{IoU}_{t_i} \geq \tau_{\mathrm{iou}} = 0.4$; cases that fail are excluded from subject-memory evaluation.

\paragraph{Failure modes.}
We characterize subject memory failure using five failure modes and one benign factor: identity change, structural distortion, appearance drift, subject disappearance, quality degradation, and benign viewpoint change. The evaluator must distinguish true subject-memory failure from benign variation caused by viewpoint, illumination, or shadow.

\paragraph{Sampling and subject extraction.}
Given a generated third-person video $V=\{I_1,I_2,\ldots,I_T\}$, we sample frames uniformly with stride $\Delta=20$, yielding indices $\mathcal{S}=\{t_1,t_2,\ldots,t_N\}$, where $t_1=1$. Each test case provides a first-frame subject mask, which initializes SAM\,2~\cite{ravi2024sam2} for mask propagation over the sampled sequence. Frames without valid subject tracking are excluded. We extract each retained subject observation as
\begin{equation}
O_{t_i} = \mathrm{crop}\!\left(I_{t_i} \odot M_{t_i}\right), \qquad t_i \in \mathcal{S},
\end{equation}
where $\odot$ denotes element-wise masking and $\mathrm{crop}(\cdot)$ isolates the masked subject region. Suppressing the background prevents the judge from using scene content as an identity cue.

\paragraph{Holistic VLM evaluation.}
We present the temporally ordered sequence $\mathcal{O}=\{O_{t_1},\ldots,O_{t_N}\}$ to the \vlmjudge{} VLM in a \emph{single} deterministic request, with the first retained crop acting as the reference and all remaining crops following in temporal order. The judge evaluates the sequence holistically rather than through pairwise frame comparisons, producing an integer subject memory score $S_{\mathrm{VLM}}\in\{1,\ldots,10\}$ together with binary diagnostic flags. The full prompt is provided in Appendix~\ref{sec:appendix_subject_memory_prompt}.

We linearly normalize the integer rating to $[0,1]$:
\begin{equation}
S_{\mathrm{sub}} = \frac{S_{\mathrm{VLM}} - 1}{9}.
\end{equation}
Thus, a rating of $1$ maps to $0$, and a rating of $10$ maps to $1$. Holistic scoring accommodates smooth viewpoint and illumination changes that can confound frame-level comparisons, while producing a single benchmark-compatible scalar without requiring an external reference-feature library.

\section{Subject Memory Prompt}
\label{sec:appendix_subject_memory_prompt}

For third-person memory evaluation, the Subject Memory Evaluation Prompt in \Cref{fig:subject_memory_prompt} scores video-level subject memory from temporally ordered subject crops.
The prompt emphasizes holistic temporal judgment, separates viewpoint and lighting changes from true identity drift, and explicitly penalizes structural collapse, disappearance, and severe quality degradation.

\section{Why Not Frame-pair Memory?}
\label{sec:frame_pair_memory}

\begin{figure}[t]
\centering
\includegraphics[width=\columnwidth]{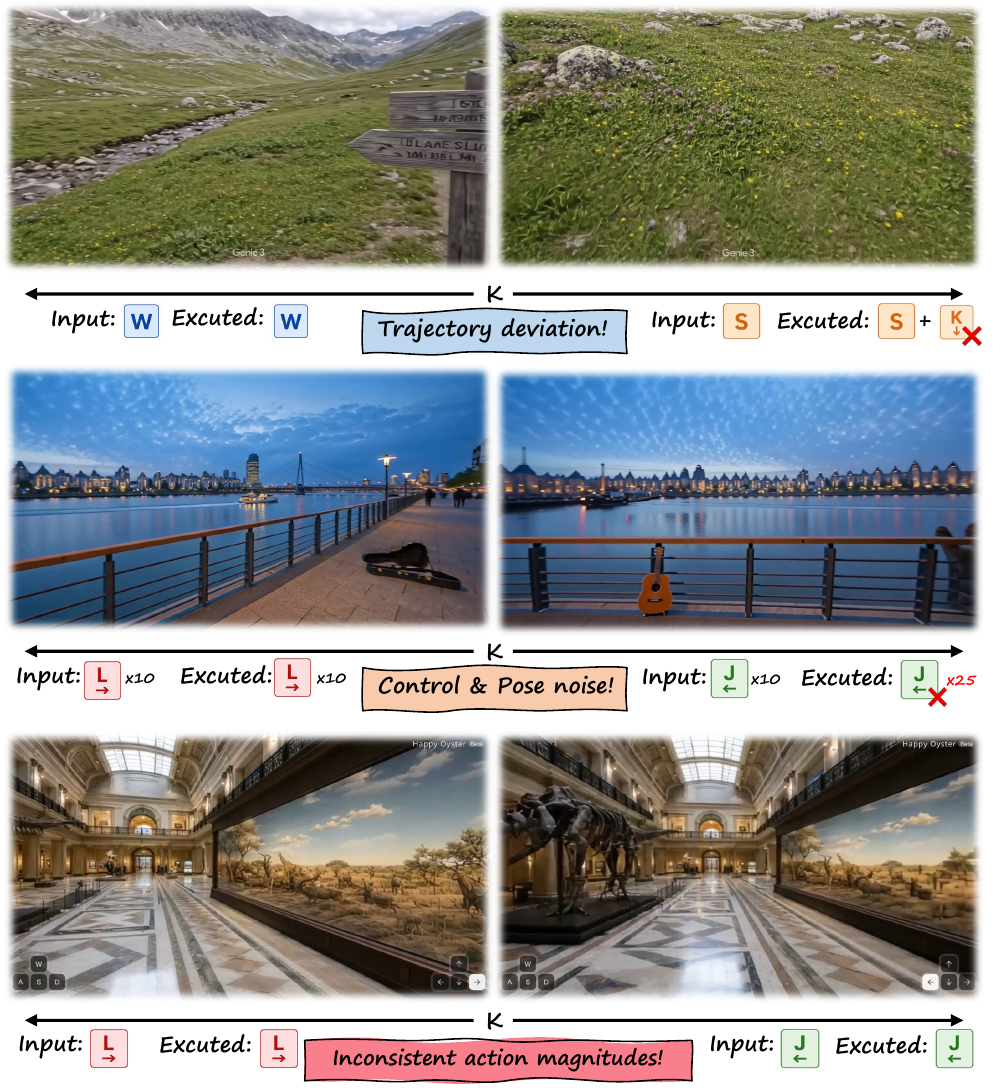}
\caption{\textbf{Failure mode of frame-pair memory evaluation.}
Frame-pair metrics assume that an observation frame and its revisit counterpart depict the same spatial location.
In long-horizon interactive rollouts, however, imperfect action execution can shift the revisit trajectory, so the paired frames correspond to different viewpoints or even different scene regions.
Image-level discrepancies in such pairs therefore reflect a mixture of memory degradation, action-following error, and pose-estimation noise, making frame-pair scores unreliable indicators of scene memory.}
\label{fig:frame_pair_failure}
\end{figure}

A natural alternative to scene-level memory is to directly compare observation and revisit frames at corresponding spatial locations. We instantiate this frame-pair approach by first reusing the action-aware transition localization step from Section~\ref{sec:scene_memory} to obtain the transition frame $t^{*}$, and then constructing geometrically corresponding pairs around $t^{*}$ via \emph{pose-aware frame pair construction}, described below. Even with $t^{*}$ given by the same procedure that anchors our scene-level pipeline, frame-pair evaluation remains insufficient for the reasons we explain at the end of this appendix. \Cref{fig:frame_pair_failure} visualizes a representative failure case.

\begin{figure*}[t]
\centering
\includegraphics[width=0.95\textwidth]{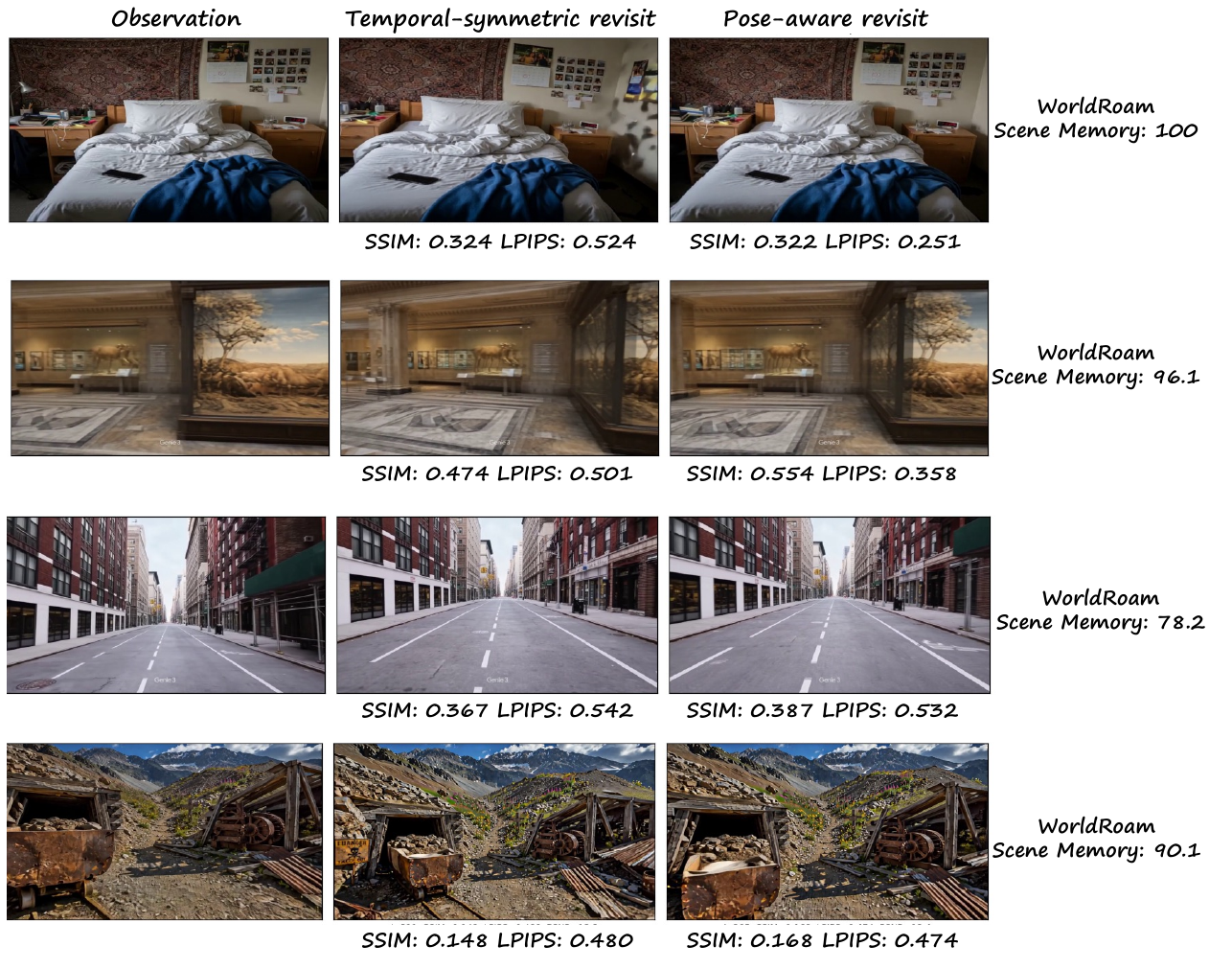}
\caption{\textbf{Frame-pair similarity can underestimate preserved scene memory.}
Each row shows the observation frame, its temporally symmetric revisit $\hat{f}_i$, and the pose-aware revisit $f_i^{*}$, with SSIM ($\uparrow$) / LPIPS ($\downarrow$) below and $S_{\mathrm{scene}}$ at right. Pose-aware pairing generally improves over symmetric pairing in these examples, illustrating how spatial misalignment affects frame-pair scores; yet substantial image-level differences remain (SSIM $\le 0.554$, LPIPS $\ge 0.251$) despite visibly preserved content. These examples suggest frame-pair discrepancies need not indicate scene-memory loss; the relatively high $S_{\mathrm{scene}}$ values are consistent with visibly preserved scene structure.}
\label{fig:frame_pair_grid}
\vspace{\baselineskip}
\end{figure*}

\paragraph{Pose-aware frame pair construction.}
Rather than pairing frames by symmetric index around $t^{*}$, camera poses estimated by the pose estimation pipeline are used to identify geometrically corresponding revisit frames. For each observation frame $f_i$, let $\hat{f}_i$ denote its temporally symmetric counterpart with respect to $t^{*}$. The revisit frame is selected by minimizing the geodesic distance to the pose of $f_i$ within a local temporal window:
\begin{equation}
f_i^{*}
=
\arg\min_{f_j \in \mathcal{N}(\hat{f}_i)}
d\bigl(P(f_i),\, P(f_j)\bigr),
\end{equation}
where $P(\cdot)$ denotes the estimated camera pose and $d(\cdot,\cdot)$ is a pose distance metric combining translational and rotational components.

This construction offers two advantages: (i)~evaluating memory on the most geometrically aligned pair minimizes contamination from incomplete revisits, and (ii)~restricting the search to a neighborhood of the symmetric index preserves the trajectory prior and mitigates the effect of noisy pose estimates.

Given the constructed pairs $\{(f_i,\, f_i^{*})\}$, standard image-level fidelity metrics are reported: MSE, PSNR, SSIM, and LPIPS.

\paragraph{Limitations of frame-pair memory evaluation.}
Despite the above mitigations, frame-pair metrics remain unreliable indicators of memory capability under the action-following quality exhibited by current world models. Three fundamental issues persist.

\noindent\emph{Systematic spatial offset from inconsistent action magnitudes.} Even when the transition frame $t^{*}$ is correctly identified, forward and backward actions often traverse different distances. This creates a systematic spatial displacement between the observation and revisit segments that no temporal re-indexing can correct. Pose-aware pairing alleviates this issue to some degree, but the search is limited to a local neighborhood and cannot compensate for large cumulative offsets.

\noindent\emph{Trajectory deviation from spurious motion.} Certain models exhibit unprompted camera movements that cause the generated trajectory to diverge substantially from the prescribed path. In such cases, the revisit segment may explore entirely different regions of the scene, rendering frame-level comparison unreliable regardless of the pairing strategy.

\noindent\emph{Compounding of control jitter and pose estimation noise.} For closed-source models accessed through web interfaces, network latency introduces temporal uncertainty in action execution, while the pose estimator itself contributes additional noise. These two error sources compound: noisy poses lead to suboptimal frame pair selection, and control jitter ensures that even correctly selected pairs depict slightly different viewpoints, both inflating the measured dissimilarity.

As a result, models that visibly preserve scene content upon human inspection can receive misleadingly poor frame-pair memory scores, not because they fail to remember, but because the evaluation framework conflates action-following deficiency with memory degradation. For these reasons, \ours{} does not adopt frame-pair memory and instead relies on the scene-level memory evaluation described in the main text, which aggregates geometric information across frames and is substantially more robust to per-frame action imprecision. \Cref{fig:frame_pair_grid} illustrates two limitations: temporally symmetric pairs can differ in viewpoint or location, and even after pose-aware pairing reduces some spatial offset, SSIM and LPIPS still indicate substantial image-level differences in examples with visibly preserved scene content.

\begin{figure*}
\centering
\includegraphics[width=0.95\textwidth]{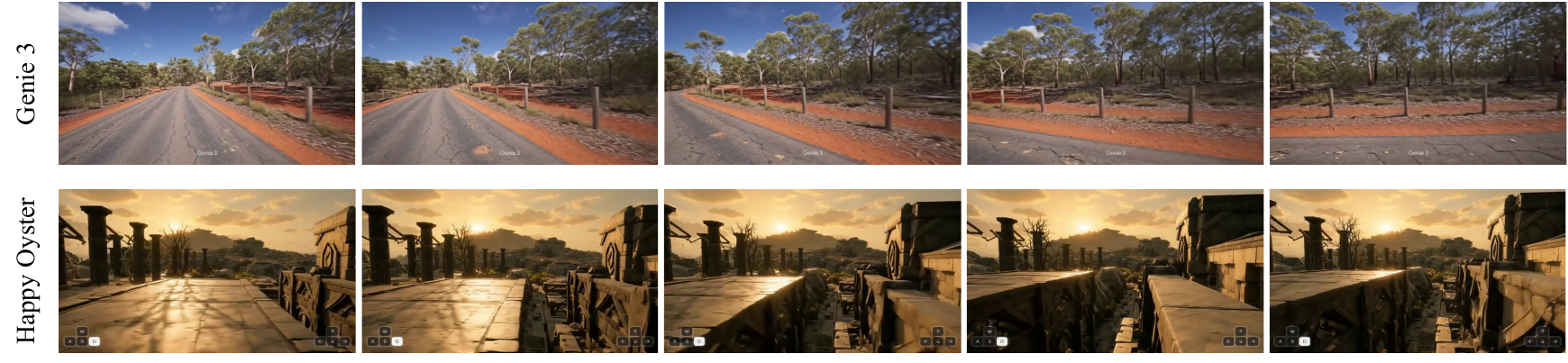}
\caption{\textbf{Action gap in closed-source models.}
\textbf{Top:} Genie~3. When the \texttt{D} key (rightward translation) is issued, the model rotates the camera to the right instead of translating the character rightward within a stable scene. The character remains centered while the entire scene rotates, making controlled navigation impossible.
\textbf{Bottom:} Happy Oyster. When the \texttt{D} key (rightward translation) is issued, the model faithfully enforces collision with a newly generated obstacle, preventing the character from moving rightward. This strong adherence to physics results in lower action-following scores.
Each row shows frames from left to right in temporal order.}
\label{fig:action_control_failures}
\end{figure*}

\section{Action Gap in Closed-Source Models}
\label{app:action_control_failures}

\Cref{tab:main_results} shows that Genie~3 and Happy Oyster trail the strongest open models in action-following scores despite their strong overall performance. We also present representative failure cases from both closed-source models to illustrate the underlying causes.

\subsection{Genie~3: Lateral-Action Camera Drift}

When the \texttt{A} or \texttt{D} key is pressed to move the character laterally, Genie~3 interprets the input as a camera rotation command rather than producing a lateral translation. As shown in \cref{fig:action_control_failures}~(top), instead of the character shifting left or right within a stable scene, the entire camera view rotates, leaving the character centered but drastically altering the visible environment. This spurious rotation makes precise steering impossible and directly illustrates a case-level action-following failure.

\subsection{Happy Oyster: Physics-Induced Action Failure}

In the illustrated rollout, Happy Oyster prioritizes collision response over commanded motion, which hurts its action-following score. As shown in \cref{fig:action_control_failures}~(bottom), when the autoregressive generation produces new obstacles along the navigation path, the model simulates collision rather than allowing the character to pass through. Once the character collides with a generated obstacle, subsequent movement actions cannot dislodge it because the collision response overrides the commanded input. This case illustrates a local trade-off between physical coherence and action following: the character remains trapped even as movement commands continue.

\section{Qualitative Study and Human Verification}
\label{sec:appendix_qualitative_human}

This appendix complements the quantitative leaderboard with a qualitative tour of \ours{} scores and a sanity check that the \vlmjudge{}-based judgments used by our physics and memory pipelines remain aligned with human perception.

\subsection{Qualitative Study}
\label{sec:appendix_qualitative_study}

\begin{figure*}[p]
\centering
\includegraphics[width=0.90\textwidth]{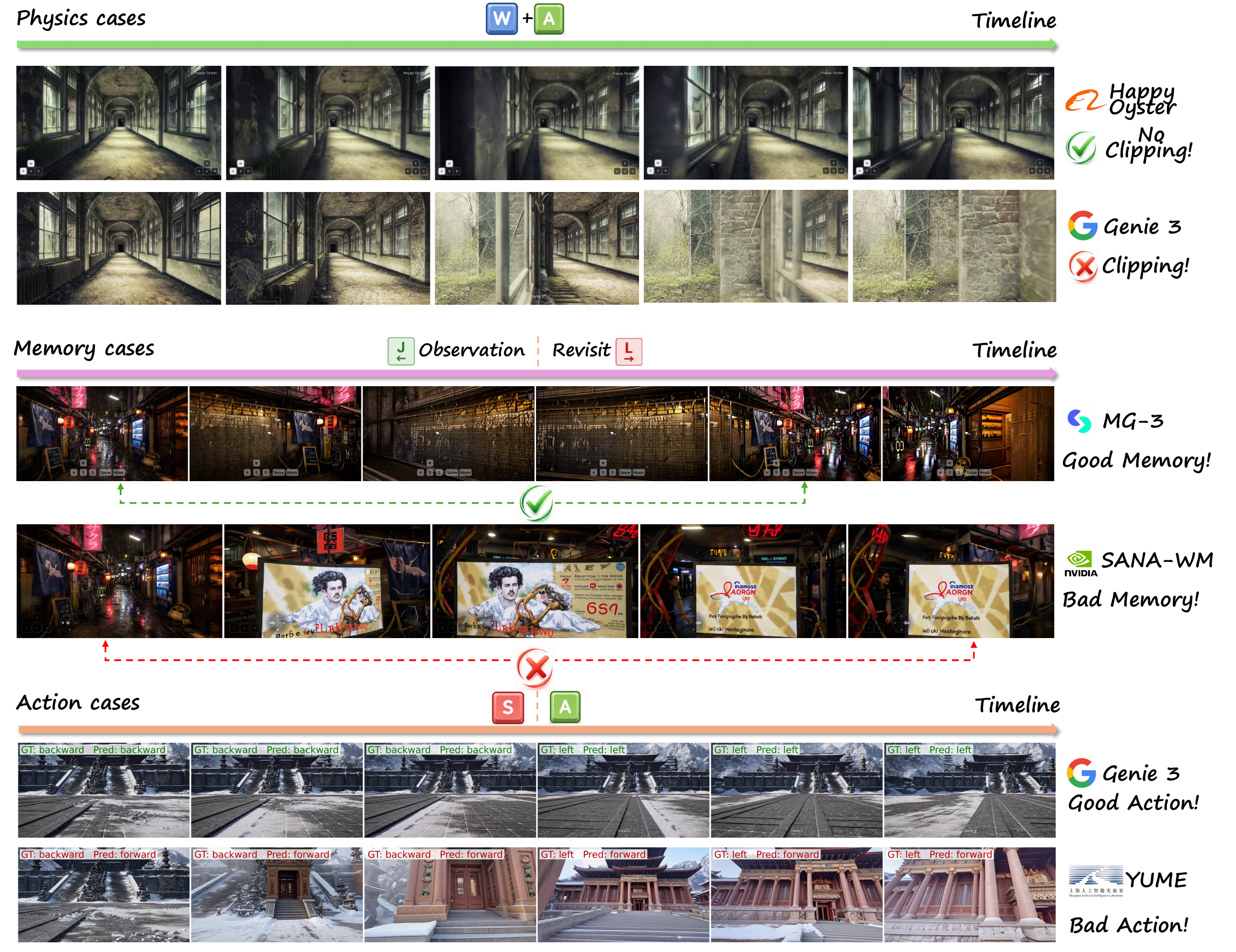}
\caption{\textbf{Qualitative examples across physics, memory, and action.}
Each block contrasts a positive and a negative rollout that share the same initial frame and action schedule.
\textbf{Physics:} $\mathtt{W}\!+\!\mathtt{A}$ in a corridor---Happy~Oyster (\emph{No Clipping}) vs.\ Genie~3 (\emph{Clipping}).
\textbf{Memory:} $\mathtt{J}\!\to\!\mathtt{L}$ Observation/Revisit in an alley---Matrix-Game~3.0 (\emph{Good Memory}) vs.\ SANA-WM (\emph{Bad Memory}).
\textbf{Action:} $\mathtt{S}\!\to\!\mathtt{A}$ (\texttt{backward}\,$\to$\,\texttt{left})---Genie~3 (\emph{Good Action}) vs.\ Yume~1.5 (\emph{Bad Action}).
Frames are shown in temporal order, left to right.}
\label{fig:qualitative_scores}
\end{figure*}

\Cref{fig:qualitative_scores} contrasts a positive and a negative rollout for each of the three dimensions---interaction physics, memory, and action following; the two rollouts in each block share both the initial frame and the keyboard action schedule, supporting comparison within the evaluated dimension.

\textbf{Physics.} Under a sustained $\mathtt{W}\!+\!\mathtt{A}$ in a narrow corridor, Happy~Oyster keeps the avatar inside the geometry (\emph{No Clipping}), whereas Genie~3 penetrates the wall into vegetation (\emph{Clipping}).
\textbf{Memory.} Under $\mathtt{J}$ (Observation) $\to$ $\mathtt{L}$ (Revisit) along a neon alley, Matrix-Game~3.0 reproduces the same storefronts on revisit (\emph{Good Memory}), while SANA-WM regenerates different murals and shopfronts (\emph{Bad Memory}).
\textbf{Action.} Under an $\mathtt{S}\!\to\!\mathtt{A}$ schedule (\texttt{backward}\,$\to$\,\texttt{left}), Genie~3's per-frame predictions match the ground truth throughout (\emph{Good Action}), while Yume~1.5 outputs \texttt{forward} in every segment (\emph{Bad Action}).

\subsection{Human Verification of VLM Judgments}
\label{sec:appendix_human_verification}

Both the interaction-physics pipeline (\Cref{sec:appendix_physics_details}) and the subject-memory pipeline (\Cref{sec:appendix_subject_memory_details}) delegate the final perceptual judgment to \vlmjudge{}. We conduct two compact human studies---one for the binary mechanics protocols and one for the ordinal subject-memory protocol---to verify that the VLM tracks human raters on these exact decisions.

\subsubsection{Physics VLM Verification}
\label{sec:appendix_physics_vlm_verification}

\begin{figure*}[p]
\centering
\includegraphics[width=0.90\textwidth]{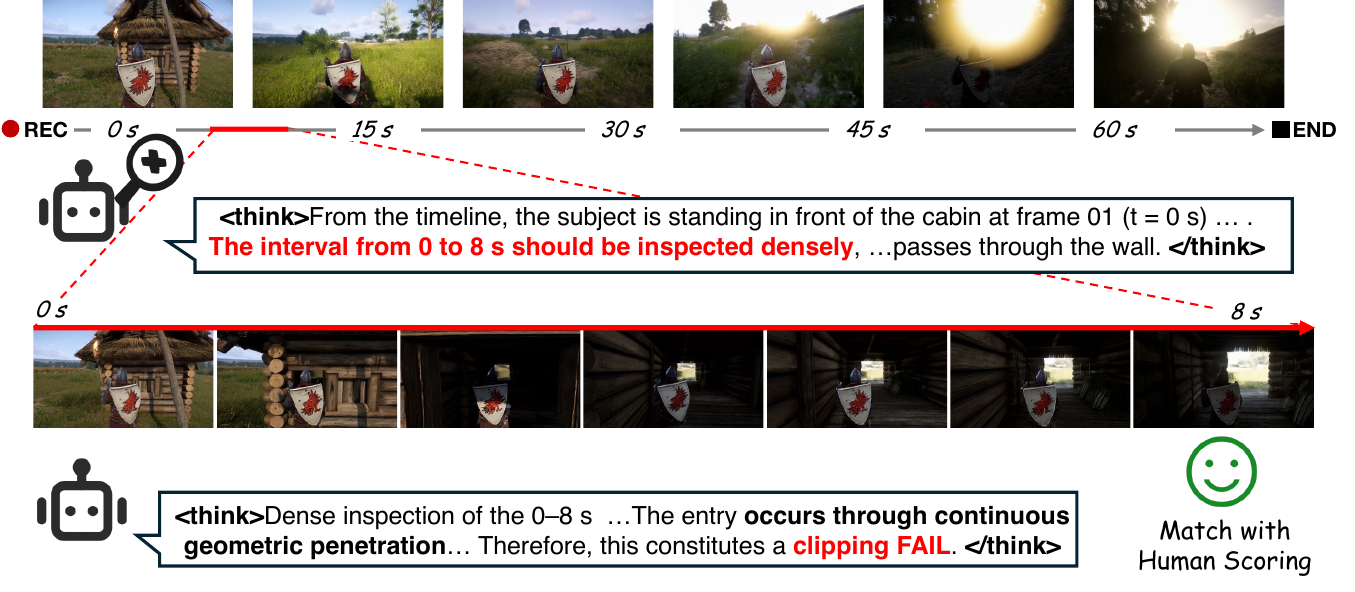}
\caption{\textbf{Physics VLM Verification.} The agent follows a coarse-to-fine scoring process: it first scans the full video timeline, progressively zooms in on the suspicious temporal interval for dense frame-level inspection, localizes the physical event, and assigns the final score.}
\label{fig:physics_vlm_verification}
\end{figure*}

From a calibration subset of 630 mechanics rollouts spanning 63 cases and 10 models, we randomly sample 100 videos for human evaluation. Twenty experts label each sampled video; the per-video majority vote serves as ground truth, and accuracy is the fraction of videos whose VLM verdict matches it, averaged over the ten models of the calibration subset. Iterative refinement of the mechanics prompts and the evidence-gathering harness increases accuracy against the human majority-vote ground truth from \textbf{52.5\%} (fixed template) to \textbf{73.2\%} on this calibration subset. Both percentages are computed on the same 100-video set used for calibration, rather than on held-out validation data.

\Cref{fig:physics_vlm_verification} visualizes a representative sampled rollout that agent successfully zooms in correct time-zone and scores matching with human ground-truth.

\subsubsection{Subject Memory VLM Verification}
\label{sec:appendix_memory_vlm_verification}

We randomly sample 60 TPV rollouts from Happy Oyster, Genie~3, LingBot-World, and HY-World~1.5, and ask 20 human experts to score each rollout on the same $1$--$10$ scale and rubric as \vlmjudge{} (\Cref{fig:subject_memory_prompt}), blind to the VLM score. Defining a (rollout, expert) pair as concordant when $|S_{\mathrm{VLM}} - S_{\mathrm{human}}| \leq 1$, the concordance rate over the $60 \times 20 = 1{,}200$ pairs reaches \textbf{88.3\%}, with a mean absolute difference of \textbf{0.83}, indicating that the judge tracks expert ratings closely under this tolerance. Establishing whether this gap lies within inter-human variation would require expert--expert agreement statistics. 

\begin{figure*}[p]
\centering
\includegraphics[width=0.90\textwidth]{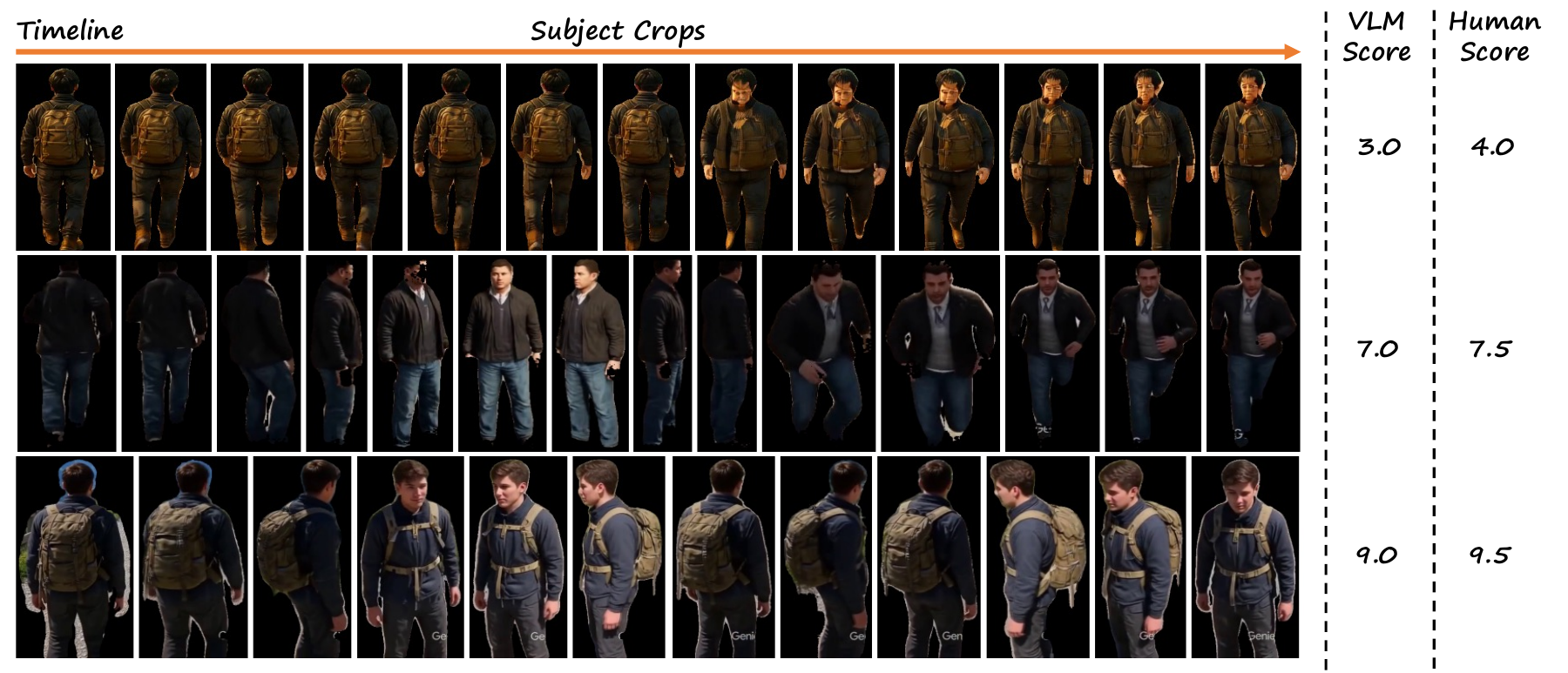}
\caption{\textbf{Subject-memory VLM verification examples.} Each row shows the temporally ordered subject crops fed to \vlmjudge{} for one sampled TPV rollout, together with the VLM score and the human expert scores on the same $1$--$10$ scale.}
\label{fig:subject_vlm_verification}
\end{figure*}

\Cref{fig:subject_vlm_verification} visualizes representative sampled rollouts as temporally ordered subject crops together with their VLM and human scores.

The three rows track subject fidelity over time: the high-scoring subject keeps facial and clothing details consistent across multiple viewpoint changes; the mid-scoring subject keeps overall body proportions but drifts in garment attributes (a tie appears in later crops); and the low-scoring subject changes in both body shape and accessories (a backpack persists after the subject turns around). In each case the VLM score lies within one point of the displayed human scores, illustrating the concordance criterion reported above.

\clearpage

\begin{figure*}[p]
\centering
\resizebox{0.90\textwidth}{!}{%
\begin{minipage}{\textwidth}
    \begin{tcolorbox}[d3box, width=\textwidth, title={TPV Mechanics Agent --- System Rubric (excerpt)}]
\ttfamily\scriptsize
You are a TPV mechanics evaluation agent. You judge ONE physical criterion per case (collision / clipping / deformation / gravity, including terrain adherence).\\
Loop (at most 4 rounds): (1) study the overview timeline; (2) call a tool to localize and densely inspect the critical event window; (3) either gather more evidence or emit a structured verdict.\\
Tools: inspect\_interval(start, end, fps, crop) --- dense frames from a time window at 1--8 fps; inspect\_frames(timestamps, crop) --- exact frames; crop in \{full, center, lower, upper\} at up to 768 px.\\
Verdict rules: PASS requires positive visual evidence; FAIL requires evidence of the violated constraint; INVALID only if the subject or target is unobservable.\\
Clipping: PASS iff the subject never traverses the obstacle (static, blocked, or a passage that visibly pre-exists contact); FAIL on body--obstacle interpenetration, the obstacle vanishing at the subject, or side-switching without a pre-existing passage.\\
The loop ends with an in-loop evidence audit: clipping PASS verdicts are re-examined for obstacle-topology artifacts; FAIL interaction verdicts get one high-recall positive-evidence pass before being committed.
\end{tcolorbox}

    \begin{tcolorbox}[d3box, width=\textwidth, title={Agent Tool Call (example)}]
\ttfamily\scriptsize
\{"thought": "Subject contacts the railing at t=16.6s; verify whether it stops or traverses.", "tool\_call": \{"name": "inspect\_interval", "args": \{"start": 16.0, "end": 24.0, "fps": 4.0, "crop": "center"\}\}\}
\end{tcolorbox}

    \begin{tcolorbox}[d3box, width=\textwidth, title={Structured Verdict (example)}]
\ttfamily\scriptsize
\{"verdict": "FAIL", "confidence": 0.95, "event\_interval": [16.6, 21.5], "subject\_crossed": true,\\
\ \ \quad "evidence": [\{"timestamp": 18.2, "observation": "railing absent at subject position"\}, \{"timestamp": 21.5, "observation": "subject on far side of railing"\}],\\
\ \ \quad "reason": "Subject switches sides while the railing disappears at its position; no pre-existing passage."\}
\end{tcolorbox}
\end{minipage}%
}
\caption{\textbf{TPV mechanics agent prompts.} Excerpt of the agent system rubric, an example tool call, and the structured verdict format. Per-criterion rubrics instantiate the same loop with criterion-specific evidence requirements.}
\label{fig:agent_prompts}
\end{figure*}

\begin{figure*}[p]
\centering
\resizebox{0.90\textwidth}{!}{%
\begin{minipage}{\textwidth}
    \begin{tcolorbox}[d3box, width=\textwidth, title={Reflection Evaluation Prompt --- Tier 1: Coarse Check (FPV)}]
\ttfamily\scriptsize
There is \{target\_object\} reflected in \{surface\_desc\}. Looking at the overall scene across all these frames, are the reflections physically plausible — i.e., consistent in shape, approximately symmetric about the reflecting surface, and changing naturally with the camera movement?\\

After your reasoning, conclude with exactly one line: Answer: yes or Answer: no
\end{tcolorbox}

    \begin{tcolorbox}[d3box, width=\textwidth, title={Reflection Evaluation Prompt --- Tier 1: Coarse Check (TPV)}]
\ttfamily\scriptsize
In this third-person view video, watch \{target\_object\} and its reflection in \{surface\_desc\}. As the subject moves, does the reflection behave physically correctly — i.e., consistent in shape, approximately symmetric about the reflecting surface, and moving consistently with \{target\_object\}?\\

After your reasoning, conclude with exactly one line: Answer: yes or Answer: no
\end{tcolorbox}

    \begin{tcolorbox}[d3box, width=\textwidth, title={Reflection Evaluation Prompt --- Tier 2: Per-Frame Scoring}]
\ttfamily\scriptsize
I show you two images. Image 1 is the REFERENCE first frame showing the original scene.\\
Image 2 is a frame from the generated video at timestamp \{timestamp\}s.\\

In the reference frame, \{target\_object\} is reflected in \{surface\_desc\}.\\

FIRST, check whether \{surface\_desc\} is clearly visible in Image 2. If the camera has moved so that \{surface\_desc\} is no longer in view, or is only a tiny sliver at the edge, score N/A — do NOT score 0 just because you cannot see a reflection. Score 0 means the surface IS visible but the reflection is wrong.\\

Score the reflection quality in Image 2:\\
- 2: Reflection is physically correct — properly flipped, content matches the real scene above, proportions and perspective are consistent with the viewing angle\\
- 1: Reflection exists but has physics errors — wrong flip direction, noticeable asymmetry, content partially mismatched, or unnatural distortion\\
- 0: The reflecting surface IS visible and occupies a reasonable area, but the reflection is missing, completely wrong, or shows unrelated content\\
- -1: INVALID — the scene has collapsed or deviated so far from the reference that reflection evaluation is not meaningful (e.g., completely different environment)\\
- N/A: The reflecting surface (\{surface\_desc\}) is NOT visible or barely visible in this frame — camera has moved away from it\\

After your reasoning, output exactly one line: Score: -1, Score: 0, Score: 1, Score: 2, or Score: N/A
\end{tcolorbox}
\end{minipage}%
}
\caption{\textbf{Reflection evaluation prompts.} Full VLM prompts for coarse reflection plausibility checks and per-frame reflection scoring.}
\label{fig:reflection_prompts}
\end{figure*}

\begin{figure*}[p]
\centering
\resizebox{0.90\textwidth}{!}{%
\begin{minipage}{\textwidth}
    \begin{tcolorbox}[d3box, width=\textwidth, title={Shadow Evaluation Prompt --- FPV with Reference Frame and Expected Shadow}]
\ttfamily\scriptsize
The FIRST image is the REFERENCE first frame showing the initial scene.\\
The remaining images are frames from the generated video.\\

The expected physical phenomenon is: \{expected\_shadow\}.\\

Compare the video frames against the reference. Does the generated video correctly show \{expected\_shadow\}?\\

Judgment criteria:\\
- A real shadow must have a recognizable silhouette shape matching the object that casts it (e.g., a person shadow looks like a person outline, a pillar shadow looks like an elongated rectangle)\\
- The shadow must have a clear boundary — vague dark patches, ambient shading, surface texture, or lighting gradients do NOT count\\
- The shadow direction must be physically consistent with visible light sources in the scene\\
- Answer 'yes' ONLY if you are confident the shadow is real and correct\\

After your reasoning, conclude with exactly one line: Answer: yes or Answer: no
\end{tcolorbox}

    \begin{tcolorbox}[d3box, width=\textwidth, title={Shadow Evaluation Prompt --- TPV with Expected Shadow}]
\ttfamily\scriptsize
In this third-person view video, \{expected\_shadow\} should be visible. Looking at these frames, does \{expected\_shadow\} appear and behave in a physically correct manner (consistent direction relative to the light source, moving with the subject)?\\

After your reasoning, conclude with exactly one line: Answer: yes or Answer: no
\end{tcolorbox}

    \begin{tcolorbox}[d3box, width=\textwidth, title={Shadow Evaluation Prompt --- TPV Fallback (No Expected Shadow)}]
\ttfamily\scriptsize
In this third-person view video, as the subject moves forward, does the subject cast a visible shadow on the ground or nearby surfaces (walls, floors, etc.)? If so, is the shadow shape recognizable and its direction consistent with the lighting in the scene?\\

After your reasoning, conclude with exactly one line: Answer: yes or Answer: no
\end{tcolorbox}

    \begin{tcolorbox}[d3box, width=\textwidth, title={Shadow Evaluation Prompt --- Generic Fallback}]
\ttfamily\scriptsize
In this scene there is a light source illuminating a wall or floor. As the camera moves, a shadow should appear or change on the surface. Looking at these frames from the latter portion of the video:\\

1. Is there a shadow visible on the wall or floor?\\
2. If yes, does the shadow move or change in a way that is physically consistent with the camera movement and the light source position?\\

Answer 'yes' if the shadow appears and behaves correctly, 'no' if the shadow is missing or behaves incorrectly.\\

After your reasoning, conclude with exactly one line: Answer: yes or Answer: no
\end{tcolorbox}
\end{minipage}%
}
\caption{\textbf{Occlusion and shadow evaluation prompts.} Full VLM prompts used for expected-shadow, fallback shadow, and generic shadow checks.}
\label{fig:occlusion_shadow_prompts}
\end{figure*}

\begin{figure*}[p]
\centering
\resizebox{0.90\textwidth}{!}{%
\begin{minipage}{\textwidth}
    \begin{tcolorbox}[d3box, width=\textwidth, title={Subject Memory Evaluation Prompt}]
\ttfamily\scriptsize
You are an expert evaluator of third-person video subject memory. The input is a temporally ordered sequence of subject crops from one generated video. The first crop is the reference subject; the remaining crops show the subject at later time points.\\
Evaluate whether the subject remains the same individual across the whole sequence, and output a video-level subject memory score from 1 to 10.\\

Assess five failure types:\\
1. Identity change: the subject becomes a different individual or category.\\
2. Structural distortion: the body, anatomy, proportions, limbs, head, face, or key parts become deformed or implausible.\\
3. Appearance drift: color, texture, clothing, hair, material, or style is truly rewritten.\\
4. Subject disappearance: the subject becomes partly or fully invisible.\\
5. Quality degradation: severe blur, low resolution, diffused edges, or loss of key details makes the subject hard to identify.\\

Important calibration rules:\\
- Viewpoint change alone is not inconsistency. Back-to-front, front-to-back, side-to-front, or far-to-close changes are acceptable if the subject can reasonably be the same individual under the new view.\\
- New frontal details, such as face or front clothing, should not be treated as contradictions when the reference was a back view.\\
- Impossible orientation conflicts are structural distortion. If the body is still back-facing but the face suddenly looks toward the camera, the score should be at most 5; if this persists or the head-body relation collapses, score 1--3.\\
- Same category does not imply consistency. If a person, animal, or vehicle remains in the same broad class but gradually becomes a blob, oval mass, fused-limb shape, or loses a clear part structure, assign a low score.\\
- Lighting and shadow changes should be treated leniently. If the subject becomes darker, bluish-gray, lower contrast, or slightly softer because of shadow, but structure, proportions, clothing shape, and identity cues remain stable, this is usually 8--9.\\
- Do not mark appearance drift unless color, texture, clothing, fur, or material changes in a way not explainable by lighting, exposure, or white balance.\\
- Do not mark structural distortion only because shadows, low contrast, or mild blur soften boundaries. Mark it only when real geometry or anatomy changes.\\
- Mild blur or reduced texture contrast can be ignored; severe blur, low resolution, or missing key details should reduce the score.\\
- Judge the full sequence holistically. Progressive deterioration is more serious than an isolated artifact.\\
- Hallucinated non-subject objects, such as a handheld item, bag, stick, or small prop, are not identity changes if the subject body and main appearance remain consistent.\\

Scoring guide:\\
10: Highly consistent identity, structure, proportions, appearance, and details.\\
8--9: Mostly consistent; only minor appearance variation, mild blur, normal viewpoint change, or lighting/shadow change.\\
7: Same subject with stable structure and main appearance, but noticeable darkening, mild-to-moderate blur, lower texture contrast, or edge softening.\\
6: Roughly consistent, but clear non-lighting appearance drift or persistent detail degradation begins to affect identity details.\\
4--5: Poor consistency, with real structural distortion, obvious appearance rewriting, partial disappearance, or severe blur.\\
2--3: Severe degradation, such as fused limbs, unclear head-body relation, blob/mass/oval collapse, or large-scale failure of key structure.\\
1: The subject clearly becomes a different object, disappears for most of the sequence, or can no longer be matched to the reference.\\

Output JSON only. Do not output anything else:\\
\{\\
\ \ \quad "subject\_memory\_score": [integer from 1 to 10],\\
\ \ \quad "identity\_change": [true or false],\\
\ \ \quad "structural\_distortion": [true or false],\\
\ \ \quad "appearance\_drift": [true or false],\\
\ \ \quad "subject\_disappearance": [true or false],\\
\ \ \quad "quality\_degradation": [true or false],\\
\ \ \quad "viewpoint\_change": [true or false],\\
\ \ \quad "lighting\_change": [true or false],\\
\ \ \quad "reason": [brief explanation of the scoring rationale]\\
\}
\end{tcolorbox}
\end{minipage}%
}
\caption{\textbf{Subject memory evaluation prompt.} Full VLM prompt used for holistic third-person subject-memory scoring and diagnostic flag extraction.}
\label{fig:subject_memory_prompt}
\end{figure*}

\end{document}